\documentclass[10pt,english]{article}

\usepackage{graphicx}                        
\usepackage{evolvingai}                      
\usepackage{textcomp}                        
\usepackage{gensymb}                         
\usepackage[numbers, sort&compress]{natbib}  
\usepackage{xr}                              
\usepackage[percent]{overpic}                
\usepackage{url}                             
\usepackage[T1]{fontenc}                     
\usepackage{babel}
\usepackage{calc}                            
\usepackage{setspace}                        

\newcommand{\captionstart}{}          
\usepackage[margin=1.875in, tmargin=1.5in]{geometry} 
\newcommand{\figwidth}{\textwidth}    
\newcommand{\documentfooter}{}        
\newcommand{\titlesettings}{          
\newgeometry{margin=1.5in, tmargin=1.5in}
}
\newcommand{\abstractsettings}{}      
\newcommand{\bodysettings}{           
\newgeometry{margin=1.875in, tmargin=1.5in}
} 

\addeditor{jh}{olive}
\addeditor{ken}{red}

\newcommand{\dir}{}     

\newcommand{\picbreeder}{Picbreeder}        
\newcommand{\picbreederweb}{Picbreeder.org}

\newcommand{\labfigo}[4]{%
\begin{overpic}[width=#2\figwidth]{#4}%
\put(-1,#3){(#1)}%
\end{overpic}}

\newcommand{\labfigof}[5]{%
\begin{overpic}[width=#2\figwidth]{#5}%
\put(#4,#3){(#1)}%
\end{overpic}}

\renewenvironment{abstract}
 {\small
  \begin{center}
  \bfseries \abstractname\vspace{-.5em}\vspace{0pt}
  \end{center}
  \list{}{
    \setlength{\leftmargin}{0.75in}%
    \setlength{\rightmargin}{\leftmargin}%
  }%
  \item\relax}
 {\endlist}
 
\usepackage{fancyhdr}

\begin{document}

\title{The Emergence of Canalization and Evolvability in an Open-Ended, Interactive Evolutionary System}

\author{  
    \small Joost Huizinga\\ 
    \small Evolving AI Lab\\ 
    \small Dept. of Computer Science\\ 
    \small University of Wyoming\\ 
    \small \texttt{jeffclune@uwyo.edu}\\
    \small Uber AI Labs\\ 
    \small \texttt{jhuizinga@uber.com}\\
    \and
    \small Kenneth O.~Stanley\\ 
    \small EPLex\\ 
    \small Dept. of Computer Science\\ 
    \small University of Central Florida\\
    \small \texttt{kstanley@cs.ucf.edu}\\
    \and
    \small Jeff Clune\footnote{Contact author}\\ 
    \small Evolving AI Lab\\ 
    \small Dept. of Computer Science\\ 
    \small University of Wyoming\\ 
    \small \texttt{jeffclune@uwyo.edu}\\
    \small Uber AI Labs\\ 
    \small \texttt{jeffclune@uber.com}\\
}

\date{}

\titlesettings
\maketitle
\thispagestyle{fancy}


\abstractsettings
\begin{abstract}
Many believe that an essential component for the discovery of the tremendous diversity in natural organisms was the evolution of evolvability, whereby evolution speeds up its ability to innovate by generating a more adaptive pool of offspring. One hypothesized mechanism for evolvability is developmental canalization, wherein certain dimensions of variation become more likely to be traversed and others are prevented from being explored (e.g.\ offspring tend to have similarly sized legs, and mutations affect the length of both legs, not each leg individually). While ubiquitous in nature, canalization is rarely reported in computational simulations of evolution, which deprives us of in silico examples of canalization to study and raises the question of which conditions give rise to this form of evolvability. Answering this question would shed light on why such evolvability emerged naturally and could accelerate engineering efforts to harness evolution to solve important engineering challenges.
In this paper, we reveal a unique system in which canalization did emerge in computational evolution.  We document that genomes entrench certain dimensions of variation that were frequently explored during their evolutionary history. 
The genetic representation of these organisms also evolved to be more modular and hierarchical than expected by random chance, and we show that these organizational properties correlate with increased fitness. 
Interestingly, the type of computational evolutionary experiment that produced this evolvability was very different from traditional digital evolution in that there was no objective, suggesting that open-ended, divergent evolutionary processes may be necessary for the evolution of evolvability. 
\end{abstract}
\bodysettings

\newpage
\section{Introduction}

The functional organisms produced by natural evolution are unfathomably diverse,
from single celled-bacteria like E. Coli to large mammals like elephants. 
The success of natural evolution is especially remarkable when one considers that
it is fueled by mostly random and unrelated changes at the genetic 
level~\cite{lynch2008genome, maki2002origins, kirschner1998evolvability, parter2008facilitated, kirschner2006plausibility}. 
As such, it is believed that natural evolution was aided by the emergence of
\emph{evolvability}~\cite{wagner1996complex,kirschner1998evolvability,abzhanov2006calmodulin,
hendrikse2007evolvability, Gerhart2007, parter2008facilitated, pigliucci2008evolvability, 
conrad1979bootstrapping, dawkins1989evolution, hindre2012new}, i.e.\ 
the emergence of genetic properties that increase the effectiveness of evolution.

Evolvability in natural systems is facilitated by many different innovations, 
a few of which are genetic structures like Hox genes~\cite{wagner2003hox, pearson2005modulating}, 
sexual reproduction~\cite{misevic2006sexual, wagner1996complex, hamilton1990sexual}, 
the evolution of mutation rates~\cite{partridge2000natural, bedau2003evolution}, 
structural organization in the form of modularity and 
hierarchy~\cite{wagner1996complex, Kashtan2005, mengistu2016evolutionary, 
clune2013originModularity, espinosa2010specialization} (discussed in greater detail below), 
and the emergence of standardized body plans~\cite{willmore2012body,fish2011satb2}.
In this paper, we will focus on a particularly interesting driver of evolvability 
known as \emph{developmental canalization}~\cite{waddington1942canalization, clarke1987developmental, 
debat2001mapping, stanley2003taxonomy},
which ties together many of the aforementioned concepts.
Canalization is the process whereby certain phenotypic dimensions of variation become resistant to genetic changes
such that other, possibly more adaptive dimensions of variation are more likely to be explored.
Here, a \emph{dimension of variation} refers to a phenotypic trait that can vary individually,
or a set of phenotypic traits that vary in concert.
For example, change in the length of the right leg of a human would be one dimension of variation,
and coordinated change in both legs represents another dimension of variation.
As it turns out, in humans it is rare that one leg becomes substantially 
longer or shorter than the other~\cite{auerbach2006limb}, 
but there exists considerable variation in leg length between individuals~\cite{bogin2010leg},
indicating that variations in human leg length have been canalized 
(specifically, the ability to individually vary leg lengths has been reduced, 
and the ability to vary both at once has been created).

Canalization is ubiquitous in natural systems and, 
as result, one might expect that forms of canalization are consistently 
encountered in models and computational simulations of evolution as well.
The opposite appears to be true;
forms of canalization are rarely reported in computational simulations of evolution 
despite significant efforts to promote and discover 
it~\cite{draghi2008evolution, kouvaris2015evolution, clune2010investigating, szejka2007evolution, 
bassler2004evolution, wagner1996does, wagner1997population, rice1998evolution, ancel2000plasticity, siegal2002waddington},
suggesting that these simulations do not adequately represent the full capacity of natural evolution.
In contrast, this paper uniquely showcases evidence for the spontaneous emergence
of canalization in Picbreeder, an interactive and open-ended system of
simulated evolution, and we discuss why and how such canalizations may have
emerged in this system, but not in others (Fig.~\ref{fig:concept}).

\newcommand{\figconcept}[1]{%
\begin{figure}[tbp]
\centering
\includegraphics[width=1\figwidth]{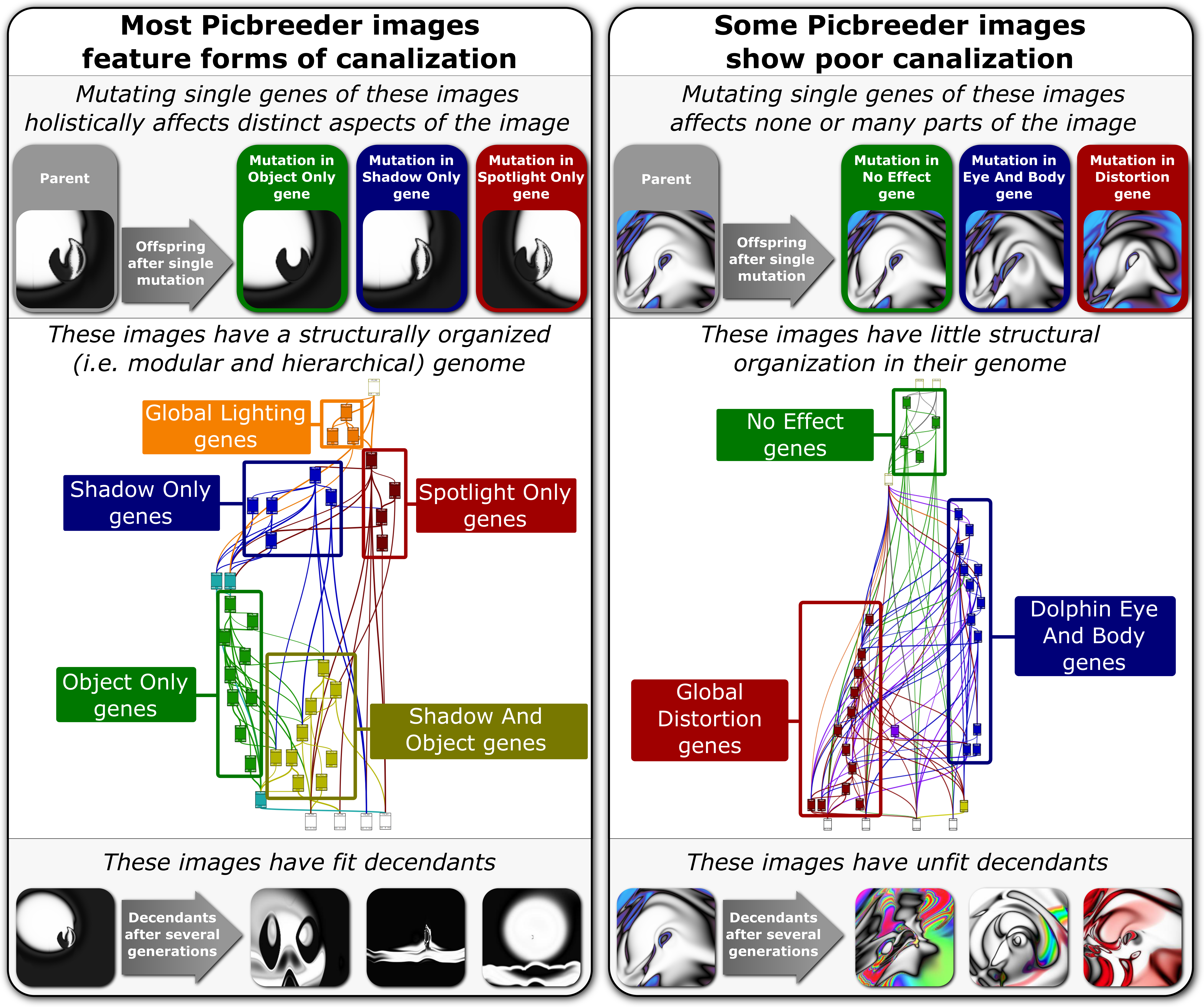}
\caption{\captionstart{}\textbf{\picbreeder{} images that have canalized dimensions of variation display structural organization in their genome and have higher quality descendants.} 
The left and right panel show the properties of two different \picbreeder{} images. 
\textbf{Left:} properties for a Picbreeder image that has canalized many intuitive dimensions of variation.
\textbf{Right:} properties for a Picbreeder image that has canalized only a few, unhelpful dimensions of variation.
The top row shows the original image, and three different variants accessible through a single mutation. 
Here, the variants in the left panel have been enlarged to better show how a single mutation 
can affect very specific parts of the image.
The middle row shows the genome of the image, with colored boxes 
around groups of genes that affect similar parts of the image. 
The bottom row shows three different descendants of the image. 
The three properties are correlated; 
images that have canalized interesting dimensions of variation tend to have structurally organized genomes, 
and their descendants are often of higher quality.
}
\label{fig:#1}
\end{figure}
}
\figconcept{concept}

It is important to note that the emergence of evolvability, including canalization, does not require 
the evolutionary process to have knowledge about future environmental changes; 
that is, evolvability is not a form of ``directed evolution''~\cite{lenski1993directed}. 
Instead, it is widely believed that evolvability can emerge based on the evolutionary history of 
lineages~\cite{dawkins1989evolution, Gerhart2007, pigliucci2008evolvability, 
conrad1979bootstrapping, kouvaris2015evolution, watson2016can, kounios2005resolving}.
In short, individuals whose genome is structured such that beneficial mutations are more likely
and detrimental mutations are less likely
have a better chance of producing viable offspring,
meaning such evolvability can be directly selected through the benefits it provides.
Provided that some forms of selection are persistent over evolutionary time while others vary,
such as mismatching legs always being detrimental while optimal leg length varies over time,
genetic structures that increased the probability of beneficial mutations in the past may do 
so in future environments as well.

Despite the fact that evolvability and canalization are often regarded as essential for the evolution of complex organisms,
their origins remain an active topic of research and 
debate~\cite{wagner1996complex,kirschner1998evolvability,abzhanov2006calmodulin,
hendrikse2007evolvability, Gerhart2007, parter2008facilitated, pigliucci2008evolvability, 
conrad1979bootstrapping, dawkins1989evolution, hindre2012new}.
The main challenge in answering questions regarding the origins of evolvability and canalization
is that they are difficult to study \emph{in vivo};
oftentimes, properties of interest can be difficult to measure, change, 
or control for~\cite{paroo2004challenges, hindre2012new}.
In addition, biological populations evolve slowly;
even rapidly reproducing microorganisms take on the order of weeks to experience 
a few hundred generations of evolution~\cite{dykhuizen1990experimental},
and the findings from these microorganisms do not necessarily generalize to their more slowly reproducing
counterparts~\cite{levin2000bacteria}.

An alternative is to study these questions in computational simulations of evolution instead.
While they may not seem as compelling as \emph{in vivo} experiments,
computational simulations can greatly improve our understanding of evolutionary processes,
and have shed light on a variety of complex evolutionary questions, including the evolution of altruism~\cite{clune2011selective, montanier2013evolution}, 
structural organization such as 
modularity~\cite{Kashtan2005, clune2013originModularity, espinosa2010specialization, wagner2005natural, Wagner2007, kashtan2007varying}, 
regularity~\cite{stanley2009hypercube, huizinga2014evolving}, 
and hierarchy~\cite{mengistu2016evolutionary, Corominas-Murtra2013}, 
mutation rates~\cite{clune2005investigations, wilke2001evolution, clune2008natural, bedau2003evolution}, 
sexual reproduction~\cite{misevic2006sexual, azevedo2006sexual},
genomic complexity~\cite{lenski2003evolutionary, lenski1999genome, wagner1996complex}, 
gene duplication~\cite{wagner2003hox, kuo2006network, pastor2003evolving}, 
and coevolution~\cite{zaman2011rapid, fortuna2013evolving, palmer2012evolved}, to name but a few.
Computational simulations are particularly attractive because the experimenters have full control over all variables 
involved in the evolutionary processes
and, provided that the fitness function is simple, modern hardware can run thousands of 
generations of evolution in just a couple of days,
allowing for rapid prototyping of hypothesis.

Unfortunately, while canalization is ubiquitous in nature~\cite{waddington1942canalization, 
clarke1987developmental, debat2001mapping},
clear examples of canalization in computational simulations are rare~\cite{draghi2008evolution, 
kouvaris2015evolution, clune2010investigating, szejka2007evolution, 
bassler2004evolution, wagner1996does, wagner1997population, rice1998evolution, ancel2000plasticity, siegal2002waddington, grimbleby2000automatic, filliat1999evolution, gallagher1996application},
meaning that we lack a proper starting point from which to conduct experiments because
we can not study canalization in computational simulations of evolution if we can not produce it in the first place.
The fact that we do not know how to reproduce canalization also means that, 
when tackling challenging engineering problems with the help of evolutionary 
algorithms~\cite{cully2015robots, livingston2016modularity, clune2009evolving, ellefsen2016planning, lohn2005evolved, chung2012structural, chiel1992robustness, lee2013evolving, yosinski2011gaits, lipson2000automatic, pallez2007eye, sims1993interactive},
those algorithms are missing a key property that made natural evolution successful,
possibly explaining why most evolutionary algorithm research restricts itself to fairly simple,
uni-modal tasks~\cite{cheney2013unshackling, blynel2003exploring, Kashtan2005}.

While such experiments are rare, the following investigations in computational 
simulation did touch upon the principles of canalization.
Draghi and Wagner demonstrated the evolution of evolvability 
in a model where the sum of two vectors needed to reach a target point in a two-dimensional space~\cite{draghi2008evolution}. 
Vectors were specified by angle and magnitude, 
but angle mutations were much less common than magnitude mutations. 
After evolution, the angles between vectors would reflect the evolutionary history; 
if the target point remained stationary, the angles between vectors was arbitrary, 
but if the target point changed frequently, the angles between vectors were close to $90\degree$, 
such that the entire space of possibly fit phenotypes could be quickly reached through magnitude mutations alone.
Here, the angle between the two vectors controlled
which dimensions of variation were more or less likely to be explored,
and the evolutionary history determined which angle became fixed in the population,
thus representing a rudimentary form of canalization.

Another form of canalization was demonstrated by Kouvaris et al., who 
worked with a model where groups of phenotypic traits had to be co-expressed to gain fitness~\cite{kouvaris2015evolution}.
That is, the front wings, hind wings, and antennae of an abstract insect consisted of several parts,
and all parts needed to be expressed simultaneously to form a functional body part.
The environment cycled between favoring individuals with front wings and antennae,
favoring individuals with both front and hind wings but no antennae,
or favoring insects without any of these traits.
Provided that the environment changed at the right frequency,
individuals evolved such that mutations would either express or 
repress entire groups (i.e. modules) of phenotypic traits
(e.g. complete wings or complete antennae),
but never cause partial expression within a group.
These groups of phenotypic traits presented a clear example of canalization,
although achieving this effect required a fairly strict set of environmental conditions to emerge.

As we will present in this paper, a possible source of canalization is
genotypic structural organization in the forms of \emph{modularity} and \emph{hierarchy}.
Following a conventional definition~\cite{clune2013originModularity, Kashtan2005, 
espinosa2010specialization, meunier2010hierarchical, klingenberg2005developmental, sporns2016modular}, 
a genome is considered modular if it consists of groups of genes that 
have many interactions with genes in the same group, 
but few interactions with genes in other groups.
A genome is considered hierarchical if interactions result in an ordered structure, 
such that interactions predominantly go from high-level structures, 
which tend to have global effects, to low-level structures, 
which are generally associated with local changes.
Structural organization in terms of modularity, hierarchy, or both, have been found
in the gene regulatory networks of many species, including E. Coli~\cite{ravasz2002hierarchical},
sea urchin~\cite{peter2009modularity}, yeast~\cite{lee2004probabilistic}, and Drosophila~\cite{olson2006gene}.
Such structural organization can lead to canalization if it changes the 
likelihood with which phenotypic traits will change.
For example, if two phenotypic traits are encoded by a single genotypic module,
a single mutation is likely to affect both traits,
whereas if the two traits are encoded by separate modules,
there is a better chance that only one of those traits is affected.
Similarly, if a genotype is hierarchically organized,
a single mutation to a high-level component is likely to affect many phenotypic traits simultaneously,
whereas a single mutation to a low-level genotypic component will probably only affect a single trait.

Research regarding structural organization has shown that,
when individuals needed to adapt to a modularly changed environment
(i.e. the overall goal in the changed environment would differ, 
but many of the sub-problems in the environment would remain unchanged),
structurally organized individuals both evolved and had increased evolvability (i.e. adapted faster)
compared to unstructured individuals~\cite{Kashtan2005, 
clune2013originModularity, espinosa2010specialization, mengistu2016evolutionary}.
These experiments implicitly also demonstrated a form of canalization,
because structurally organized individuals were much more likely to rewire sub-problems than unstructured individuals.
In other words, for structurally organized individuals, 
dimensions of variation related to environmental sub-problems were much more likely to be explored,
whereas dimensions of variation related to more holistic changes in the behavior 
of the individual were less likely to be explored.
However, most of this research focused on individuals with a direct encoding 
(i.e.\ the phenotype and genotype are equivalent),
precluding the wide array of genetic interactions present in biological 
organisms~\cite{waddington1942canalization, clarke1987developmental, debat2001mapping, fish2011satb2}.
Research that did examine the effects of structural organization with a developmental encoding
did not report on forms of canalization~\cite{huizinga2014evolving, huizinga2016aligning}.

The above experiments provide some proofs of concept for 
the evolution of canalization in computational simulations,
but their models are simple.
The present research shows the evolution of canalization in a more complex, open-ended system:
the images evolved on \picbreederweb{}, a website for the 
interactive evolution of pictures (Fig.~\ref{fig:concept} top).
We also show that many \picbreeder{} genomes display structural organization 
in the forms of modularity and hierarchy,
and we present examples where the structural organization directly 
corresponds to the observed canalizations (Fig.~\ref{fig:concept} middle).
In addition, the results suggest that these structurally organized 
genomes are generally more fit in terms of offspring (Fig.~\ref{fig:concept} bottom).
Lastly, we will discuss the differences between \picbreeder{} and other computational simulations of evolution,
and argue that the emergence of canalization may be directly facilitated by the ever changing, divergent,
goalless nature of \picbreeder{}.
The implication is that, as has been recently 
argued~\cite{lehman2016critical, pugh2016quality, lehman2012benefits, stanley2015greatness}, 
the success of natural evolution may not be due to short-term competition over common resources,
but is enabled instead by the long-term tendency to invade new niches and avoid competition altogether.

\section{Methods}

\subsection{\picbreeder}
\label{sec:picbreeder}

\picbreederweb{} is a website, first presented by \citet{secretan2008picbreeder},
where users can interactively and collaboratively evolve images. 
Users visiting the site can ``breed'' images similar to how one might breed livestock;
the user starts with an initial population of images from which 
the user can select the images he or she finds most promising.
Those will then be mated and mutated to form the next generation of images, and the process repeats.
The user can continue this process until satisfied or bored, 
and can then choose to publish the result to the website, 
such that the results can serve as a seed for other users. 
Since its inception, over $10,000$ images have been published on \picbreeder{}~\cite{picbreeder2016statistics}. 

The evolutionary process is driven by the 
NeuroEvolution of Augmenting Topologies (NEAT) evolutionary algorithm~\cite{stanley2002evolving}.
NEAT is an algorithm for the evolution of networks. It starts with simple networks,
and slowly increases the size of the networks by adding nodes and connections.
To evolve images with NEAT, the images are represented through an artificial 
genetic encoding called Compositional Pattern Producing Networks (CPPNs), 
as described in Section~\ref{sec:cppn}.
Whenever a CPPN is mutated, every weight in the network has a chance of being changed 
by replacing it with a random number drawn from a normal distribution 
with a mean equal to the original weight of the connection and a variance of 1.
In addition, there is a small chance of adding a connection between two unconnected nodes,
and there is a small chance of adding a new node onto an existing connection.
When multiple images are selected, their underlying CPPNs may be combined through crossover.
To perform crossover between networks, following the convention of NEAT, nodes and connections in the 
network are first aligned by matching \emph{historical markings}; 
unique identifiers that are assigned to every node and connection the first time they are added to a CPPN.
Nodes and connections that are present in both parents will be randomly selected from either parent,
whereas nodes and connections only present in one parent will always be added.
The original NEAT algorithm also includes fitness sharing through speciation, 
added to preserve diversity within the population,
but fitness sharing is not in effect on \picbreeder{} because the 
individuals that get to reproduce are directly chosen by the user.
Further details and parameters are described in~\cite{secretan2008picbreeder} and~\cite{secretan2011picbreeder}.

When starting evolution, the user can choose to start from scratch, or to branch from an existing image.
If the user chooses to start from scratch, 
the initial population of images will consist of simple geometric patterns, 
as specified by the initial small, randomized genomes (Sec.~\ref{sec:cppn}). 
However, if the user chooses to start from an existing image, 
the initial population will consist of direct offspring of the selected image. 
For the purpose of measuring the reproductive success of an image, 
we define the \emph{fitness} of an image as the number of \emph{direct descendants} of that image, 
where a direct descendant is defined as an image that was branched, evolved, and 
published directly by a single user from the original image
without any of the intermediate forms being published.
This measure of fitness encapsulates both the quality of the parent,
because interesting images have a higher chance of being selected by a user for further evolution,
and the evolvability of the parent,
because users are unlikely to publish descendants if they were 
unable to introduce any interesting new changes in said descendants.
This metric is noisy (an image placed on the front page for a long period of time, 
such as an ``editors pick'' or top-rated image, 
may have many more descendants than a qualitatively similar image that did not make it to the front page), 
but it is arguably informative when taken in aggregate.

It is important to note that, in contrast to most classic experiments with evolutionary algorithms, 
\picbreeder{} has no overall goal. 
That is, while any individual user will select images that are aesthetically pleasing or interesting in some way, 
there is no ``target image'' that needs to be found. 
In addition, while users may form goals during a session, long lineages are often evolved by many different users, 
who may all have different strategies and motivations during image selection.

\subsection{CPPNs}
\label{sec:cppn}

The genomes of the \picbreeder{} images are known as Compositional Pattern Producing Networks 
(CPPNs)~\cite{stanley2007compositional},
which are an abstraction of developmental processes. 
CPPNs have been described at length many times previously~\cite{stanley2009hypercube, 
stanley2007compositional, gauci2007generating, d2007novel, cheney2013unshackling, secretan2011picbreeder},
so here we only briefly describe them and how they abstract developmental biology.
Consider the development of any multi-cellular organism:
The organism will start as a single stem cell, which will multiply over time to form the mass of the organism. 
To form different functional parts of the organism, stem cells will have to determine what kind of cell to become 
(e.g.\ muscle, bone, neuronal, etc.), i.e. their \emph{cell fate}. 
The proper fate of a cell depends on its location in the developing organism; 
a cranial cell may have to become part of the central nervous system, 
whereas a distal cell may have to become part of a claw. 
In developing organisms, a cell can effectively glean its location by measuring the concentrations 
of different proteins and other chemicals, jointly referred to as \emph{morphogens}~\cite{turing1952chemical}, 
which form gradients throughout the developing organism. 
For example, if there exists a morphogen that is only produced at the extreme anterior of the organism, 
but slowly diffuses throughout the entire organism,
the concentration of that morphogen provides location information with respect to the anteroposterior 
(front to back) axis. 
If a sufficient number of these morphogens are present over 
different axis (anteroposterior, dorsoventral, mediolateral, etc.), 
a cell can determine its location and hence its fate.

\newcommand{\figcppn}[1]{%
\begin{figure}[tb!]
\centering
\includegraphics[width=1\figwidth]{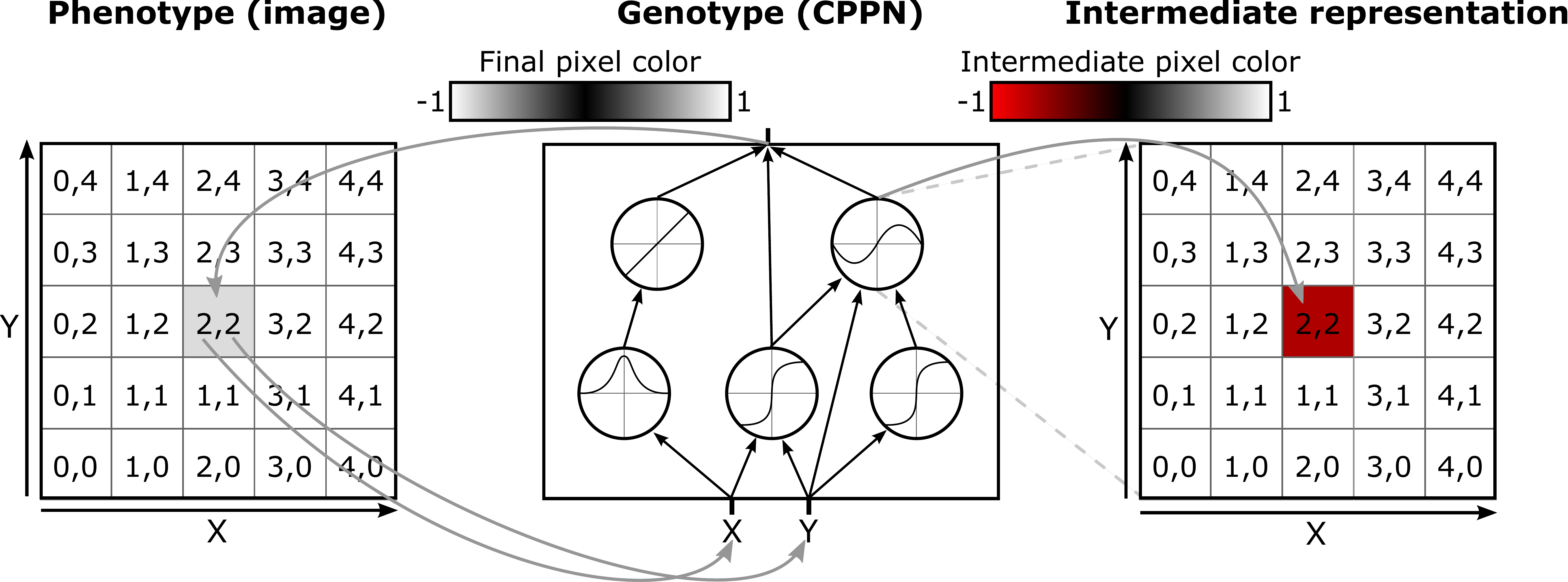}
\caption{\captionstart{}\textbf{CPPNs are functions from position to fate.} 
\textbf{Left:} To encode an image with a CPPN, every pixel in the image needs to have a geometric coordinate. 
\textbf{Middle:} The geometric coordinate of each pixel is then passed to the CPPN, 
which computes one or more values that determine the color of the pixel. 
For grayscale images (shown), the network will have a single output determining pixel intensity, 
whereas for color images the network will have two additional outputs for hue and saturation. 
\textbf{Right:} The activation of intermediate nodes in a CPPN is visualized in the 
same way that the final output of the network is visualized, namely,  
by mapping the output of the intermediate node to pixel color. 
However, to make the functioning of intermediate nodes more clear, 
negative values for these nodes are mapped to red, rather than white.}
\label{fig:#1}
\end{figure}
}
\figcppn{cppn}

While these morphogens are effective at signaling position information to developing cells, 
simulating the diffusion of such morphogens is computationally expensive,
which is why simulated diffusion-based artificial organisms are often restricted in size~\cite{bongard2003evolving}. 
However, in computational simulations of development, global positional information can be relayed directly to a cell, 
without the need to simulate diffusion. 
Inspired by this idea, CPPNs are functions from global positional information to cell fate;
they take the position of a cell, such as the x and y coordinates of a pixel, 
and return its fate, in this case the color value (Fig.~\ref{fig:cppn}).

An arbitrary function from position to cell fate is not sufficient to capture the power of developmental biology.
For example, diffusing chemicals can spread smoothly in all directions, 
giving rise to symmetry.
Genes can respond to their own gradients, enabling repeated patterns.
Genes can also compose different gradients by 
responding only when multiple different morphogens are present (or absent) at the same time.
To capture these properties, CPPNs are compositions of regular functions with specific behaviors; for example, 
Gaussian functions can provide symmetry, sine waves can provide repetition, 
and step functions like sigmoids can confer the ability to respond only when all necessary gradients are present
(Fig.~\ref{fig:cppn}, middle).
	
Each node in the CPPN is associated with one of these functions, 
and nodes interact with each other through weighted connections. 
At any node, the incoming values are multiplied by the corresponding connection weights, 
and the sum is passed to the function of that node. The input to the network is a geometric coordinate, 
and the output represents the cell fate, which will be the color of a pixel in an image. 

Within a CPPN, every node can be considered as a gene that produces a unique morphogen,
and the output of the node can be considered as the expression pattern of that gene.
Because the \picbreeder{} CPPNs describe 2-D images, the expression of an intermediate node can
be visualized by creating a 2-D image where each pixel is colored according to the 
output of the node at that location (Fig.~\ref{fig:cppn}, right).
For the purpose of such visualizations, pixels for intermediate nodes are colored from red ($-1$), 
to black ($0$), to white ($1$). 

Different paths in the CPPN may result in different intermediate patterns, 
which may later be combined to form the final output. 
As such, CPPNs can model many different interactions including pleiotropy, 
redundancy, and different developmental pathways without sacrificing computational tractability,
making the model appropriate for the study of evolvability and canalization.

\subsection{Analysis}
\label{sec:examiner}

To analyze the CPPNs produced by \picbreeder, we developed a tool called CPPN-Examiner (CPPN-X). 
It makes it possible to pick any connection in the network and slowly change its weight
while directly observing the effect on the pattern produced.
The tool also allows labeling these connections and thereby makes it possible to create a fully annotated version 
of the network (e.g. Fig.~\ref{fig:networkTrans} right).
To ensure that our analysis is relevant to all Picbreeder images,
we faithfully modeled the CPPN-X tool after the online code base,
written by Secretan et al.~\cite{secretan2008picbreeder, secretan2011picbreeder}.
We opted for creating a separate, offline tool to allow for additional computational optimizations of the CPPN,
which greatly increased the speed at which the program can render the effect of weight changes,
enabling us to view the effect of weight changes smoothly and in real time.
The source code is available at \url{www.evolvingai.org/CPPN-X}.

Connections in genomes were annotated according to the following procedure. 
First, a not-yet-labeled connection was selected and we swept across the possible values for 
that connection from its minimum ($-3$) to its maximum value ($3$) at a $0.1$ interval, 
viewing each intermediate image produced.
In some rare cases, the $0.1$ interval would be too coarse to properly observe the effect of the sweep and, 
in those cases, we decreased the interval to $0.01$.
Then, we qualitatively classified the resulting change and annotated the connection accordingly. 
When assigning a label, we ignored background changes that occurred when the weight got far (generally 1 unit or more) from its original value (see SI Fig.~\ref{SI-fig:sweep} for examples). We ignored these changes because we are interested in the effect of small genetic mutations, as those are considered to be an effective basis for modeling evolutionary processes~\cite{bulmer1971effect}. It is likely that every functional connection will cause some background changes if its weight is changed by a sufficiently large number and, as such, we do not believe that this effect should define the primary function of that connection.

After all connections were labeled, 
we merged classes with similar effects 
(e.g. merged classes such as ``Move Spotlight Left-Right'' and ``Move Spotlight Up-Down'' into a ``Spotlight Only'' class)
to reveal the high-level functional decompositions of the genomes.
In effect, this kind of manual experimentation and annotation is like a 
kind of artificial bioinformatics for Picbreeder CPPNs.
Files containing the fine-grained decompositions are available for 
download at: \url{www.evolvingAI.org/PicbreederCanalization}.

Nodes were assigned labels according to the majority label among their incoming connections.
Because we did not vary any attributes of the nodes,
this labeling holds no additional information,
and only serves to improve visual clarity.

It is important to note that this analysis of canalization is inherently subjective,
as it requires a human observer to classify the nature of each change.
While an objective measure of canalization would have been preferable, 
to the best of our knowledge, 
there does not exist an appropriate, objective measure of canalization for images at this time.
For example, naive metrics, such as localized change of pixel values in response to mutations,
do not cover all relevant forms of canalization.
Moving an object from one location in an image to another location 
in the image without changing the shape of the object is considered an important form of canalization,
but such canalization does not result in localized changes in pixel intensities.
Conversely, a local change in pixel intensities affecting arbitrary parts 
of different objects in an image is generally not considered a form of canalization,
yet it would be valued as such.
It may seem that such issues could be resolved by employing techniques based on automated object 
recognition systems (e.g. deep neural networks~\cite{krizhevsky2012imagenet}),
as these systems are often associated with a sense of objectivity not attributed to human observers.
However, there are still many problems that need to be resolved before an 
automated system can replace a human observer in the current domain, 
as these systems may see objects that are not actually there~\cite{nguyen2015deep},
may misclassify objects due to imperceptible changes~\cite{szegedy2013intriguing}, 
and can inherit their own bias from the training data~\cite{kamishima2011fairness}.

Despite the lack of an appropriate method for objectively measuring canalization, 
we do not believe that studying canalization in images should be avoided 
just because humans are (currently) the only agents capable of properly interpreting the data.
Indeed, many fields, such as those that study animal and human behavior, 
have to rely on human judgments (e.g. of whether two animals are fighting, cooperating, hugging, 
etc.)~\cite{dawkins2007observing, wemelsfelder2001assessing, altmann1974observational}.
As in those fields, and as has been argued before specifically in the context of 
harnessing human judgements in evaluating evolutionary algorithms~\cite{stanley2016art}, 
while it is important to note that the judgments are made by humans, and are thus subjective, 
more is learned through good, albeit imperfect, measuring devices than by not performing any measurements at all.
In addition, to facilitate an open discussion regarding our results and interpretations, 
many examples are included in this paper and its supplementary material, 
and the visualization tool, capable of accessing and analyzing our complete data set of Picbreeder genomes,
is freely available so that readers may judge for themselves 
the extent to which they agree with our subjective interpretations.

To validate that the previously described labeling process was fair and not biased by our knowledge of the hypothesis of canalization, 
we tested whether independent people would provide similar labels as those presented in this paper.
To do so, we compared the labeling presented in this paper against the labels provided by individuals recruited through Amazon's Mechanical Turk program,
a service where one can pay workers to conduct arbitrary tasks online.
Because it was too expensive to obtain the number of labels necessary for proper statistical analysis (which was at least 30 labels per connection) 
for every connection in every genome analyzed in this paper and its SI, we were only able to conduct this validation for a single genome.
However, because the process used to label all images was the same (and we did not know at that time we would perform extra validation on any genome), 
if the process is found to be sufficiently accurate for one genome, it is likely that the process was accurate for all genomes. 
We decided to conducted the test on the central, focal genome-and-image pair presented in this paper named ``Spotlight Casting Shadow'' (Fig.~\ref{fig:networkTrans}).
The results show that the Mechanical Turk workers, who were independent and not informed of our hypothesis,
assigned labels to connections similar to those in the presented labeling (SI Fig.~\ref{SI-fig:mtgenome})
and that the presented labeling is not an outlier among the Mechanical Turk workers on three different metrics
that measure to what degree a labeling matches the aggregate data obtained from the Mechanical Turk experiment (SI Fig.~\ref{SI-fig:match}).
As such, the analysis confirms that our labeling process was indeed fair and consistent with the labelings obtained from independent individuals.
The full analysis and experimental details can be found in the SI (SI Sec.~\ref{SI-sec:mechanical_turk}).

\section{Results}
\label{sec:results}

\newcommand{\figotherimages}[1]{%
\begin{figure}
\begin{center}
Apple body size\\
\renewcommand{\dir}{Figures/DovApple}
\includegraphics[width=0.19\figwidth]{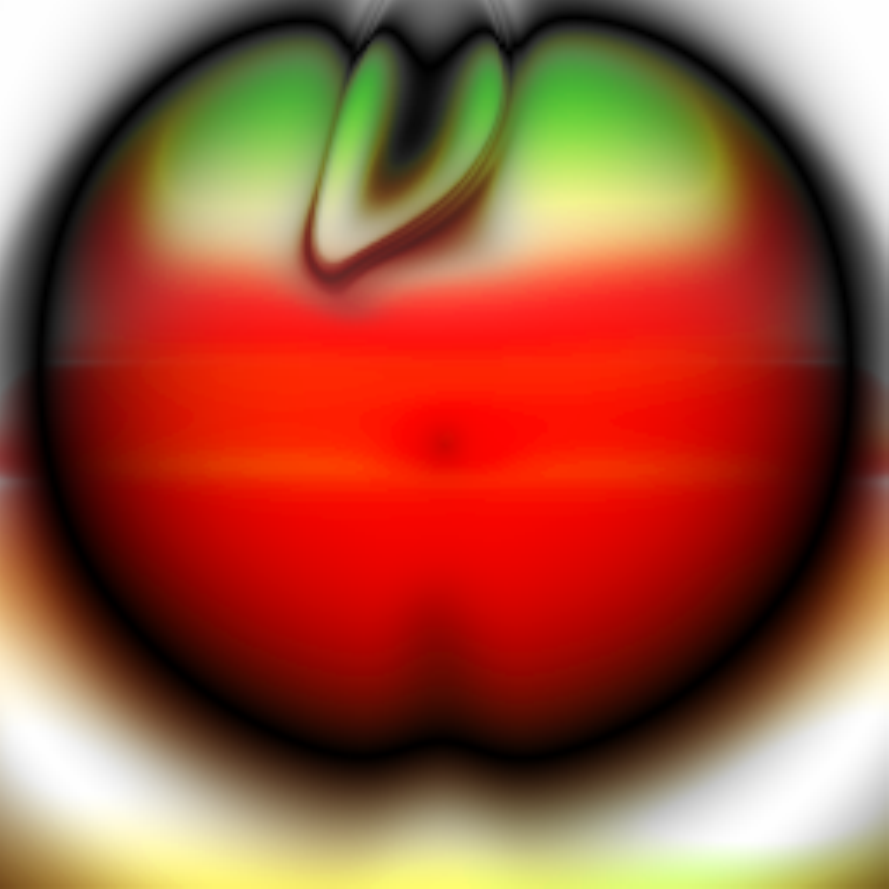}
\includegraphics[width=0.19\figwidth]{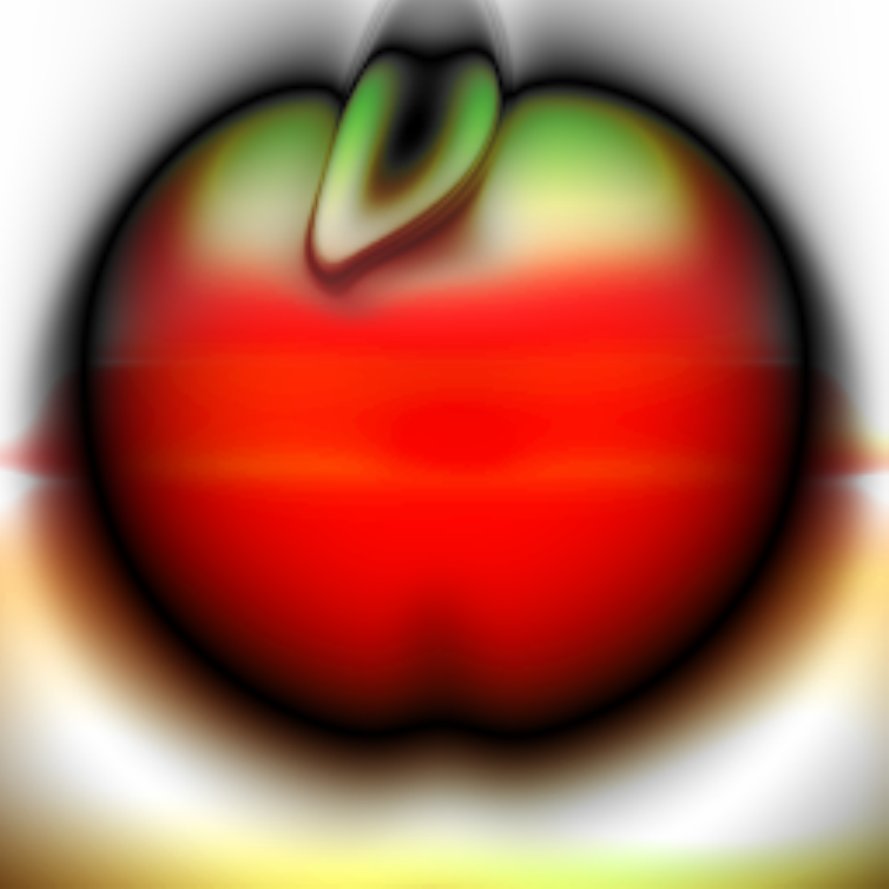}
\includegraphics[width=0.19\figwidth]{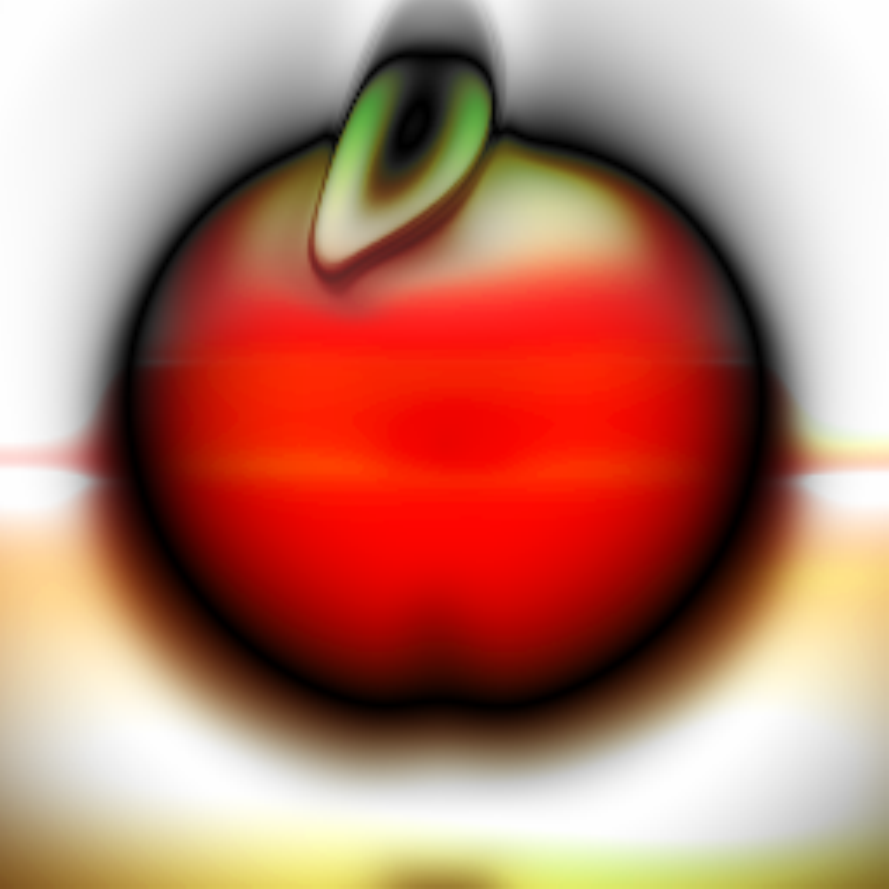}
\includegraphics[width=0.19\figwidth]{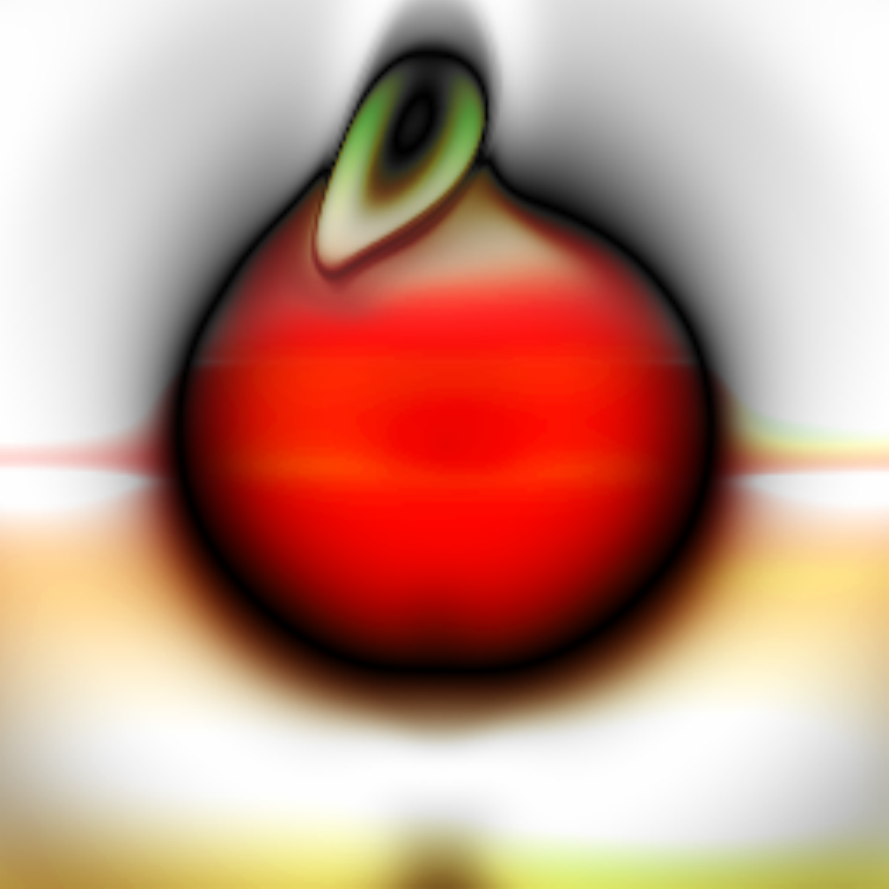}
\includegraphics[width=0.19\figwidth]{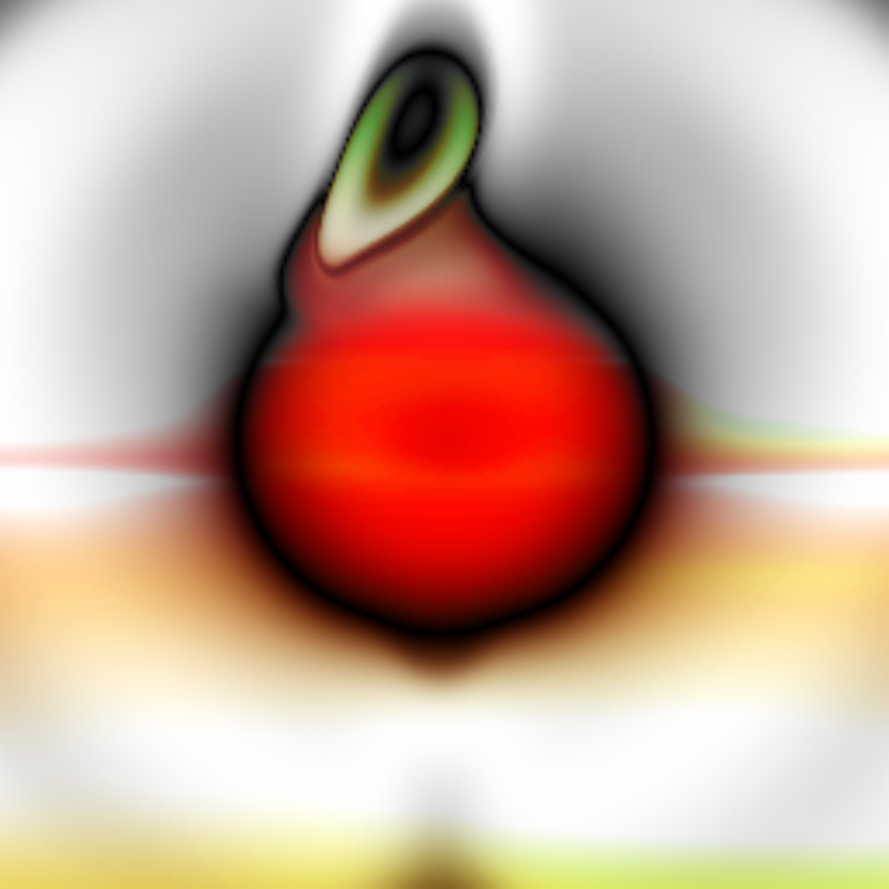}

Apple stem swing\\
\includegraphics[width=0.19\figwidth]{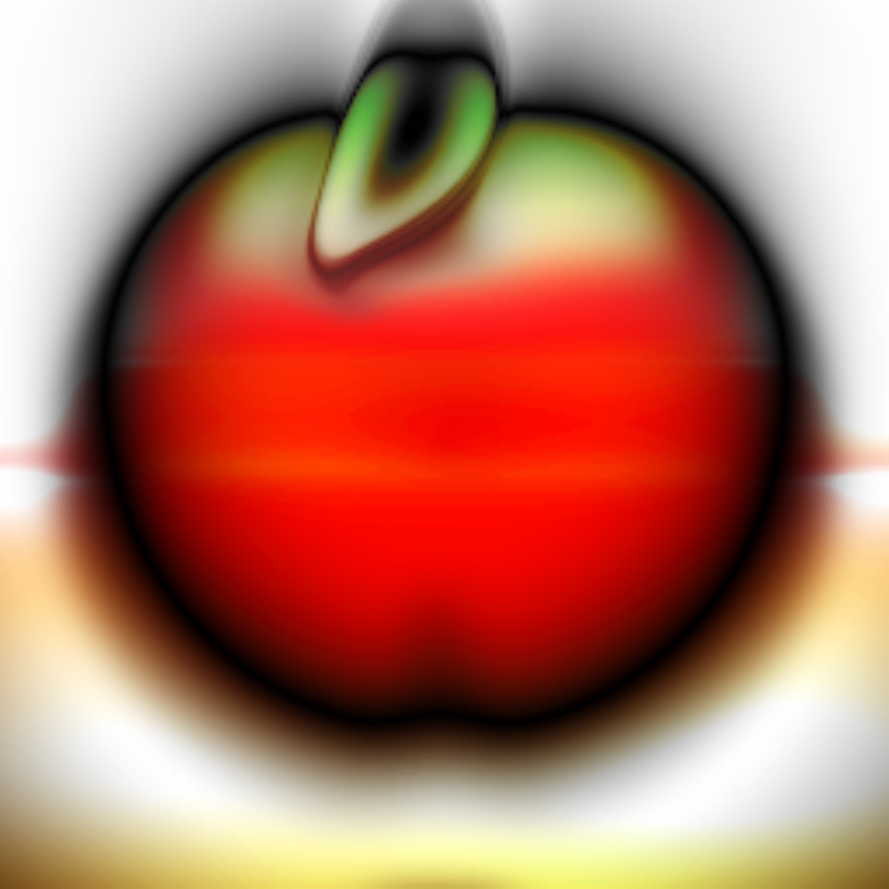}
\includegraphics[width=0.19\figwidth]{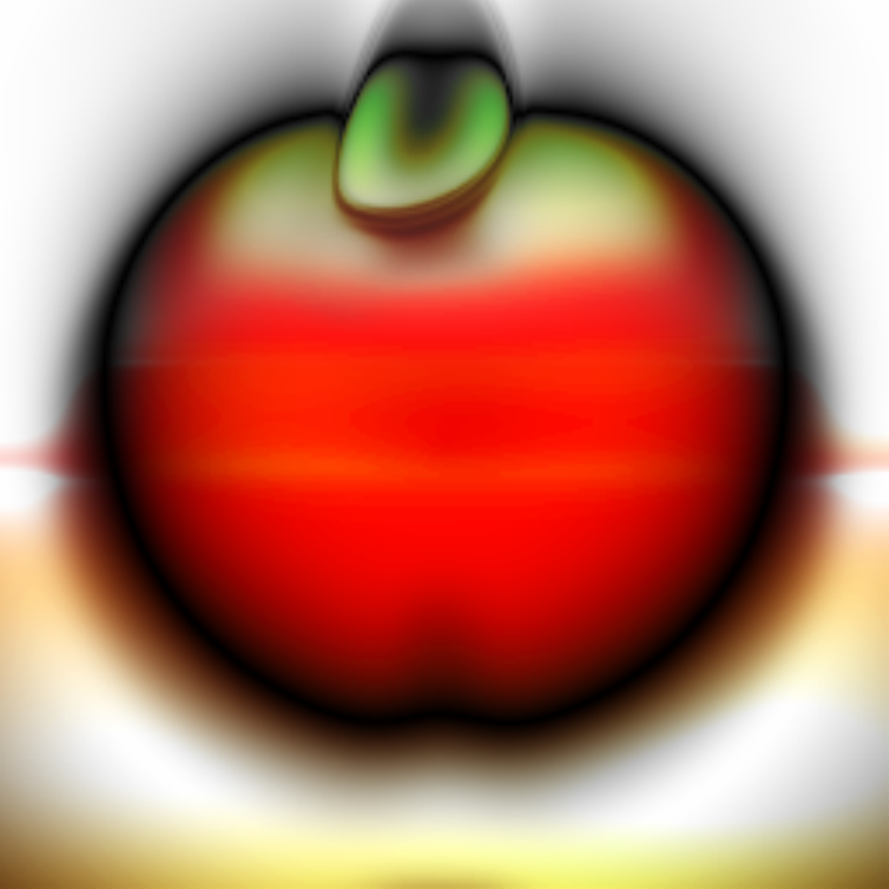}
\includegraphics[width=0.19\figwidth]{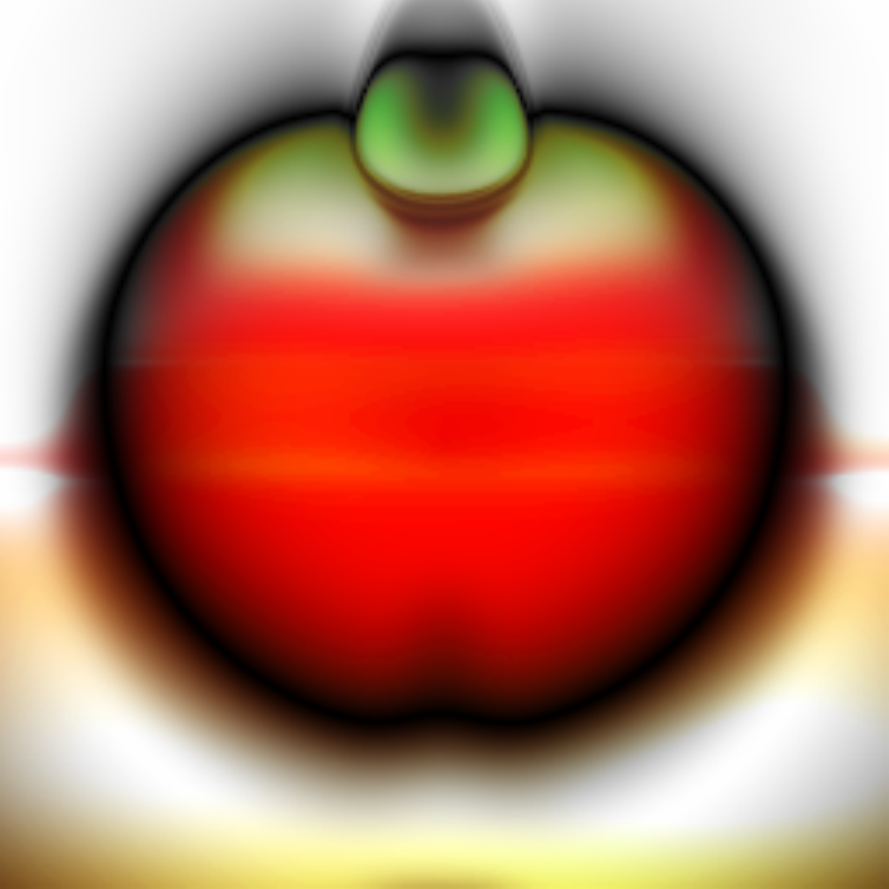}
\includegraphics[width=0.19\figwidth]{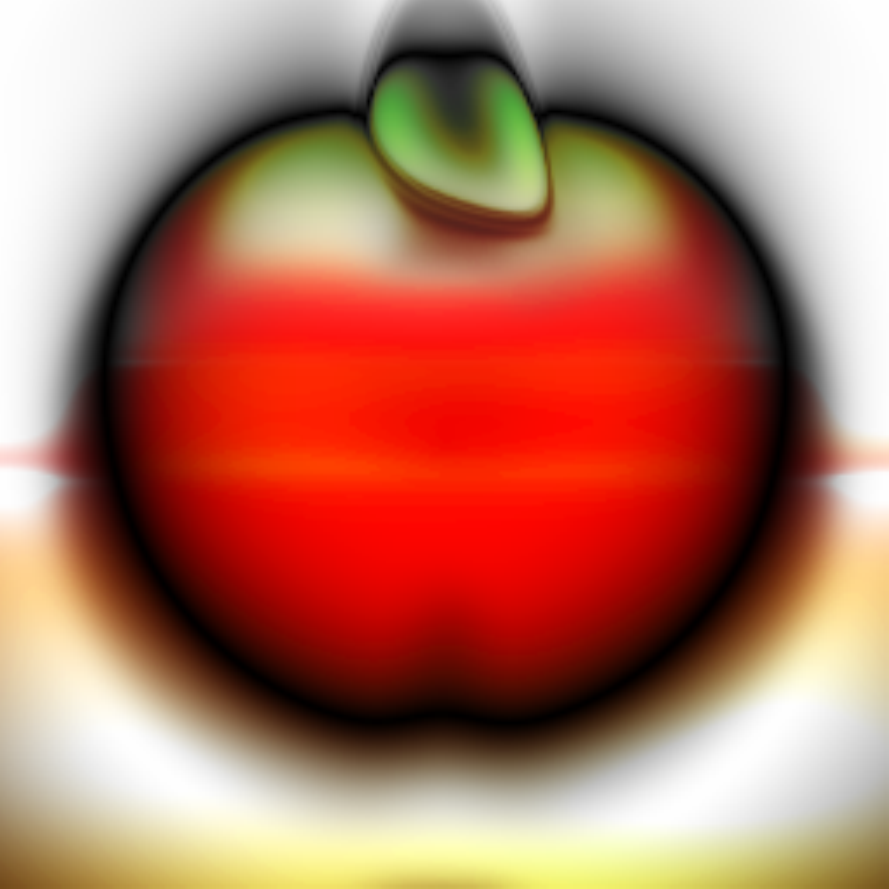}
\includegraphics[width=0.19\figwidth]{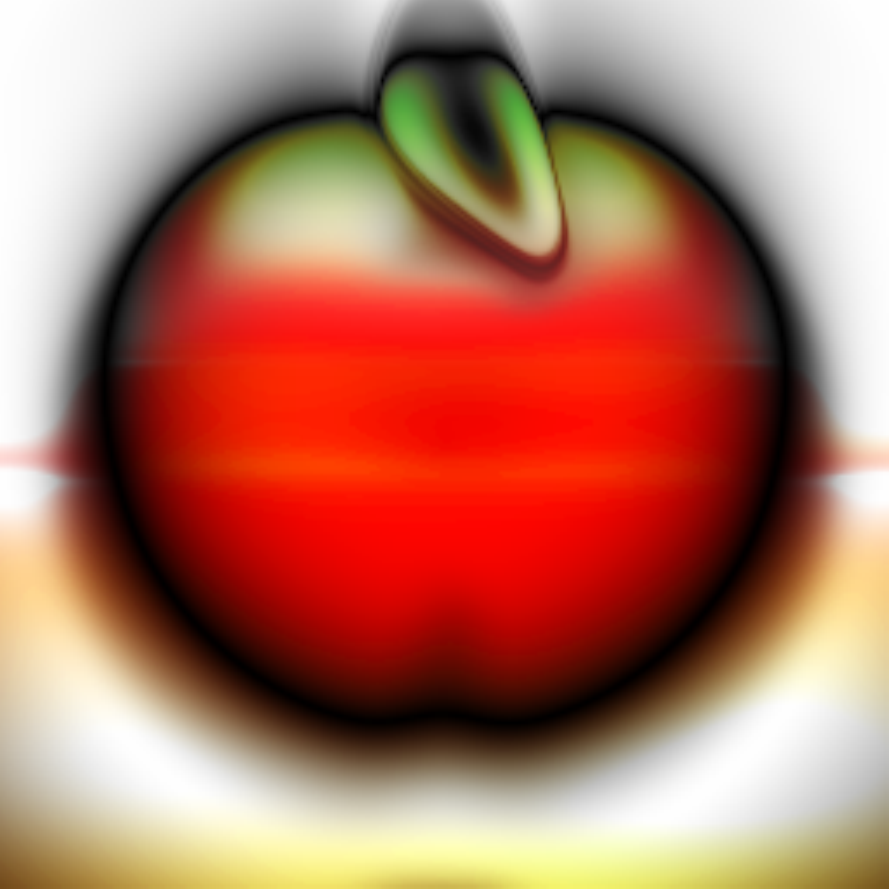}

Insect with shrinking antennae\\
\renewcommand{\dir}{Figures/DovBug}
\includegraphics[width=0.19\figwidth]{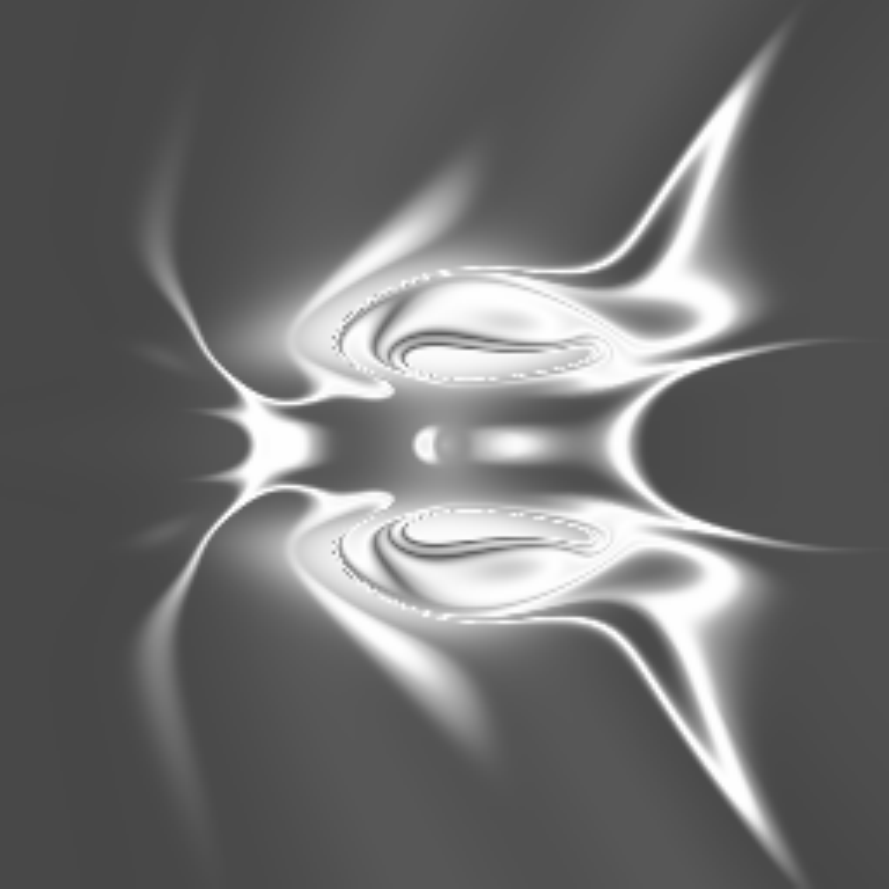}
\includegraphics[width=0.19\figwidth]{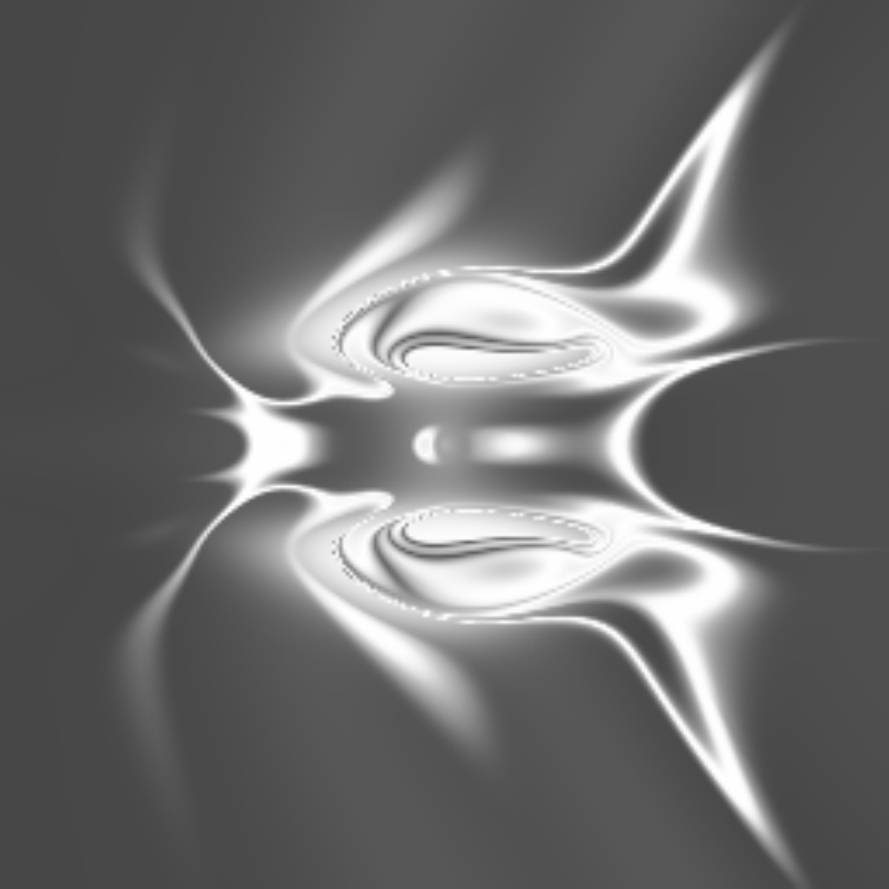}
\includegraphics[width=0.19\figwidth]{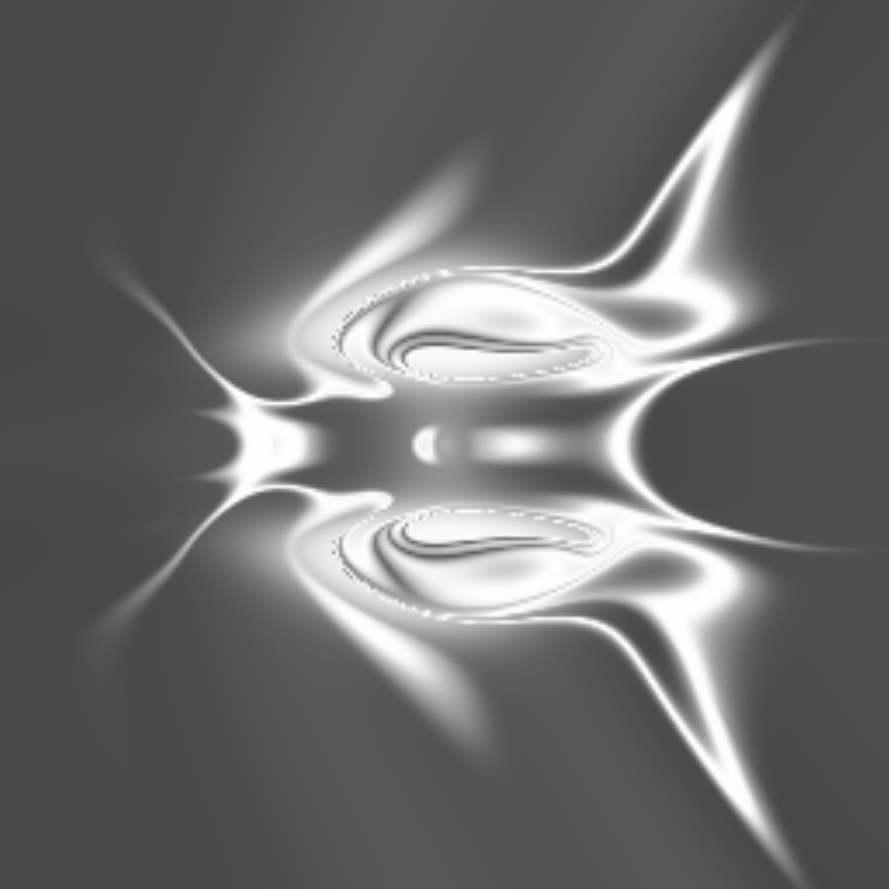}
\includegraphics[width=0.19\figwidth]{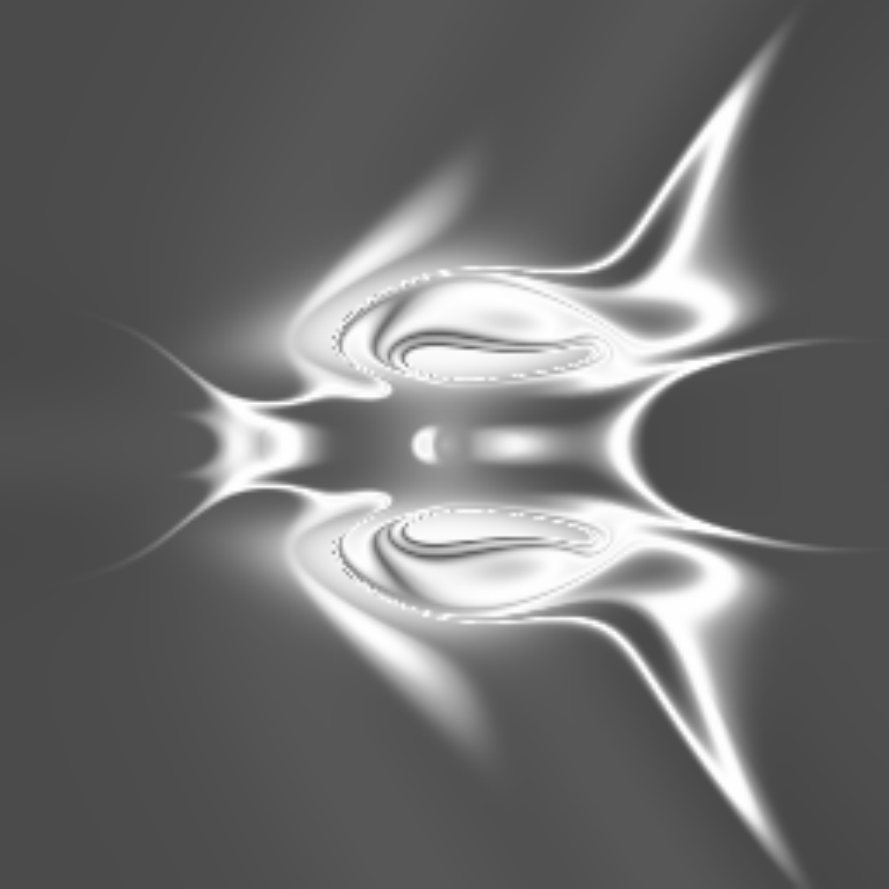}
\includegraphics[width=0.19\figwidth]{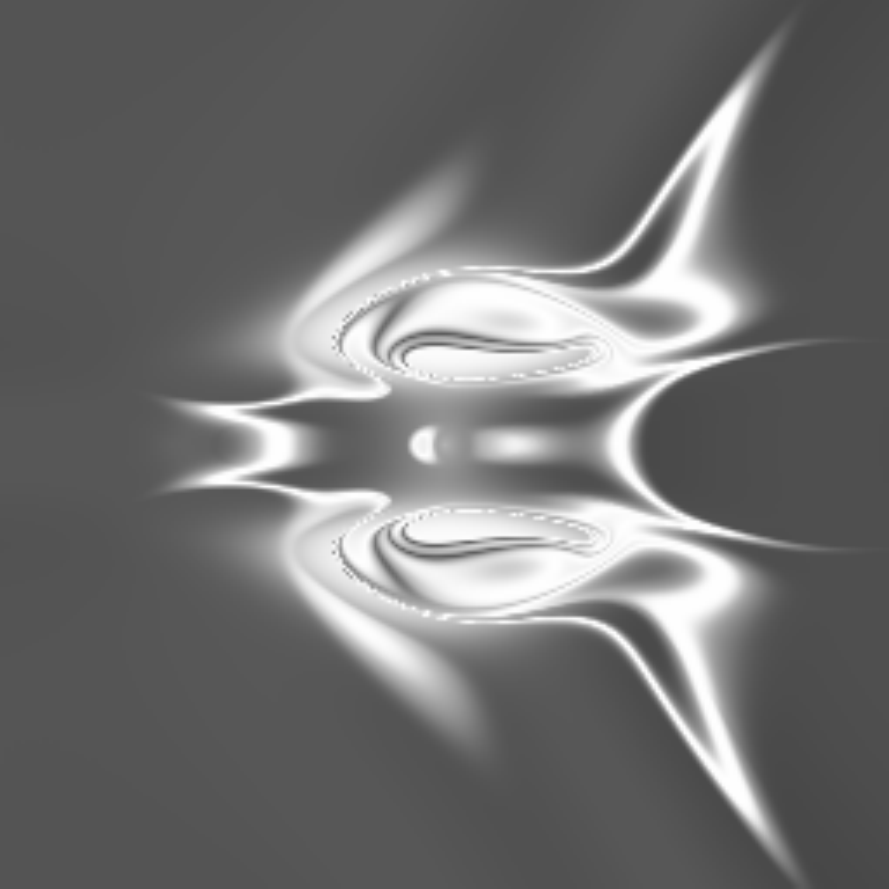}

Insect growing horns\\
\includegraphics[width=0.19\figwidth]{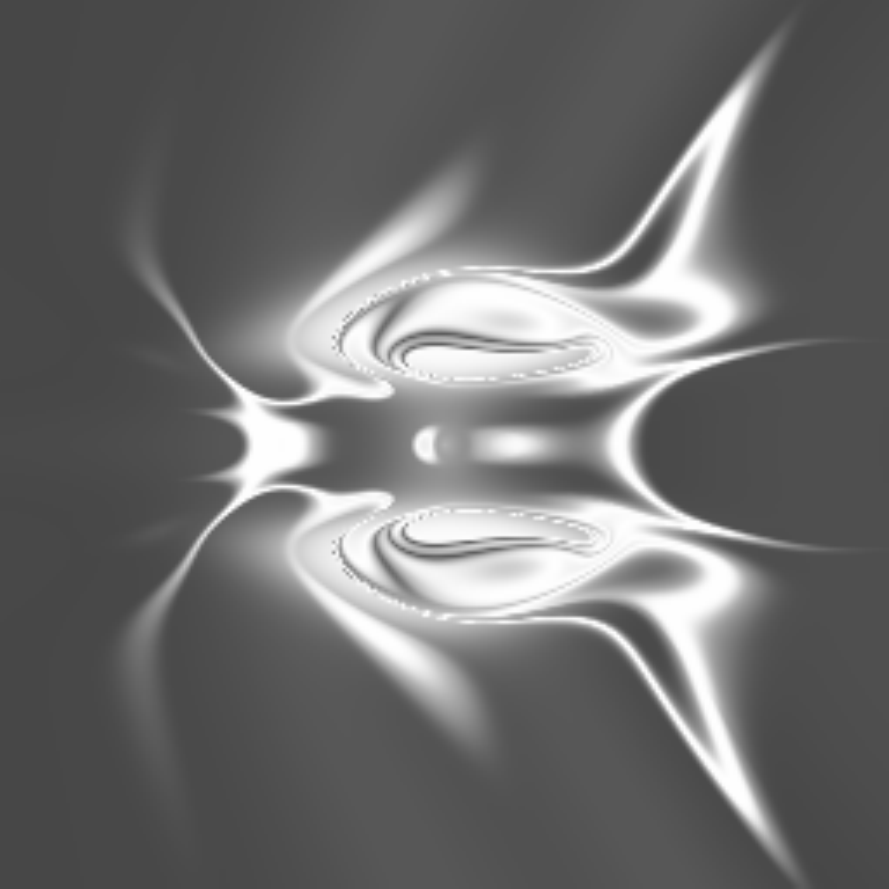}
\includegraphics[width=0.19\figwidth]{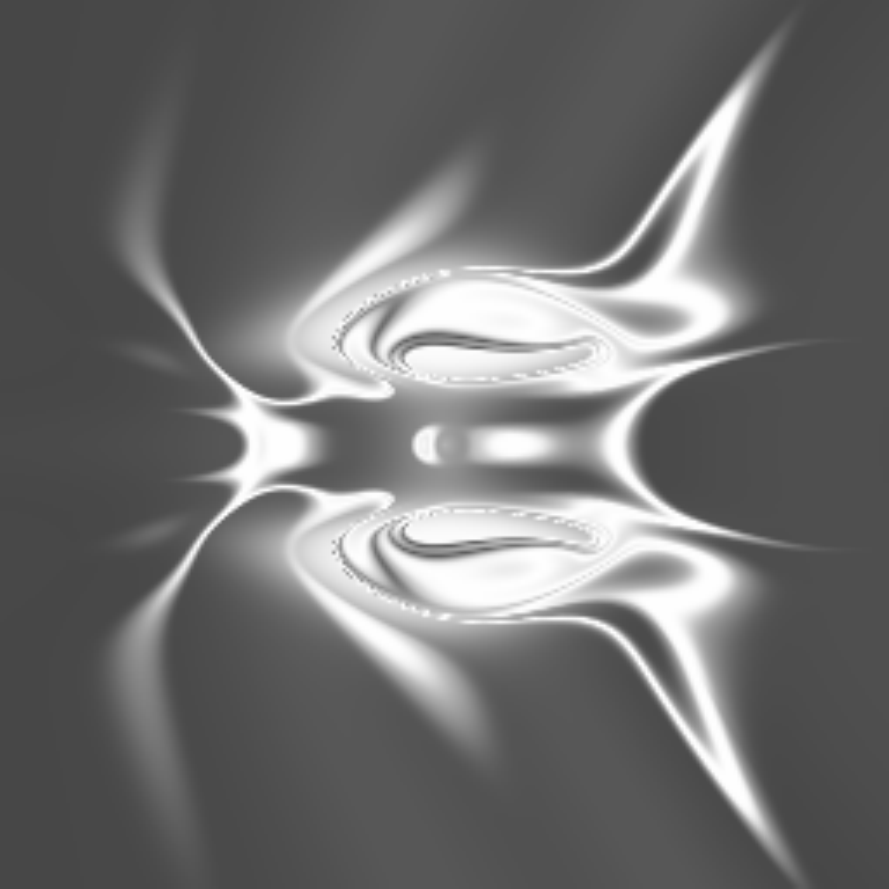}
\includegraphics[width=0.19\figwidth]{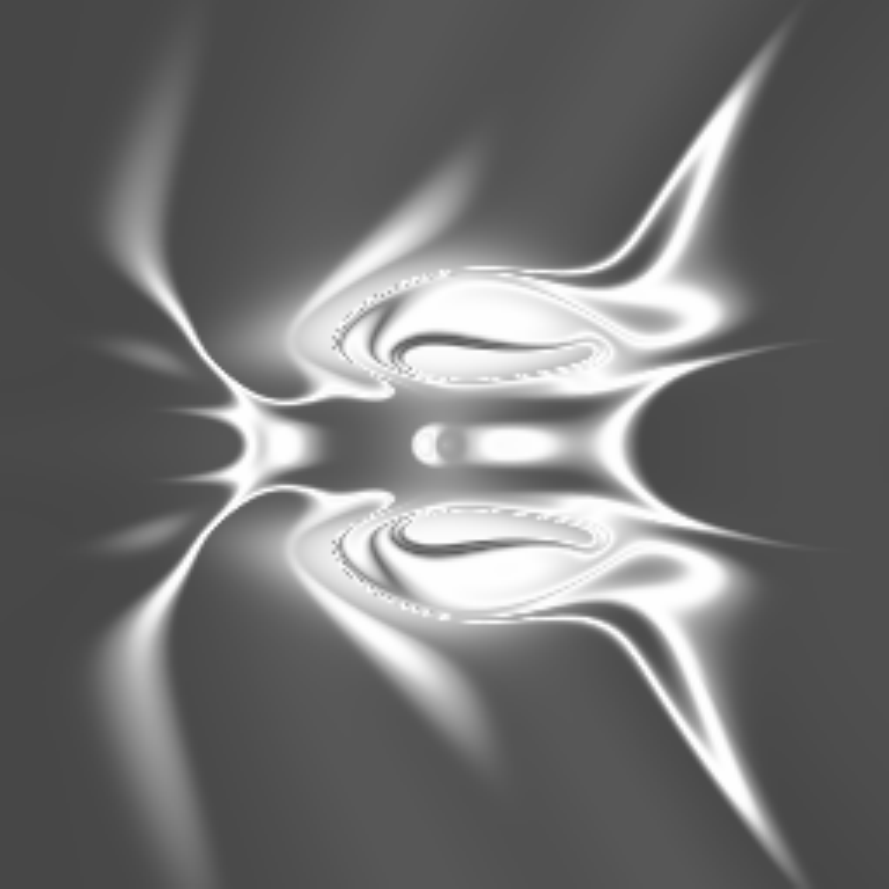}
\includegraphics[width=0.19\figwidth]{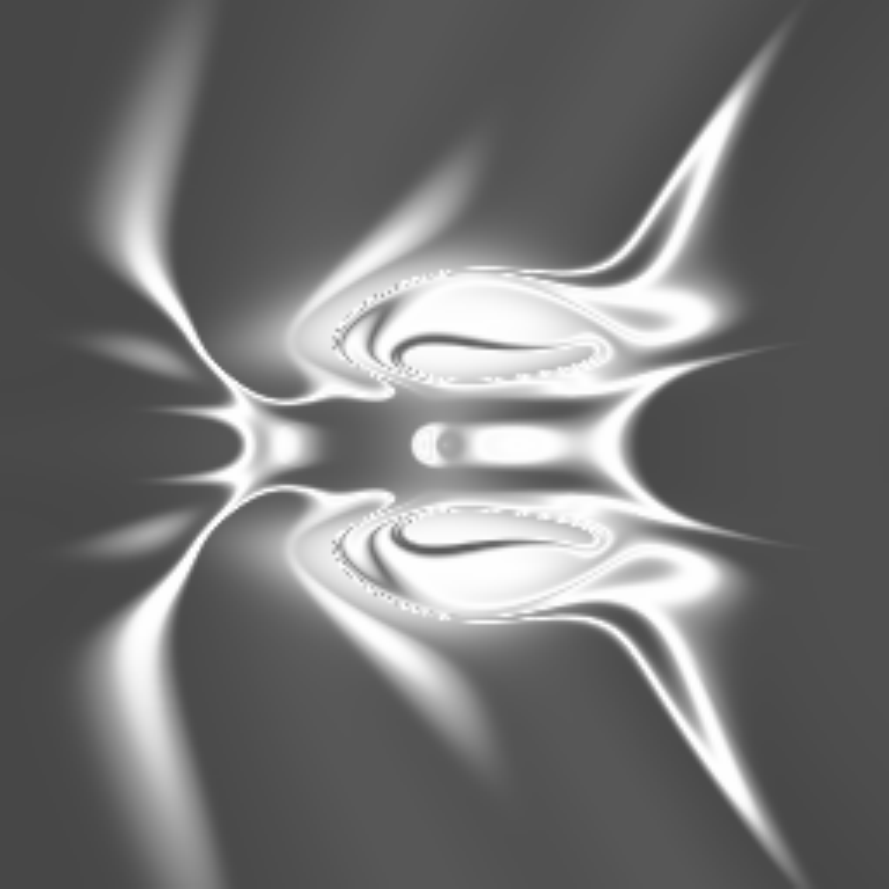}
\includegraphics[width=0.19\figwidth]{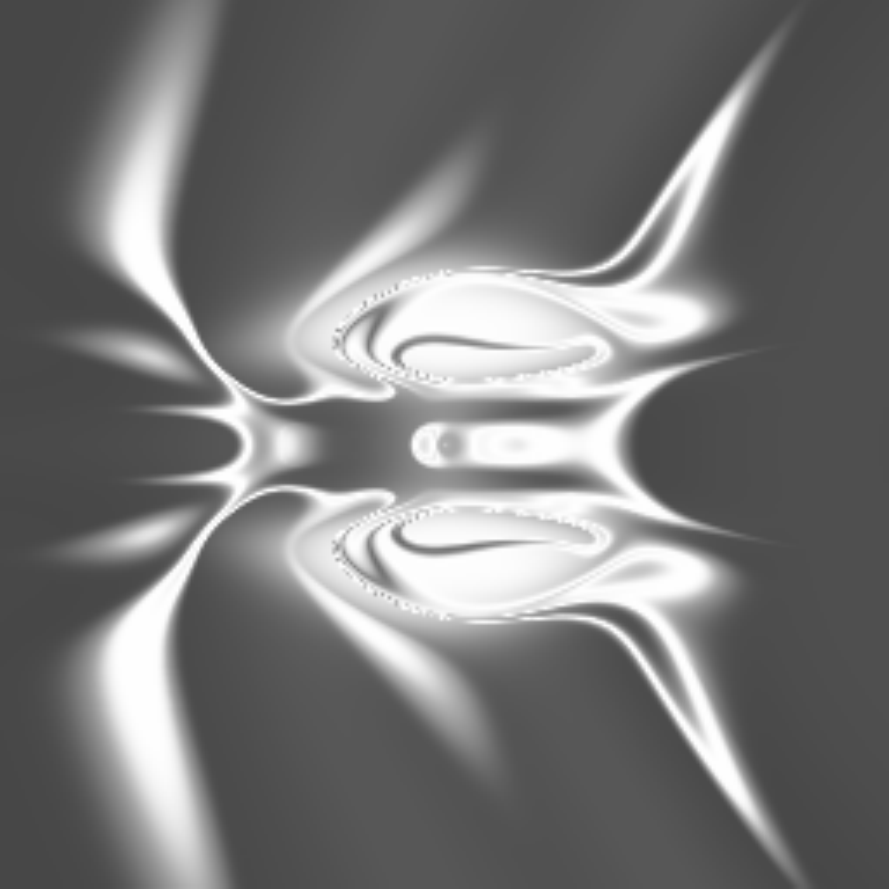}

Skull winking with left and right eye\\
\renewcommand{\dir}{Figures/DovSkull}
\includegraphics[width=0.19\figwidth]{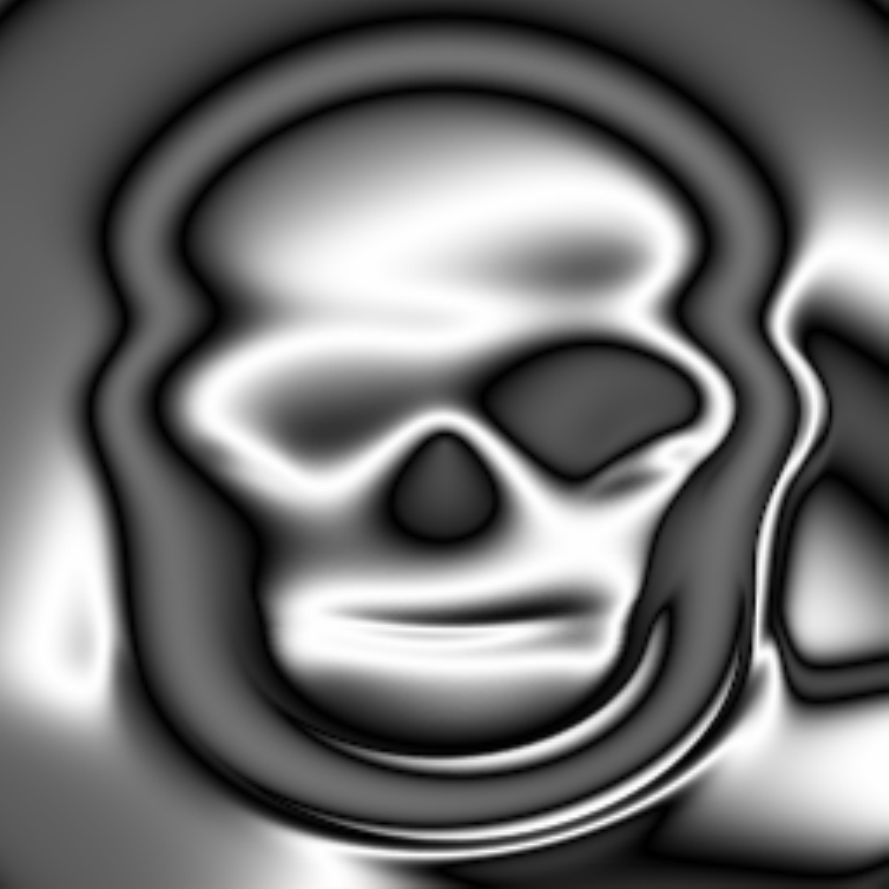}
\includegraphics[width=0.19\figwidth]{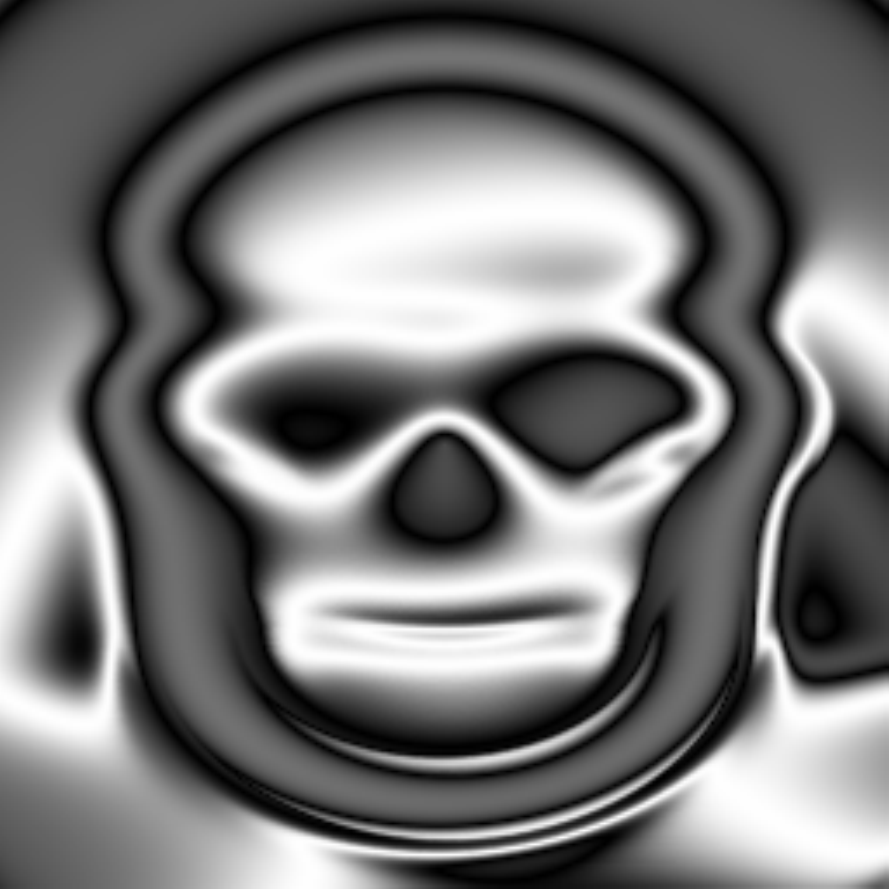}
\includegraphics[width=0.19\figwidth]{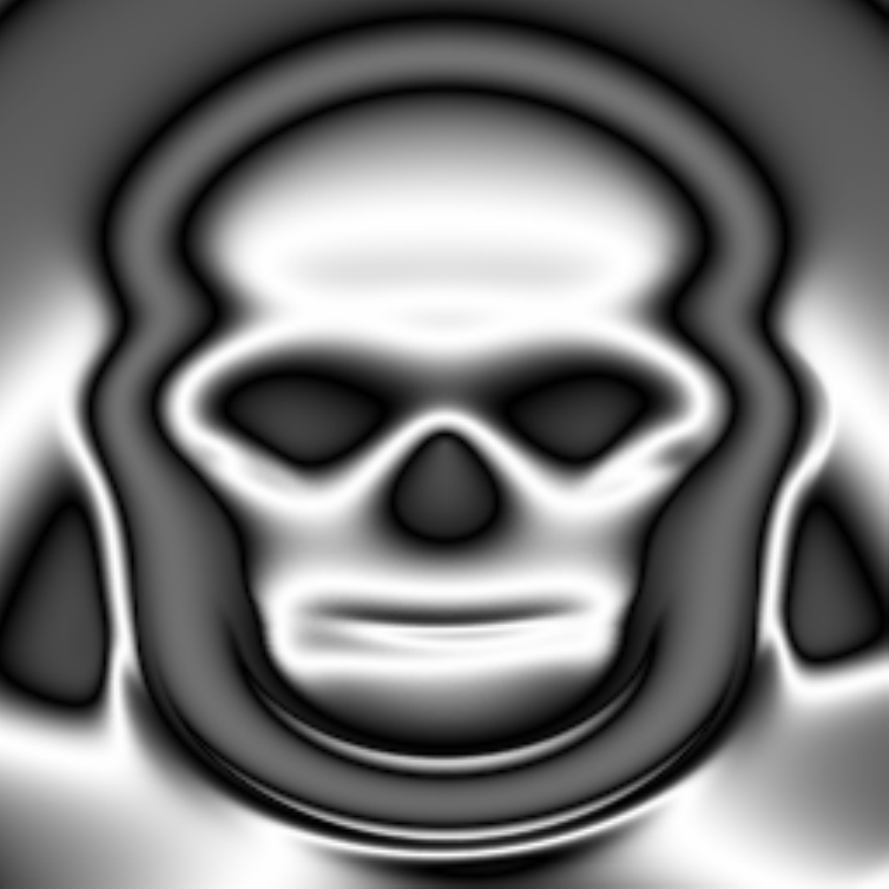}
\includegraphics[width=0.19\figwidth]{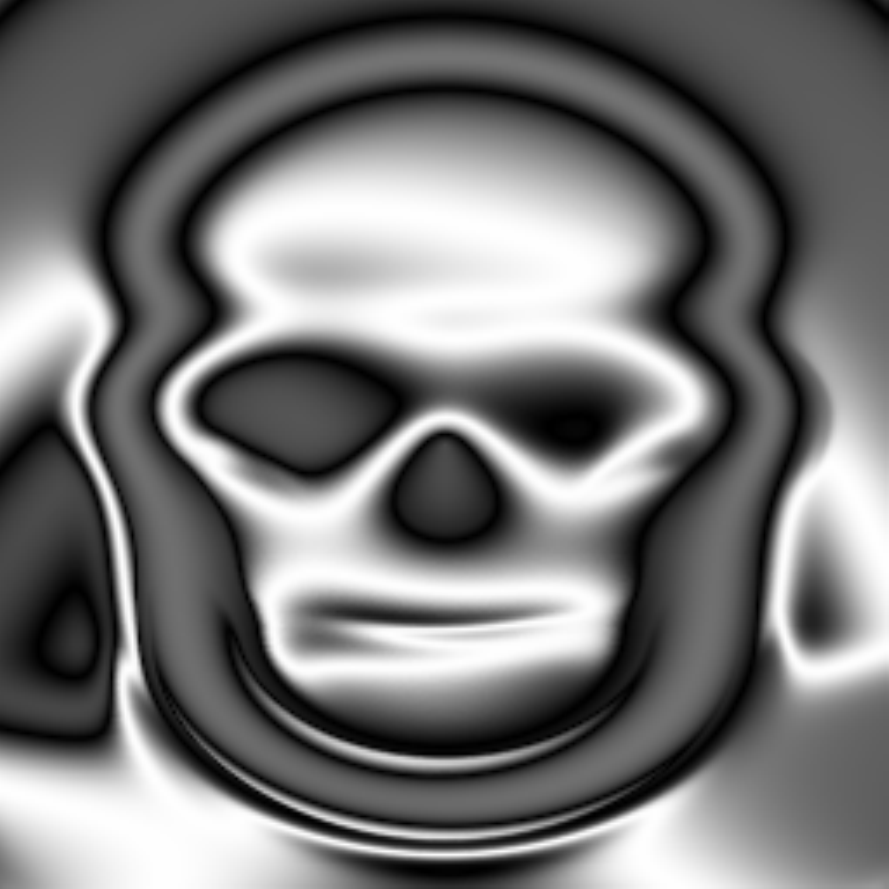}
\includegraphics[width=0.19\figwidth]{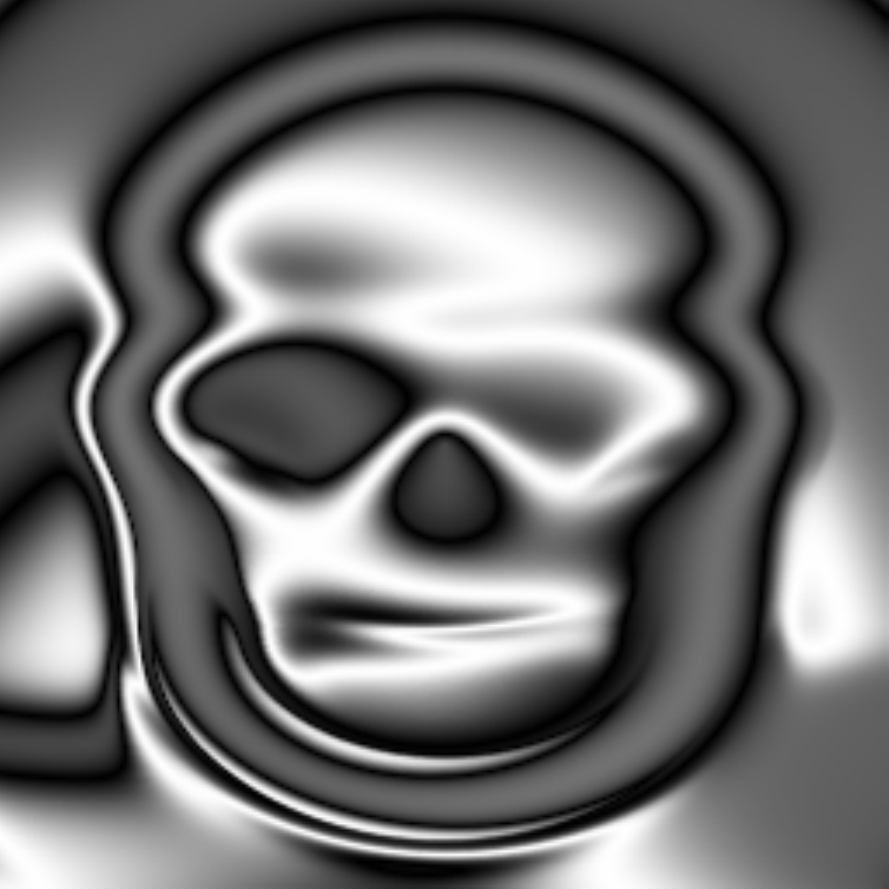}

Skull mouth open-close\\
\includegraphics[width=0.19\figwidth]{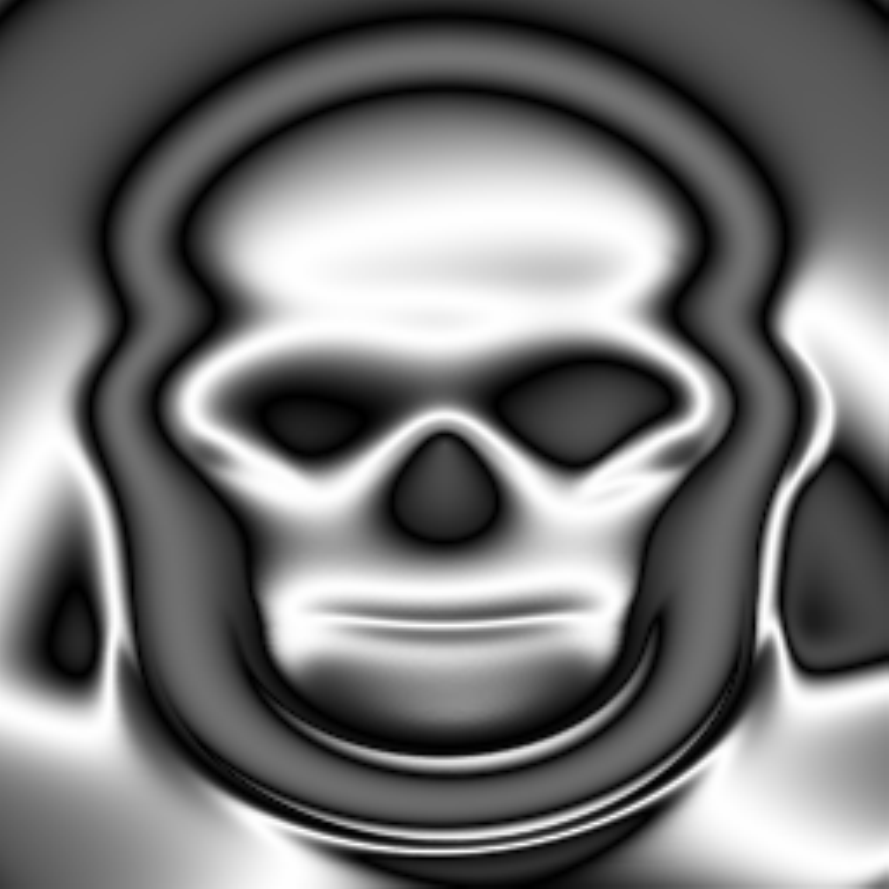}
\includegraphics[width=0.19\figwidth]{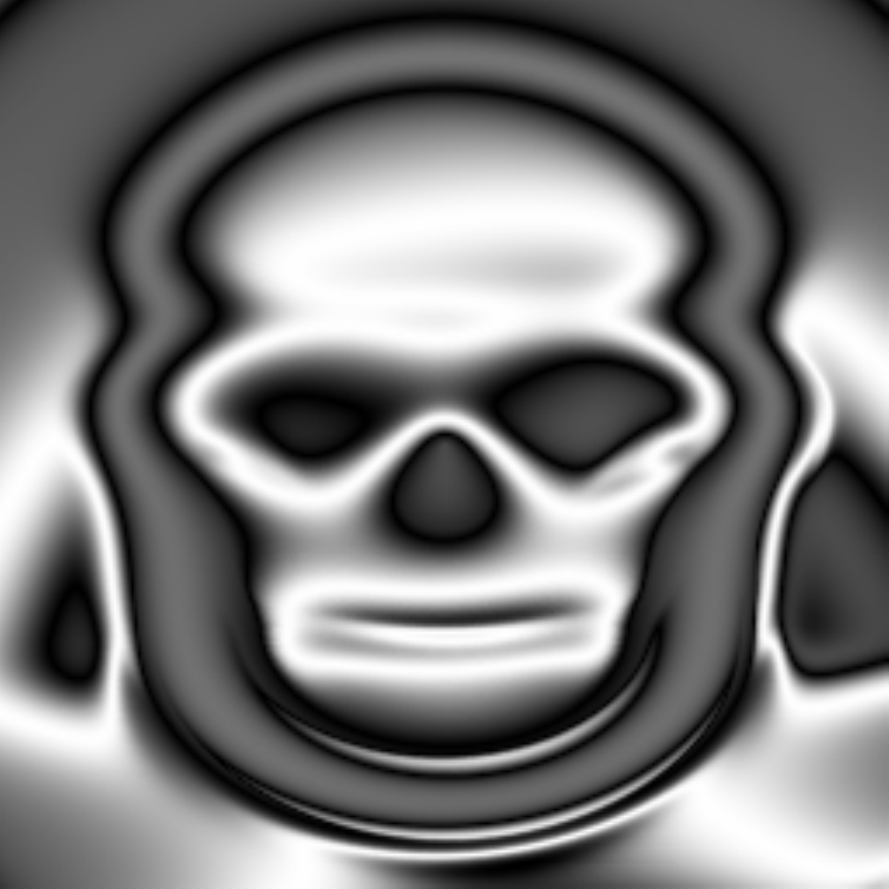}
\includegraphics[width=0.19\figwidth]{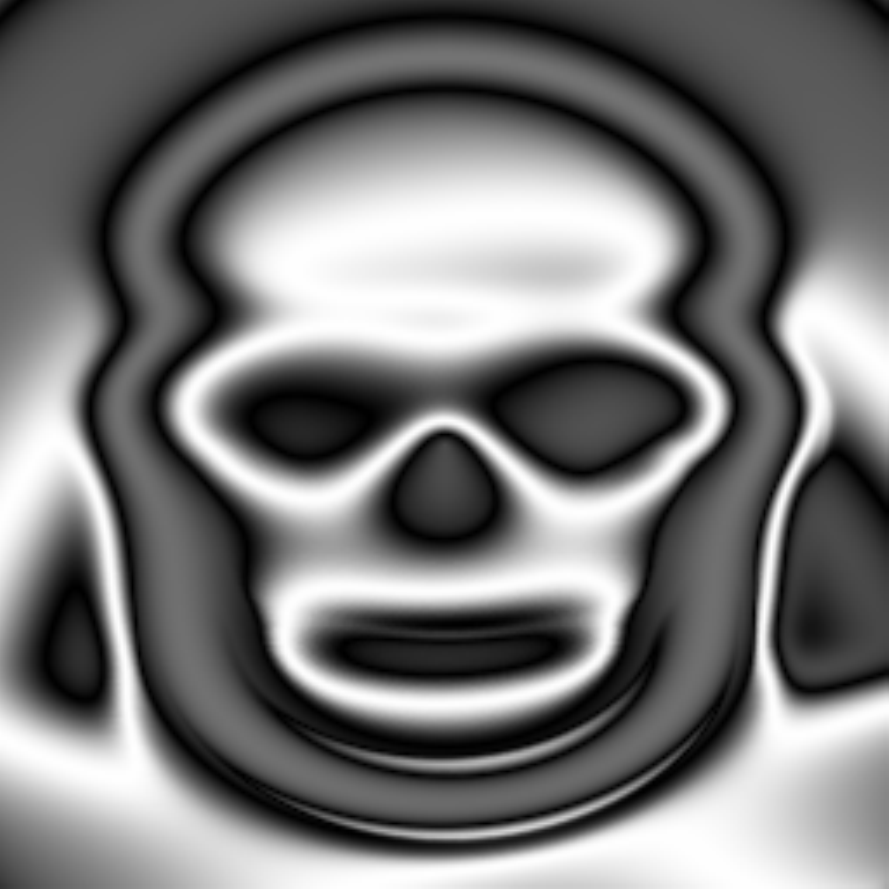}
\includegraphics[width=0.19\figwidth]{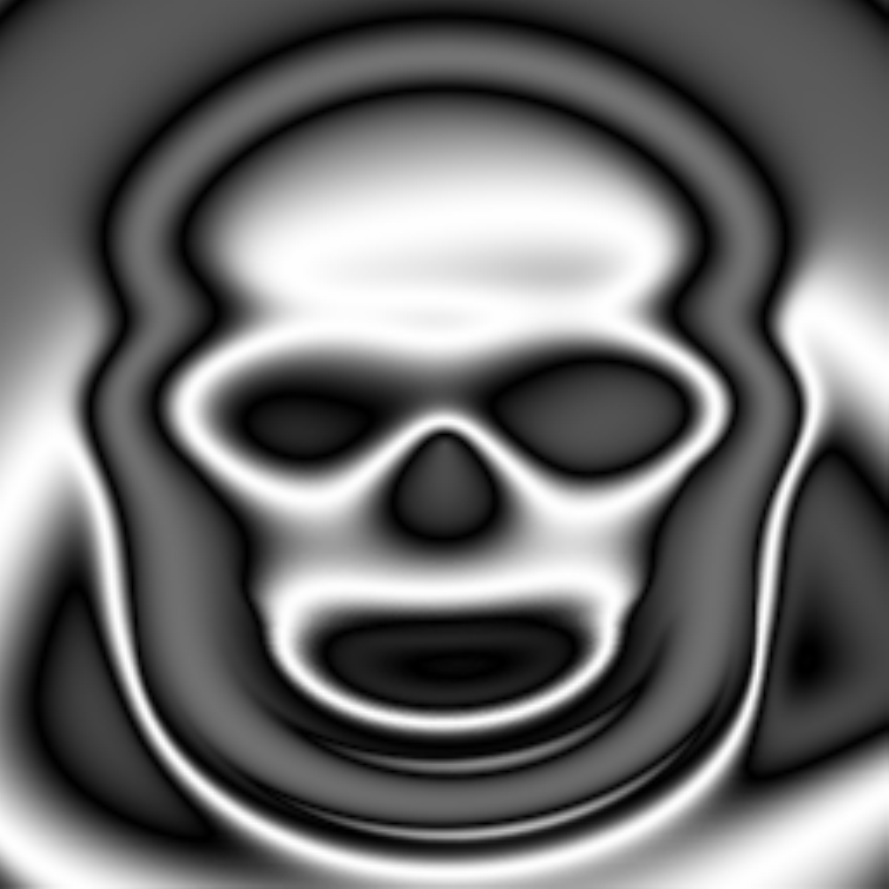}
\includegraphics[width=0.19\figwidth]{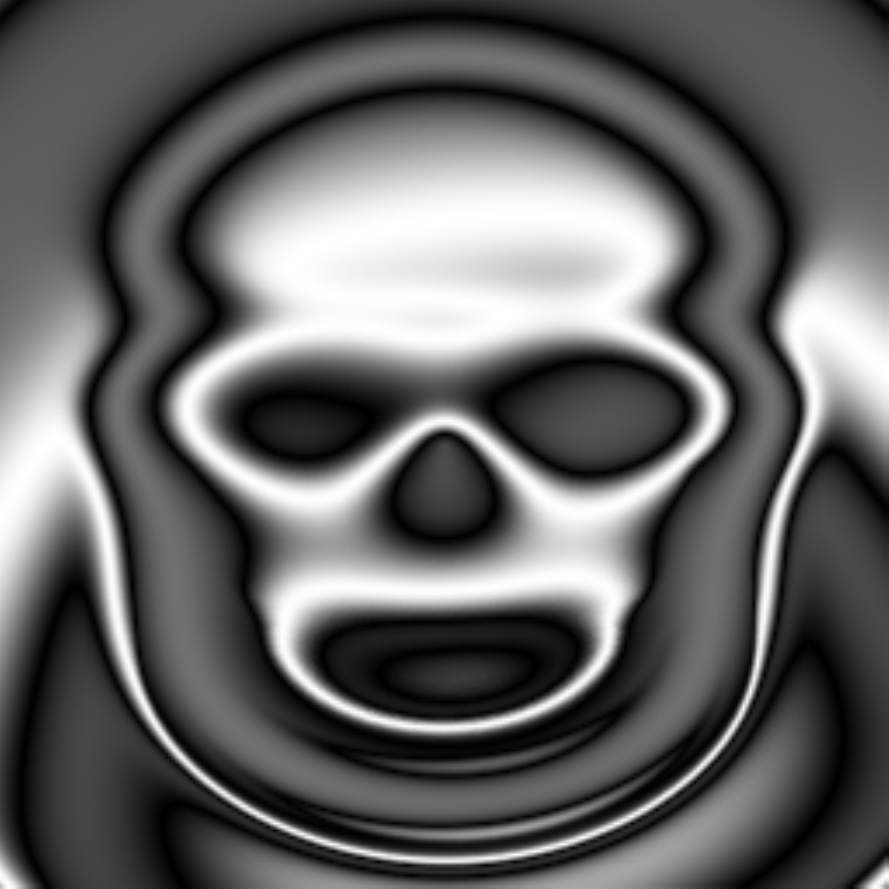}

\end{center}
\caption{\captionstart{}\textbf{Many images on Picbreeder have canalized dimensions of variation.} From left to right, each row represents the effect of sweeping over a single gene (connection) in the underlying CPPN genome, along with our subjective interpretation of the effect of changing that connection.}
\label{fig:#1}
\end{figure}
}
\figotherimages{otherImages}

To examine whether \picbreeder{} images canalized dimensions of variation, 
we selected one image that a user titled ``Spotlight Casting Shadow'' (Fig.~\ref{fig:limelight}). 
We selected it because it visually appears to contain a clearly distinct object in the image (the object), 
a correlated attribute (the shadow), and an independent, but also conceptually distinct entity (the spotlight). 
We thus wondered how these entities would respond to changes in the genome.
One possibility was that this image would behave like a face seen in clouds. 
To us, human observers, such a shape may appear to consist of various different entities,
such as eyes, a nose, and a mouth.
However, as the clouds change shape in the wind, one would not expect any of these components to be preserved. 
For example, it is exceedingly unlikely for the expression on the face to cycle through different expressions, 
or the eyes to open and shut, or the entire face to expand appropriately, etc. 
Instead, most often shapes seen in clouds are ephemeral, 
and quickly morph back into an amorphous cloud (or perhaps an entirely different shape), 
without any regard for the meaning once assigned to the shape and its parts.
The same could have been true for this Picbreeder image,
where the relationships between the different entities within the image would be solely within the eyes of the beholder,
and where changes to the genotype would simply cause the image to become scrambled in unrecognizable ways.
However, as described below,
we discovered that the genome not only evolved to enable the different 
aspects of the image to be independently controlled while preserving their meaning, 
but that the dimensions of variation for these objects are sensible in that they 
enable changes to the image in a way that humans might expect the objects to be manipulated.

To test whether such conclusions extended to other images, we then
analyzed the 12 most branched images from \picbreeder{} 
(recall that being branched can be considered a form of fitness in this system). 
We specifically tested whether their CPPN genomes contained links that affected a single, 
qualitative aspect of the image. 
To make this determination, we annotated the genome as described in Sec.~\ref{sec:examiner}.
Images of all 13 fully labeled networks, including representative examples of variation, 
can be found in SI (SI Sec.~\ref{SI-sec:si_analyzed_genomes}).

Canalizations of dimensions of variation were found in every \picbreeder{} image we examined. 
While the images differed in the quality and quantity of canalizations, 
even the images that seemingly consist of arbitrary patterns have canalized some interesting dimensions of variation. 
Two example dimensions of variation for three different images are shown in Fig.~\ref{fig:otherImages}.
Full videos of these and other transitions are available at: \url{www.evolvingai.org/PicbreederCanalization}.

We picked the Spotlight Casting Shadow image to present in detail in this paper (Fig.~\ref{fig:limelight}), 
though most other images we analyzed have 
qualitatively similar properties (SI Sec.~\ref{SI-sec:si_analyzed_genomes}).
While the image itself appears to have separate components (the object, its shadow, and the spotlight), 
it could have been the case that genetically these features were not decomposed and could not be altered independently.
Surprisingly, however, the CPPN genome does contain individual 
connections specialized to modify only one of these three different entities. 
We found dimensions of variation corresponding to the size of the object, the size of the shadow, 
and the size and position of the spotlight (Fig.~\ref{fig:limelight}). 
Moreover, we also found connections that change multiple entities in a coordinated fashion; 
the object and the shadow can be modified together such that 
both objects can grow or shrink simultaneously, 
which is the behavior one would expect if the shadow was actually cast by the object.
Note that such canalization is not an inherent, inevitable property of CPPNs:
the image titled ``Dolphin'' (Fig.~\ref{fig:concept} right) features several visually distinct components, such as
the eye of the dolphin, the snout of the dolphin, the head of the dolphin, and the water in the background,
but most genes in the Dolphin CPPN are highly pleiotropic, and affect all of those components simultaneously.

\newcommand{\figlimelight}[1]{%
\renewcommand{\dir}{Figures/DovObjectCastingShadow}
\begin{figure}
\begin{center}
Components of the Spotlight Casting Shadow image.\\
\includegraphics[height=0.15\figwidth]{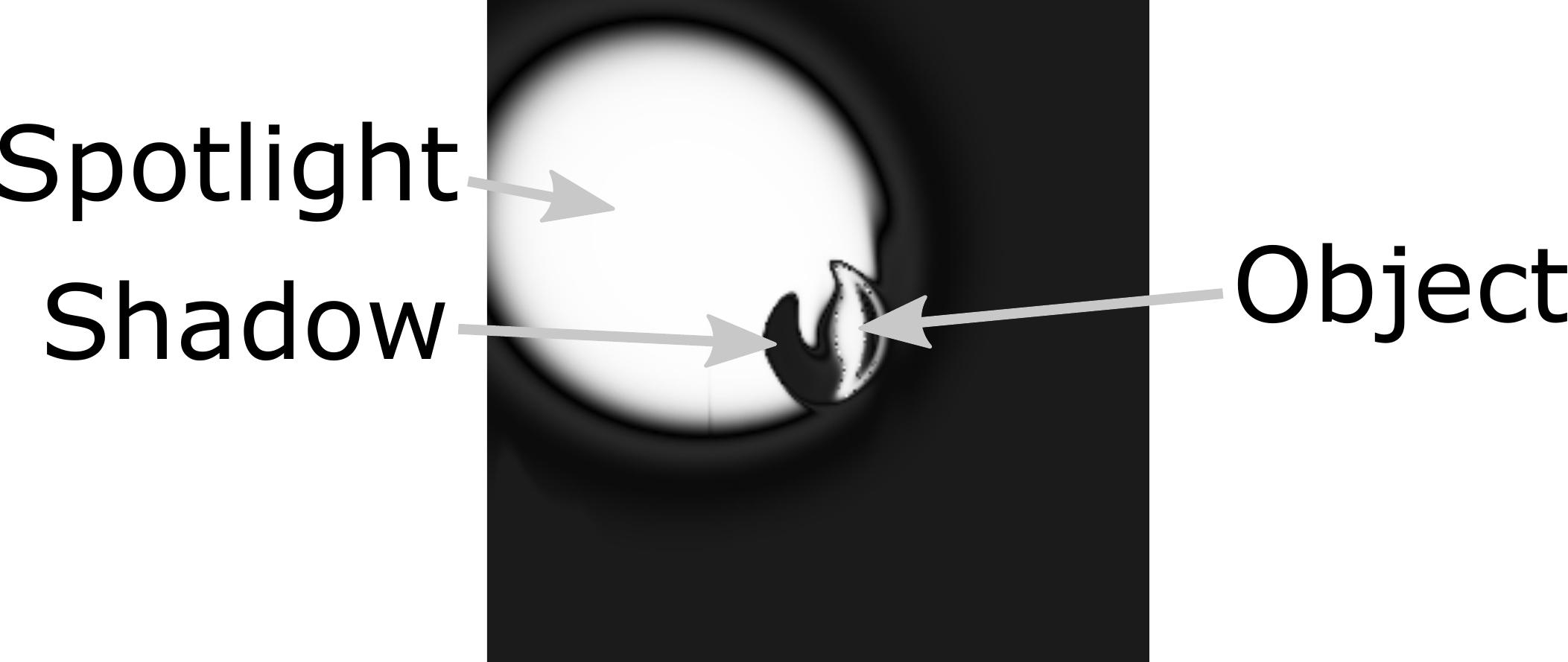}

Object and shadow size together (Fig~\ref{fig:networkTrans}, Link 1)\\
\includegraphics[width=0.15\figwidth]{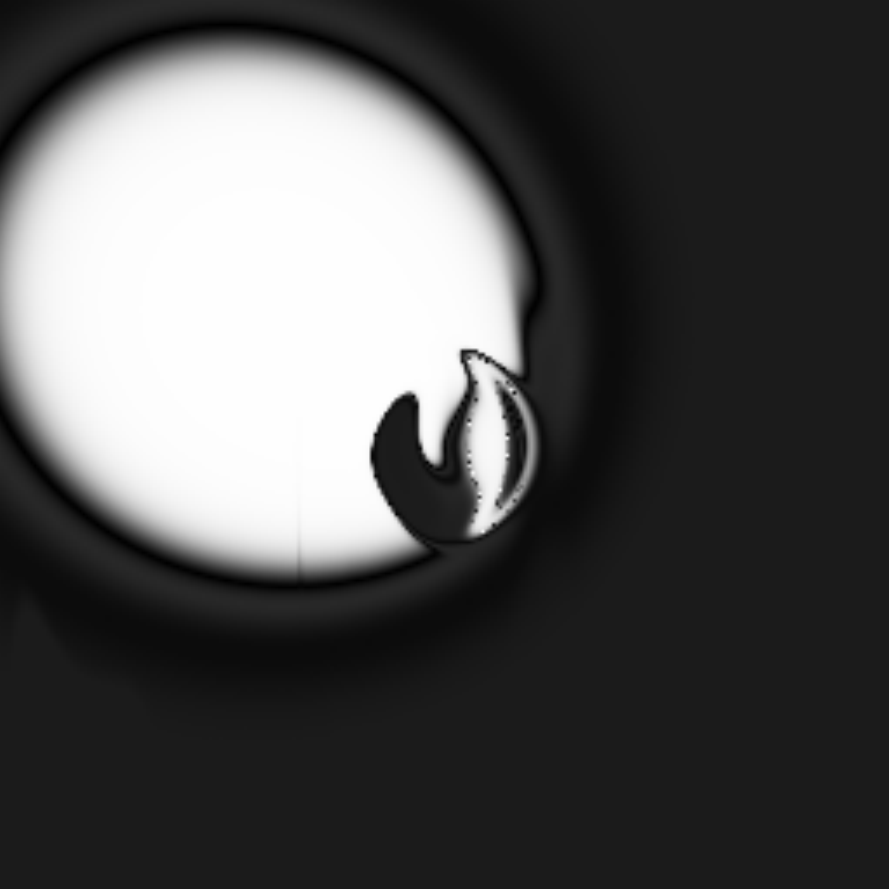}
\includegraphics[width=0.15\figwidth]{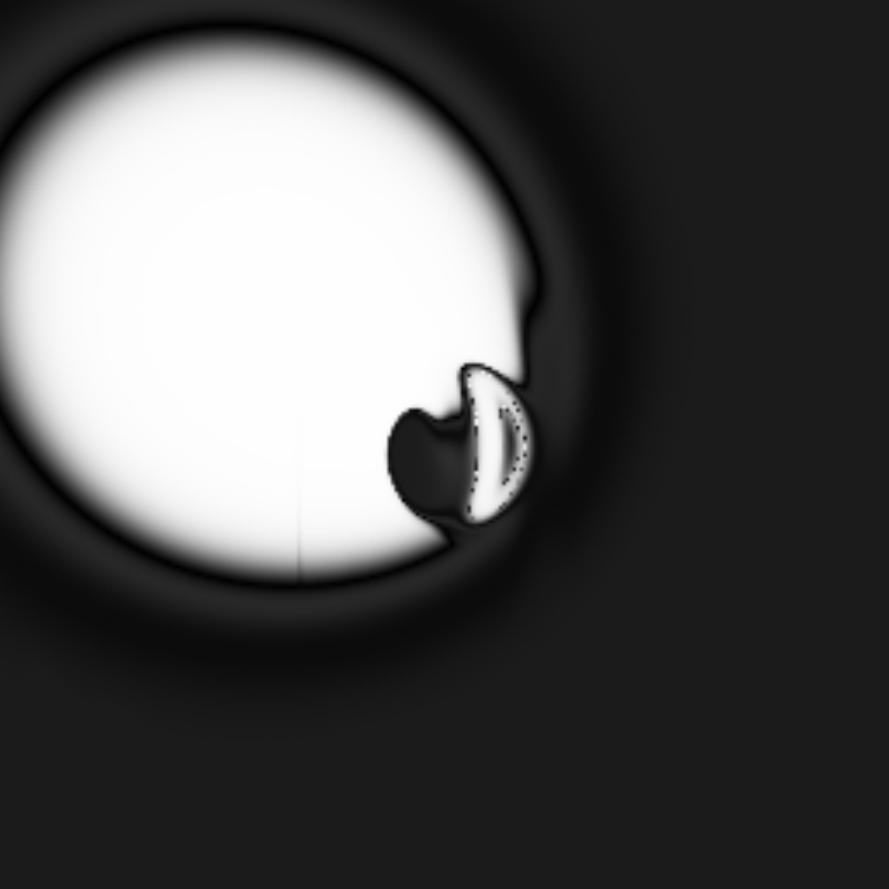}
\includegraphics[width=0.15\figwidth]{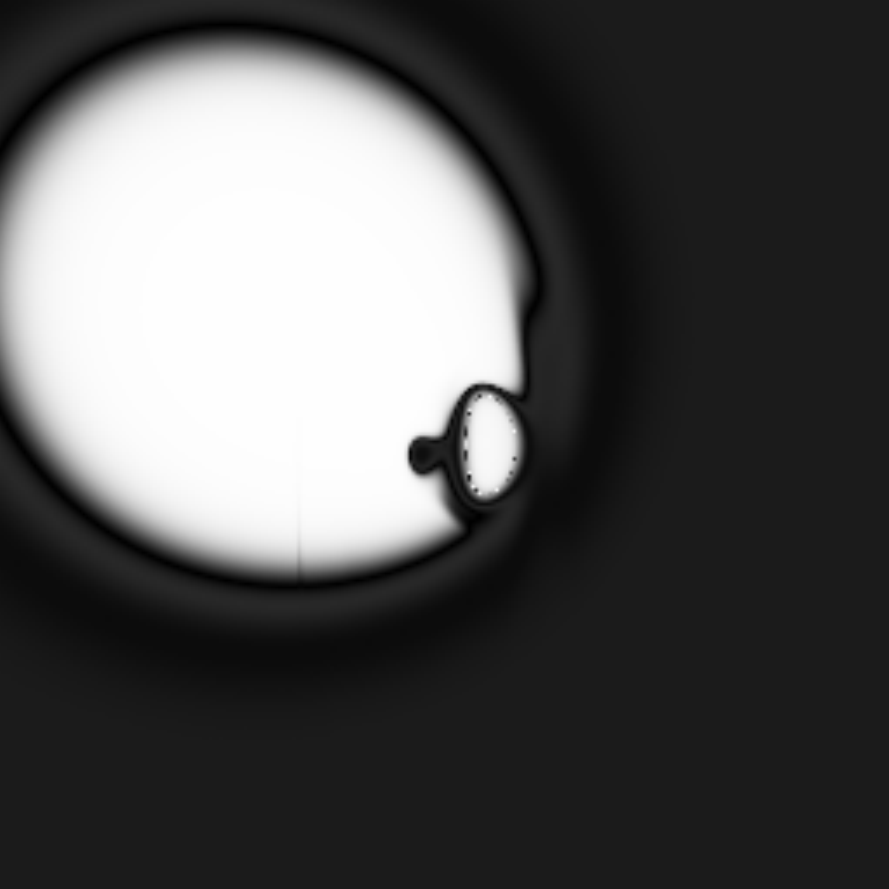}
\includegraphics[width=0.15\figwidth]{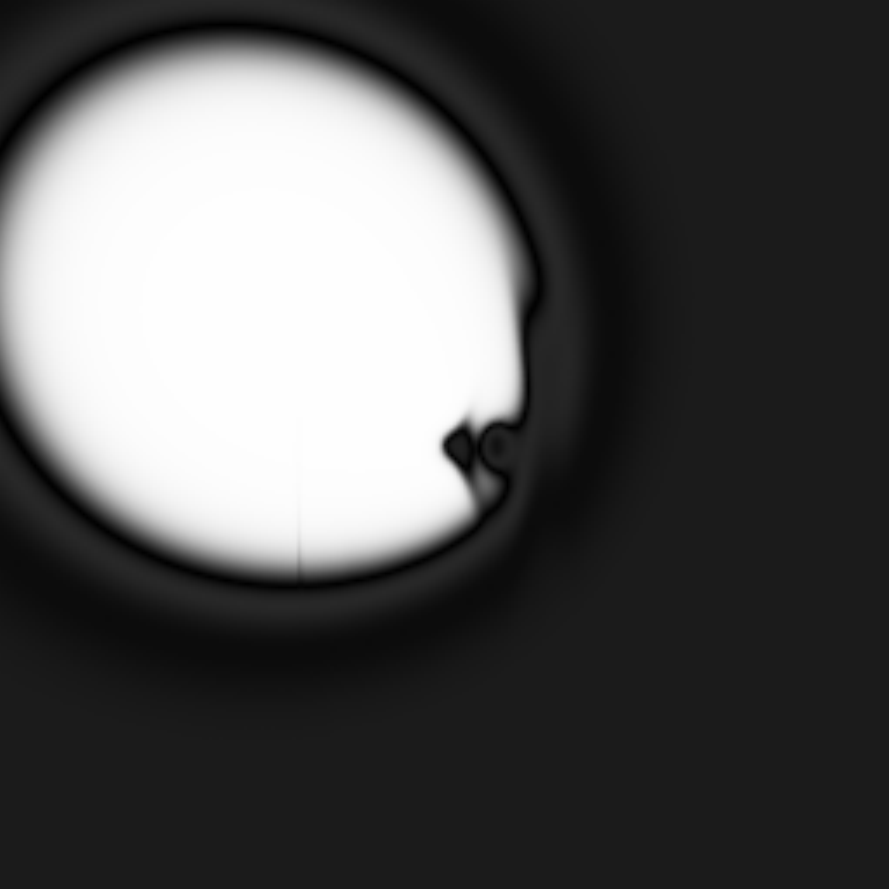}
\includegraphics[width=0.15\figwidth]{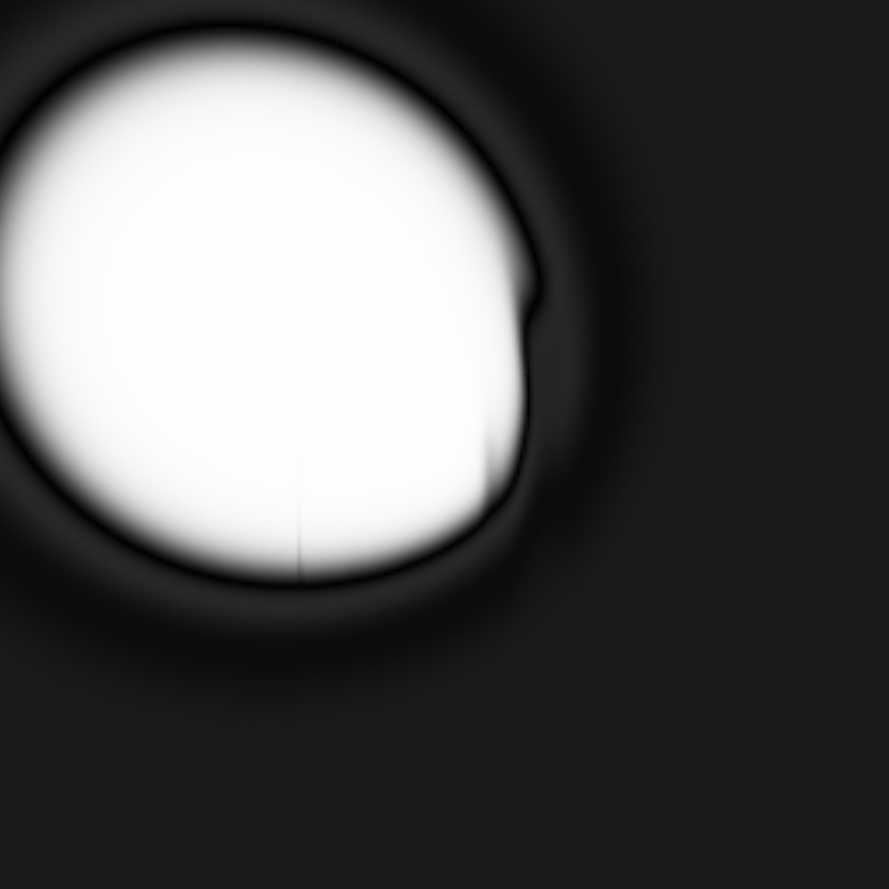}

Shadow size without object (Fig~\ref{fig:networkTrans}, Link 2)\\
\includegraphics[width=0.15\figwidth]{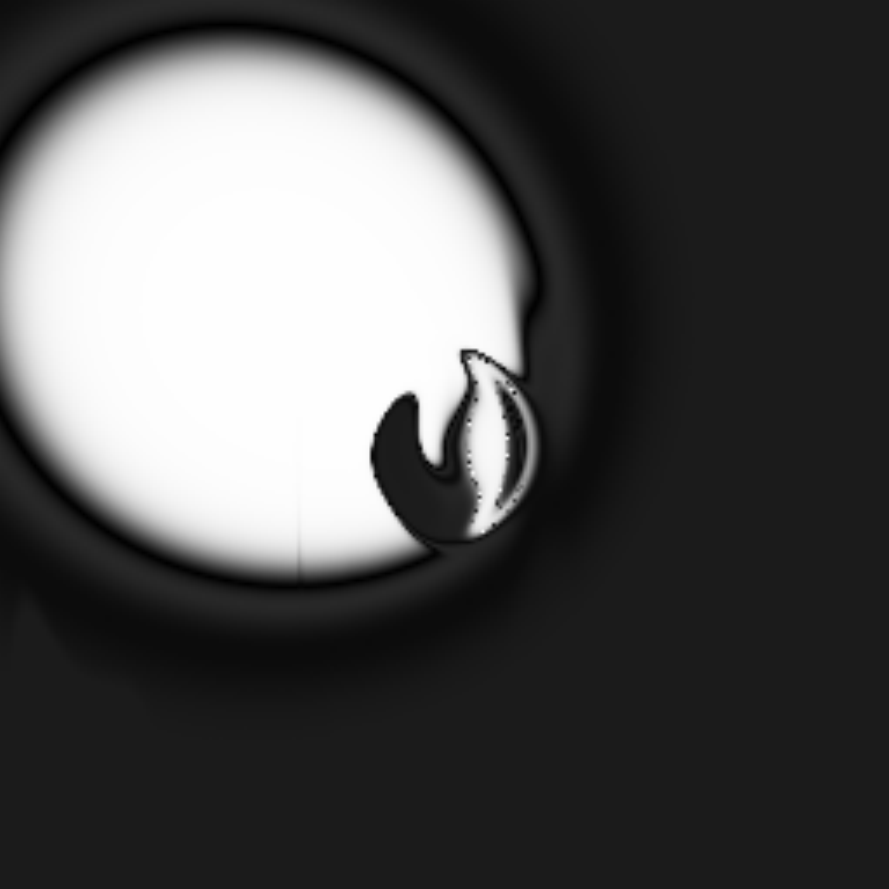}
\includegraphics[width=0.15\figwidth]{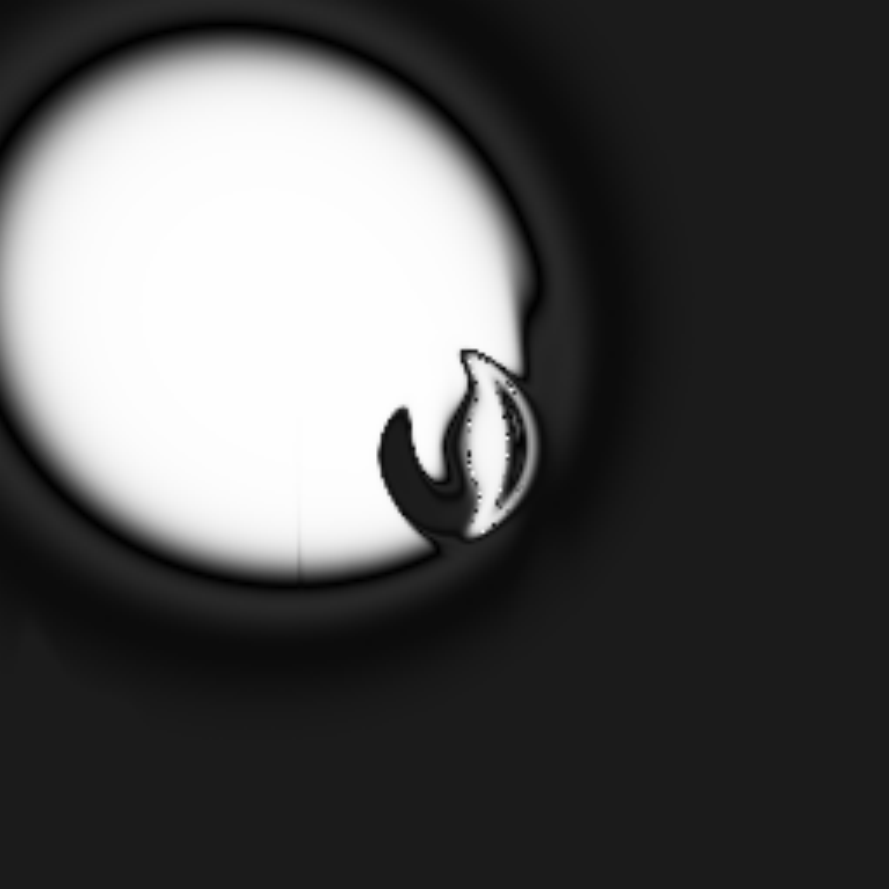}
\includegraphics[width=0.15\figwidth]{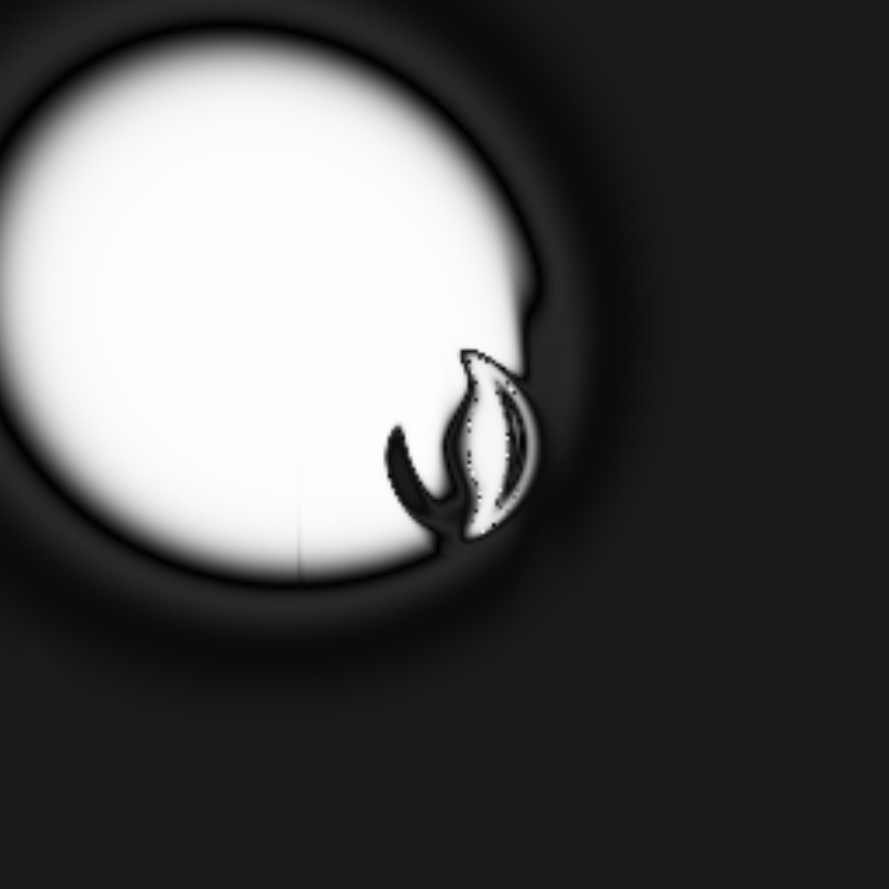}
\includegraphics[width=0.15\figwidth]{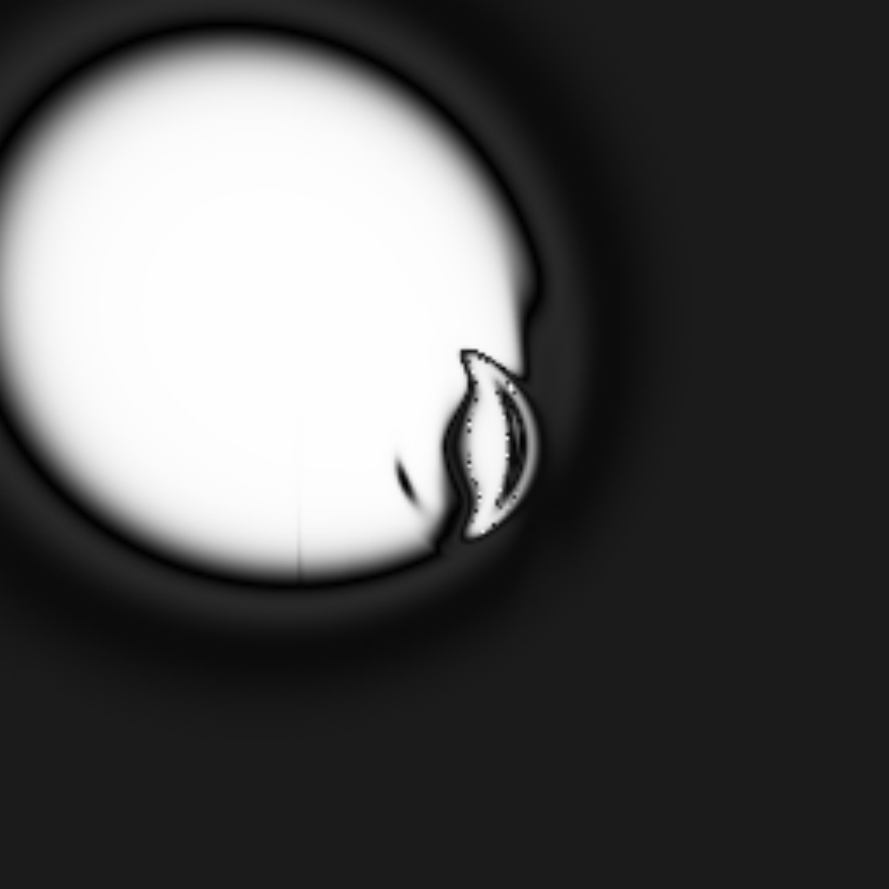}
\includegraphics[width=0.15\figwidth]{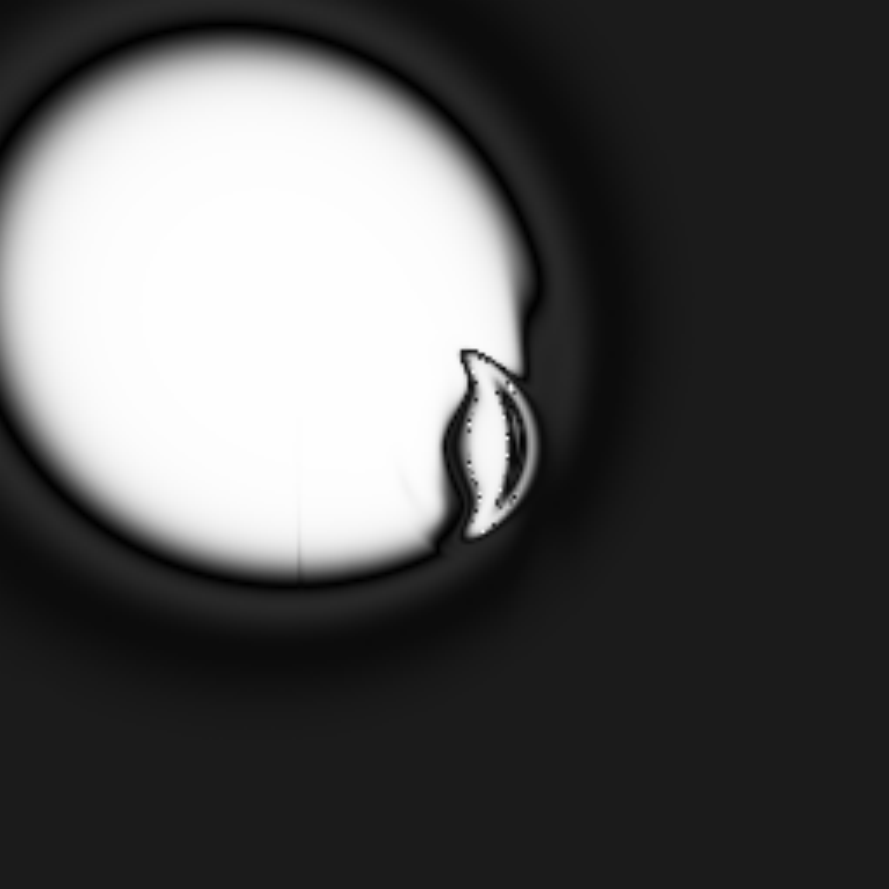}

Object size without shadow (Fig~\ref{fig:networkTrans}, Link 3)\\
\includegraphics[width=0.15\figwidth]{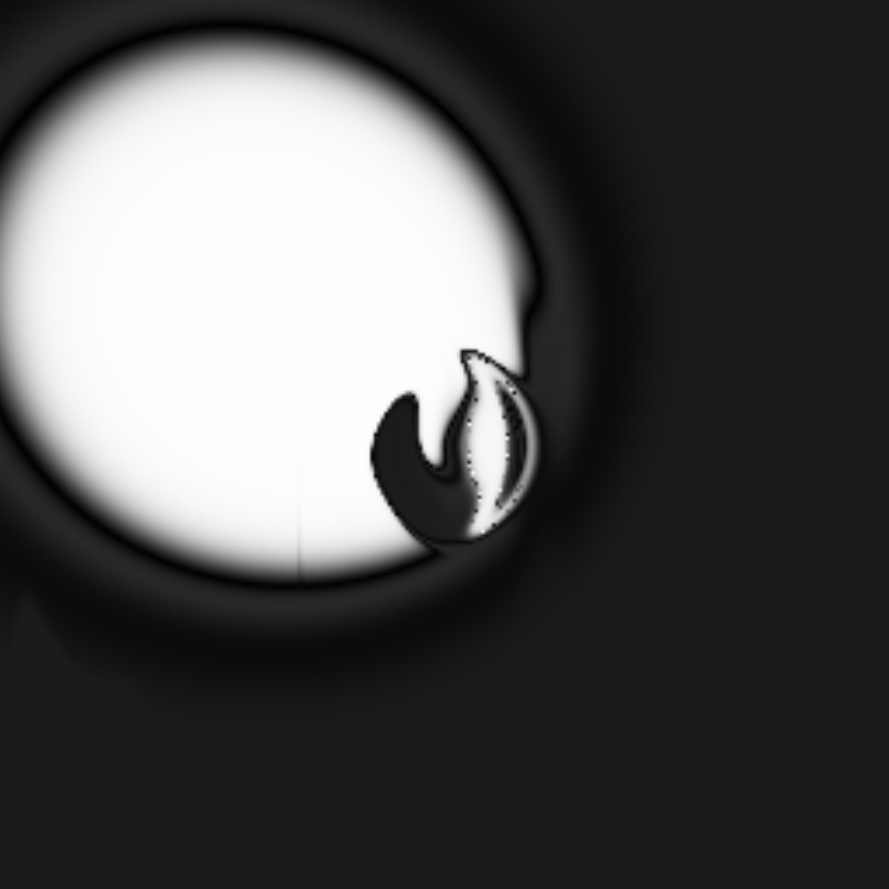}
\includegraphics[width=0.15\figwidth]{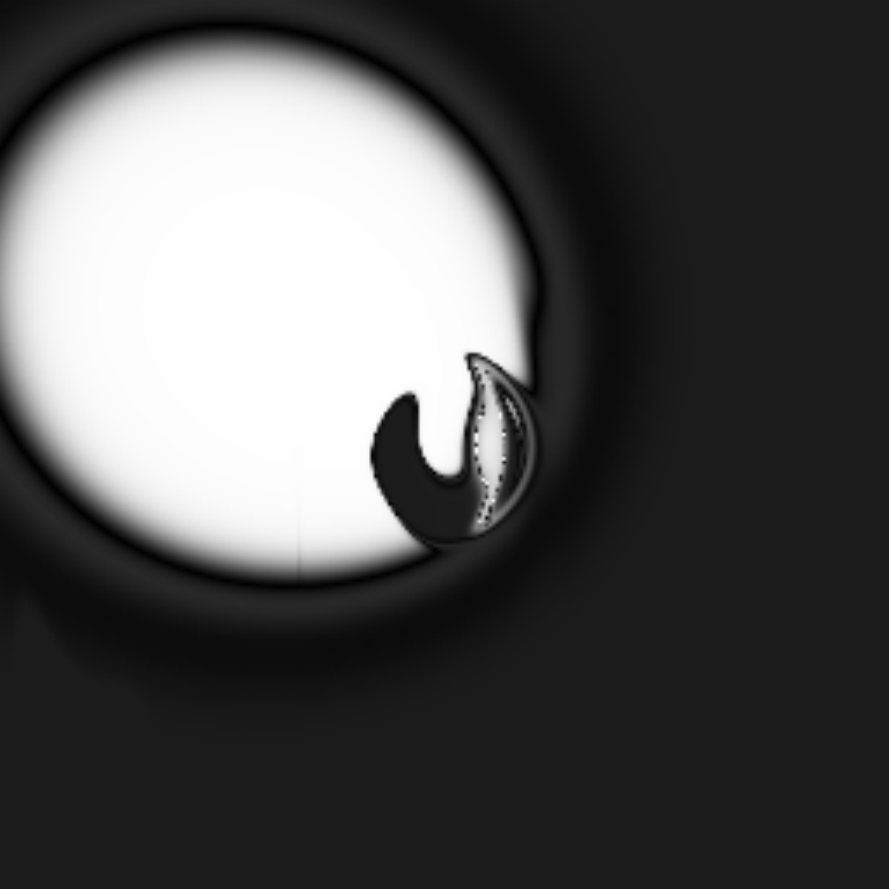}
\includegraphics[width=0.15\figwidth]{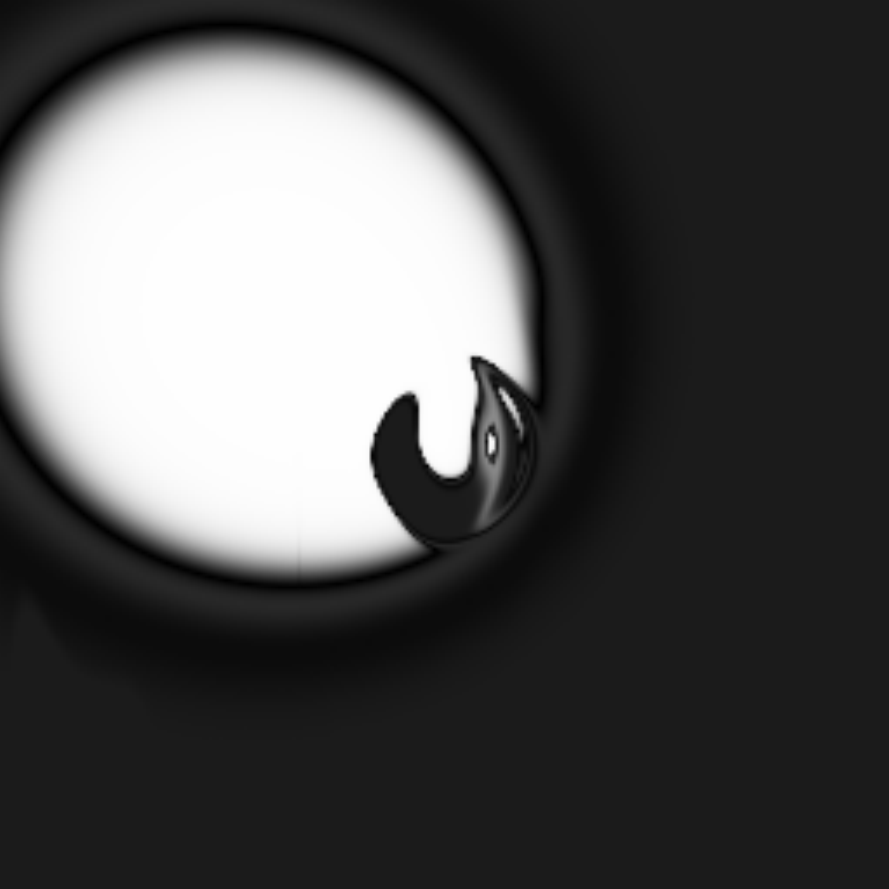}
\includegraphics[width=0.15\figwidth]{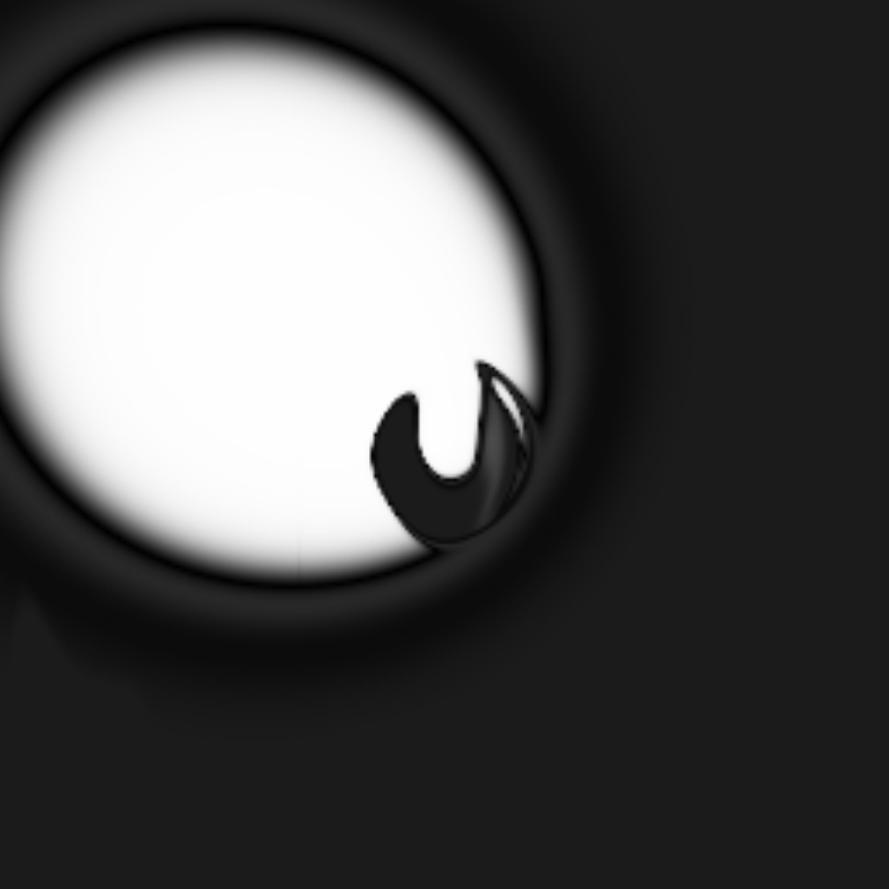}
\includegraphics[width=0.15\figwidth]{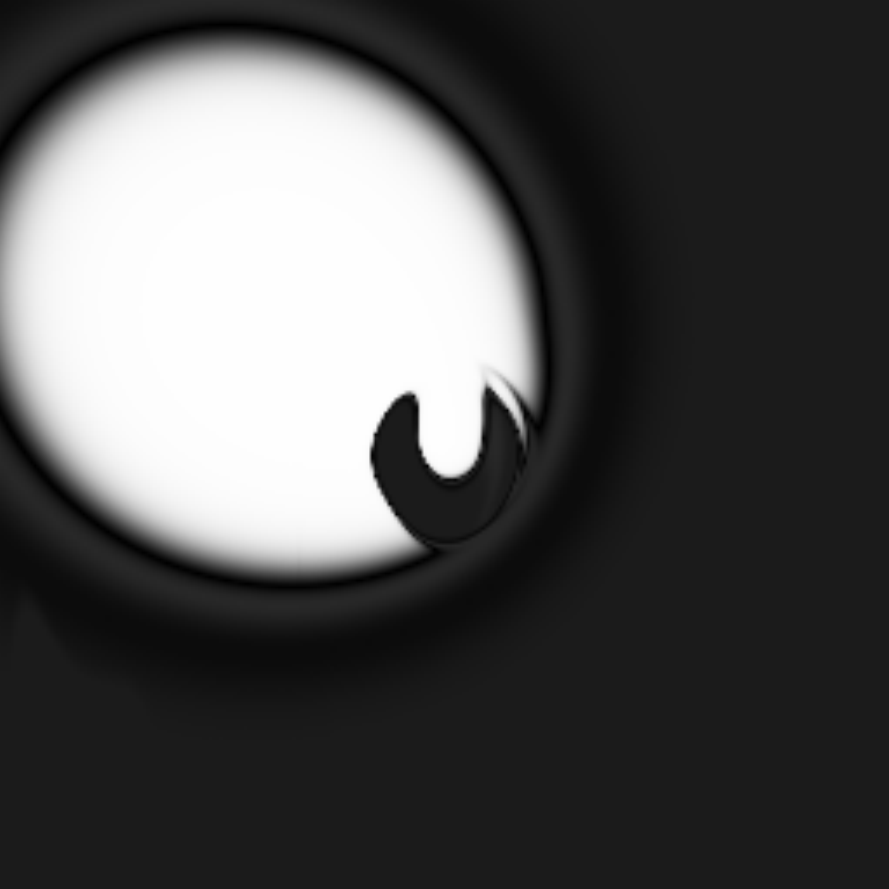}

Spotlight from left to right (Fig~\ref{fig:networkTrans}, Link 4)\\
\includegraphics[width=0.15\figwidth]{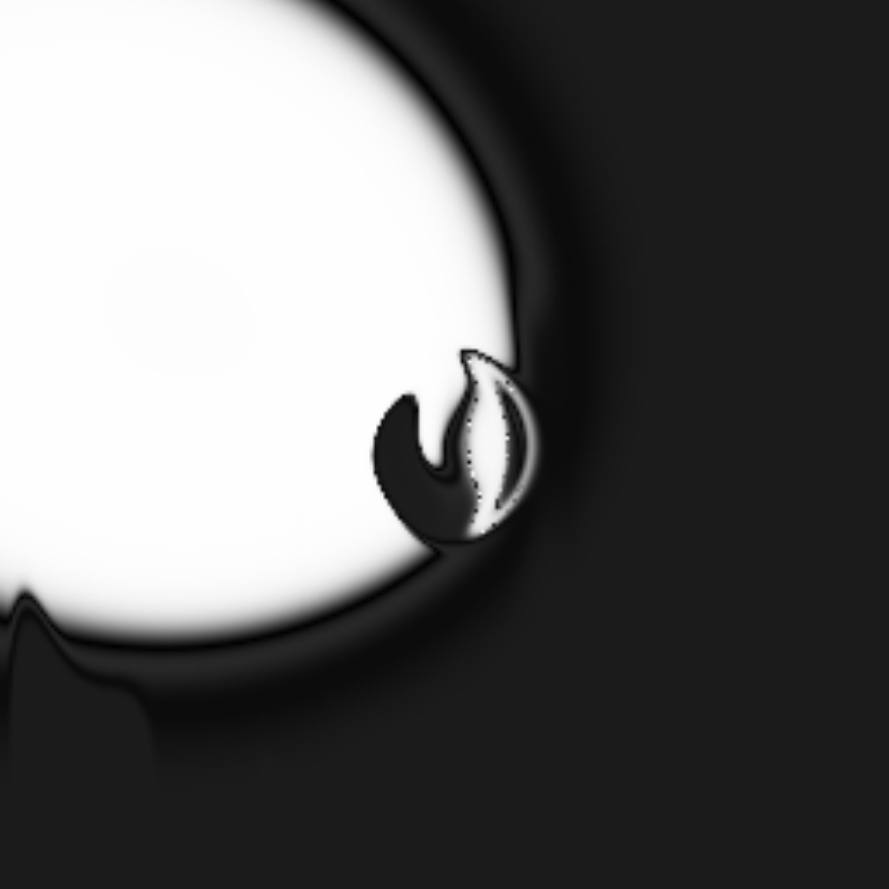}
\includegraphics[width=0.15\figwidth]{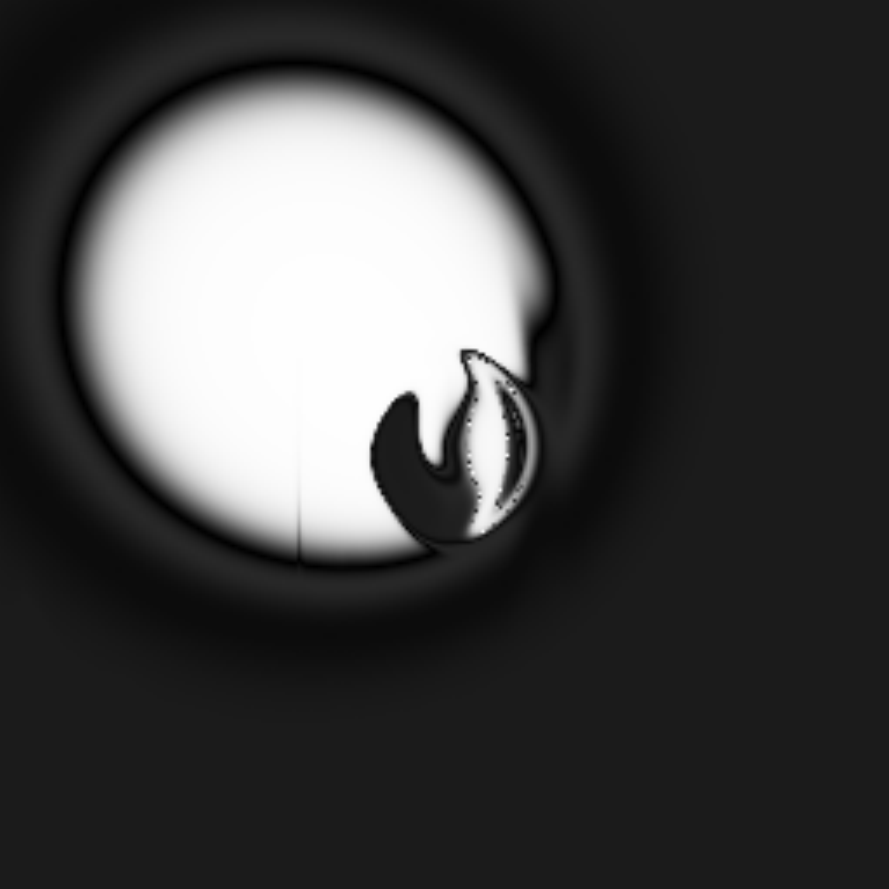}
\includegraphics[width=0.15\figwidth]{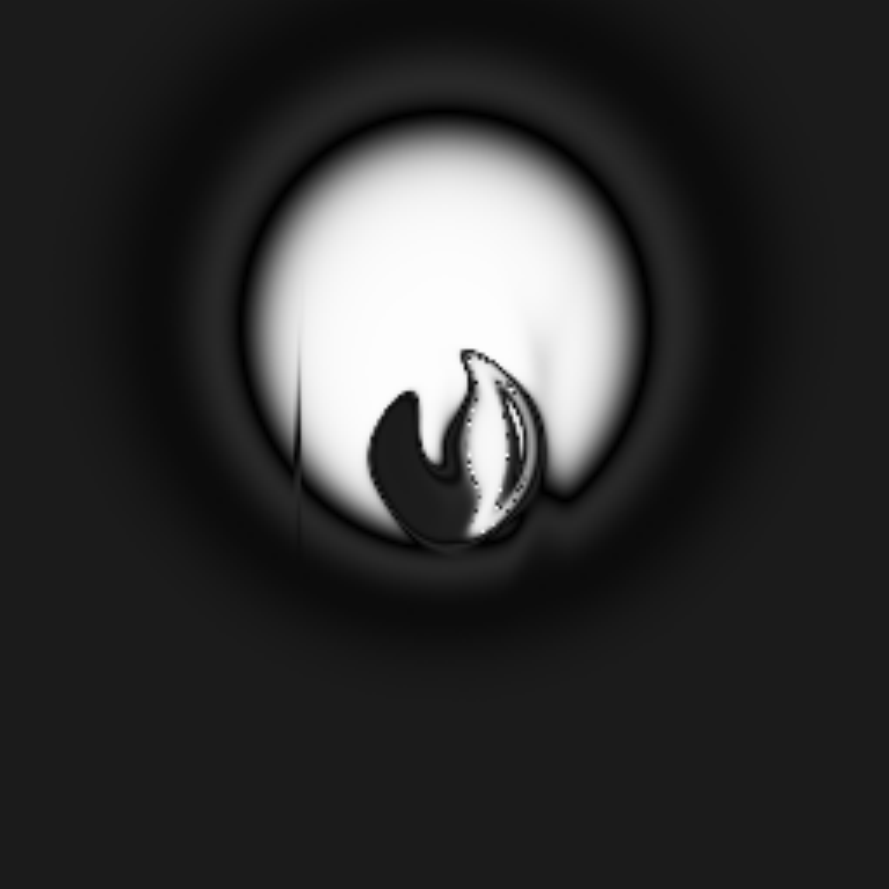}
\includegraphics[width=0.15\figwidth]{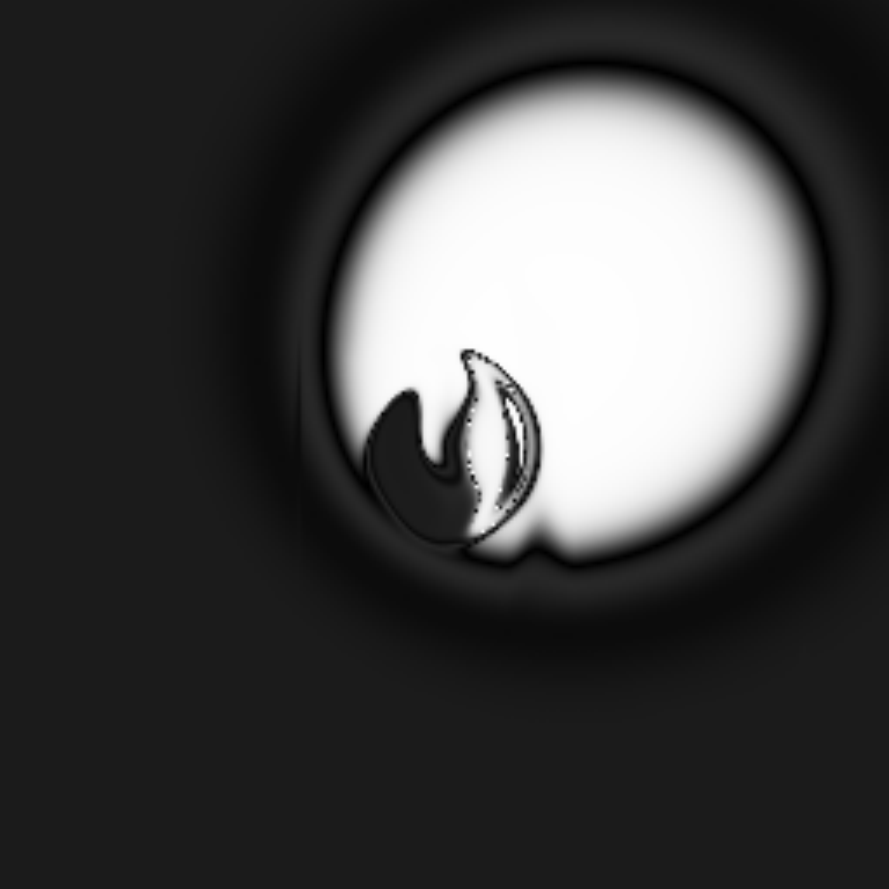}
\includegraphics[width=0.15\figwidth]{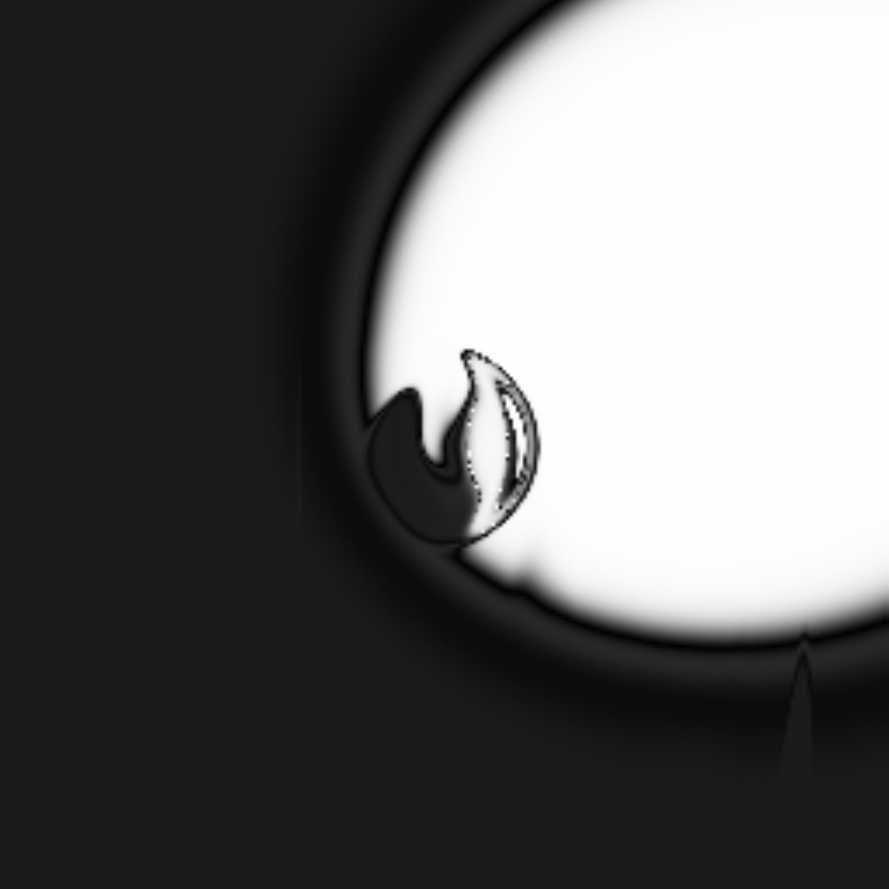}

Spotlight from top to bottom (Fig~\ref{fig:networkTrans}, Link 5)\\
\includegraphics[width=0.15\figwidth]{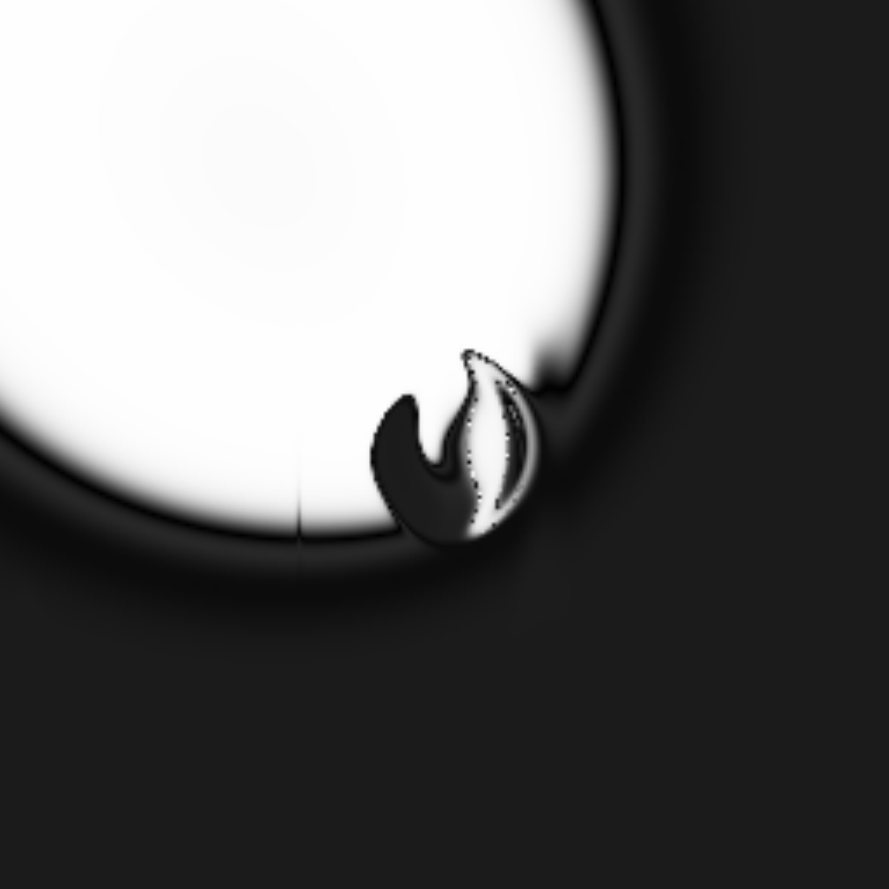}
\includegraphics[width=0.15\figwidth]{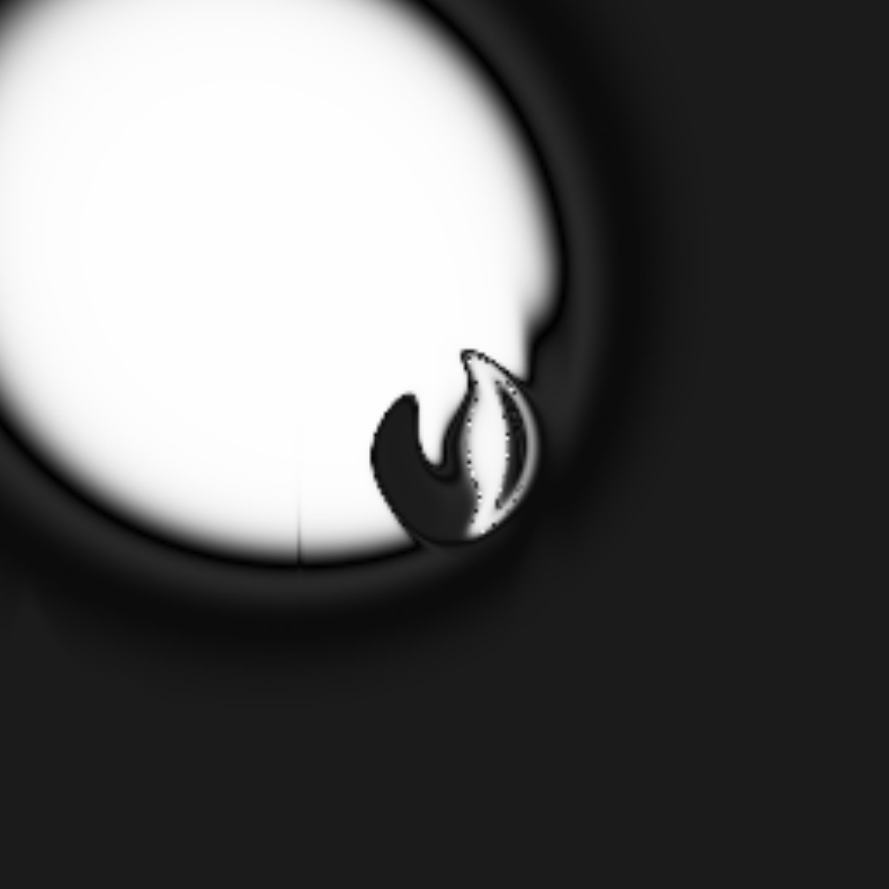}
\includegraphics[width=0.15\figwidth]{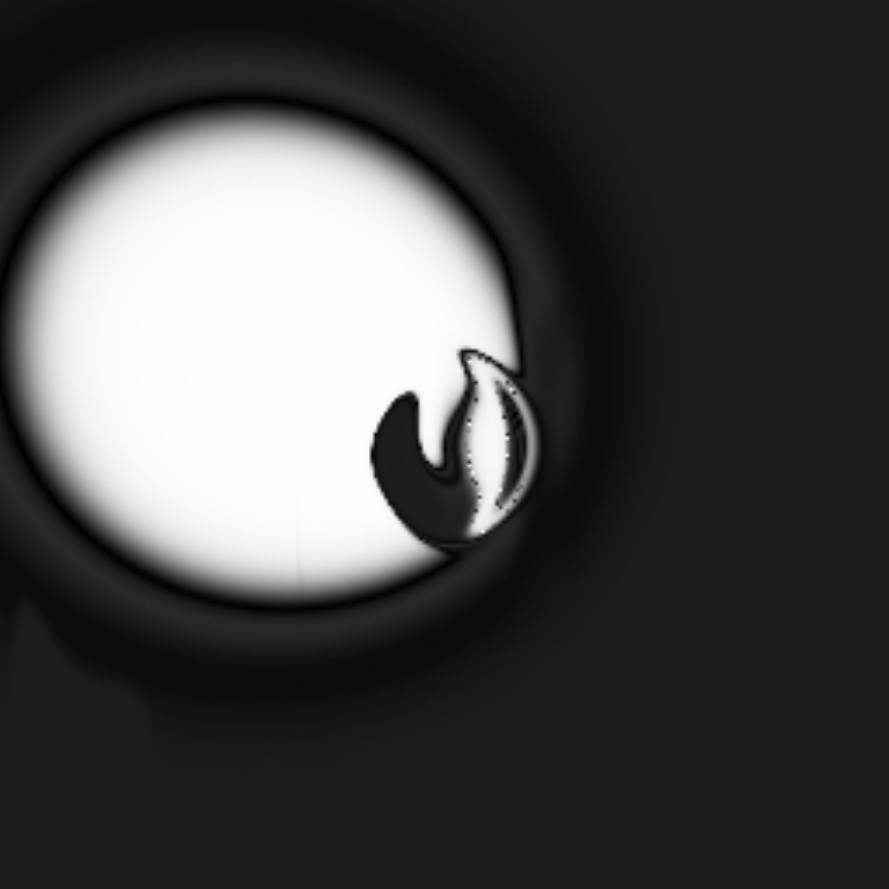}
\includegraphics[width=0.15\figwidth]{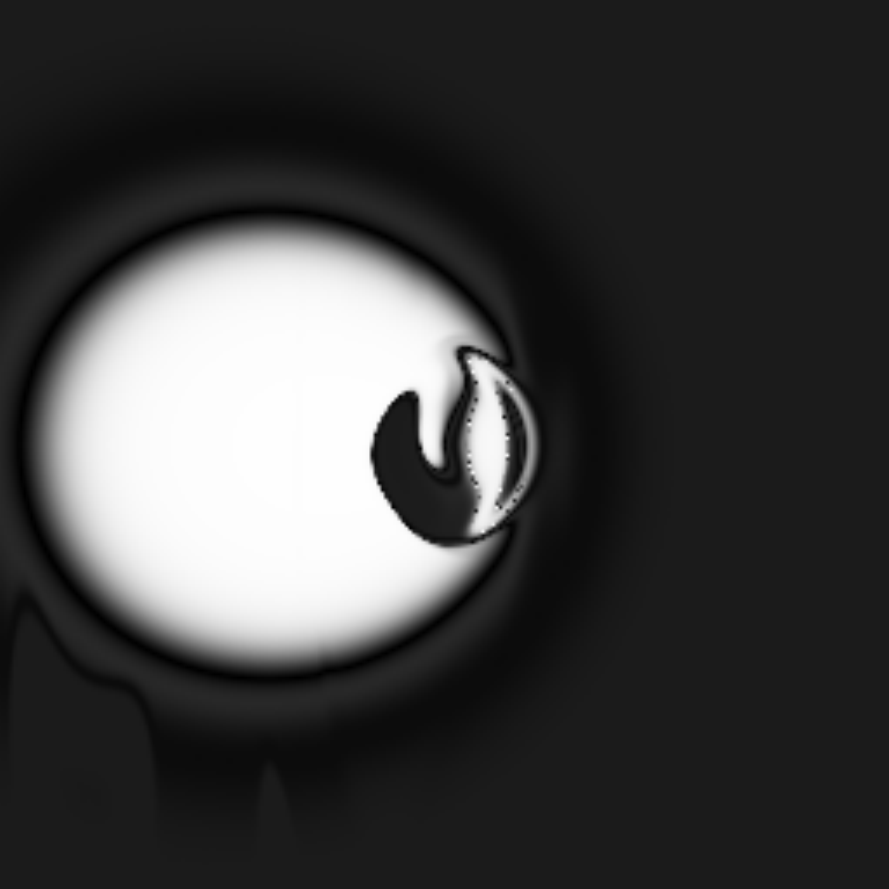}
\includegraphics[width=0.15\figwidth]{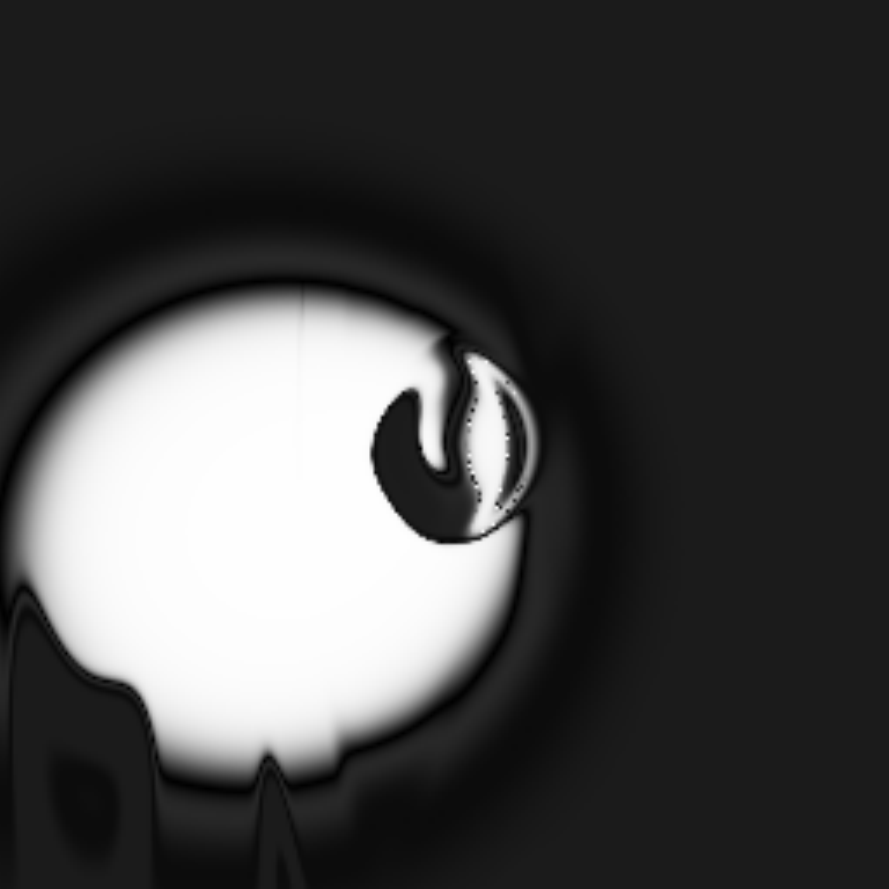}

Spotlight size (Fig~\ref{fig:networkTrans}, Link 6)\\
\includegraphics[width=0.15\figwidth]{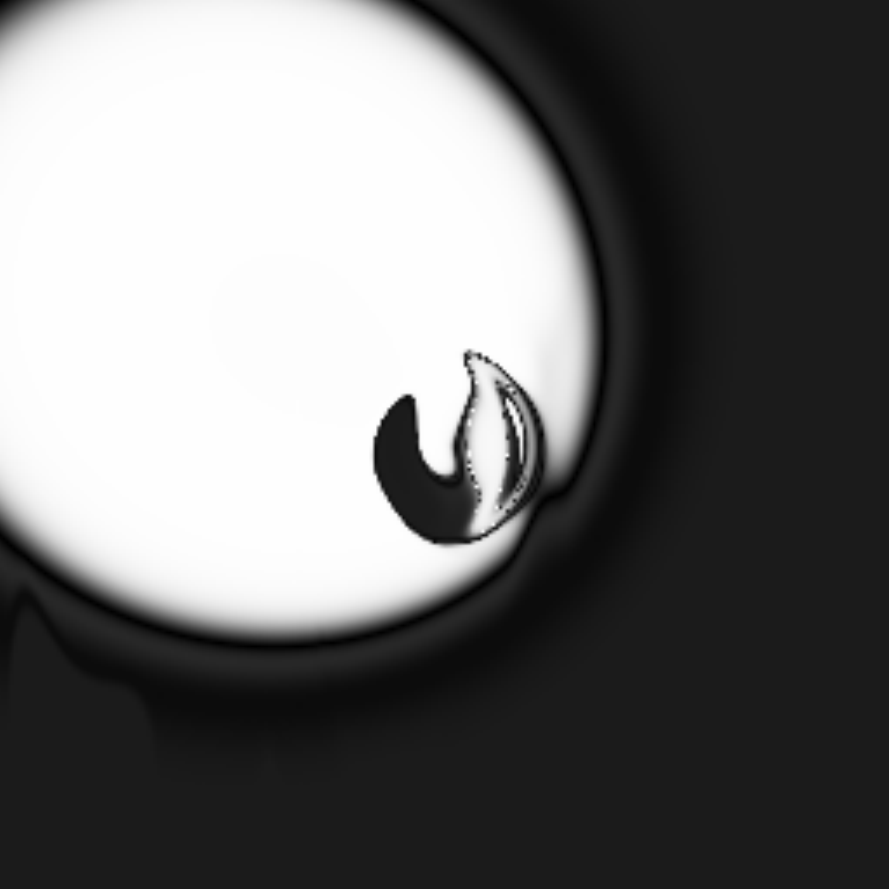}
\includegraphics[width=0.15\figwidth]{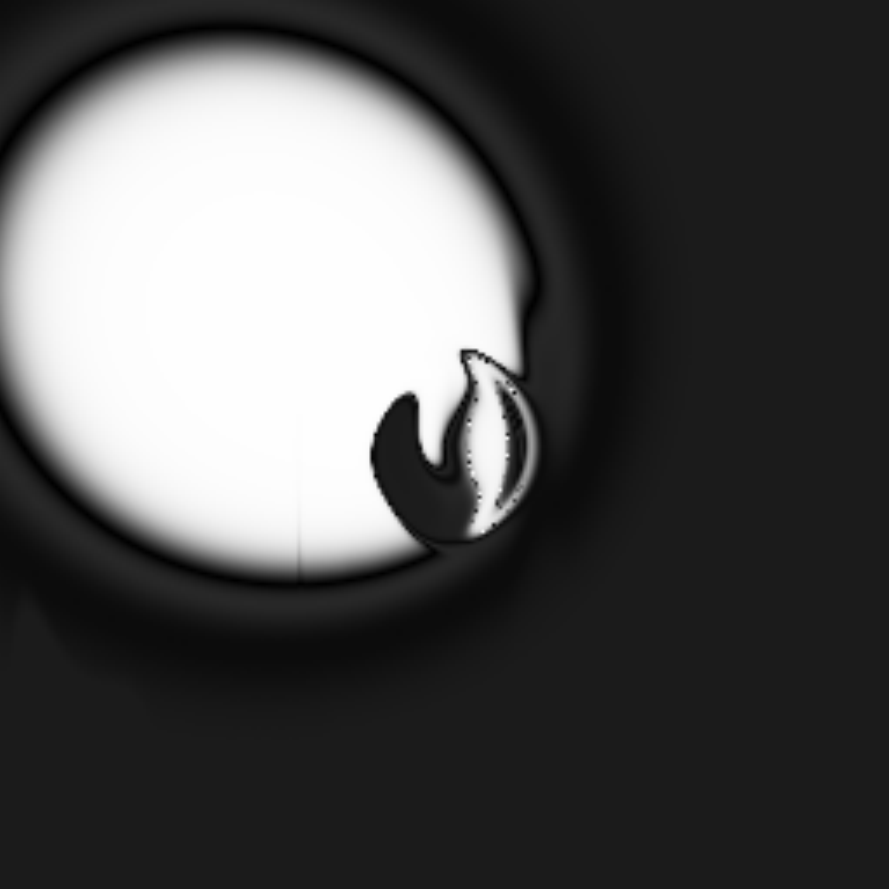}
\includegraphics[width=0.15\figwidth]{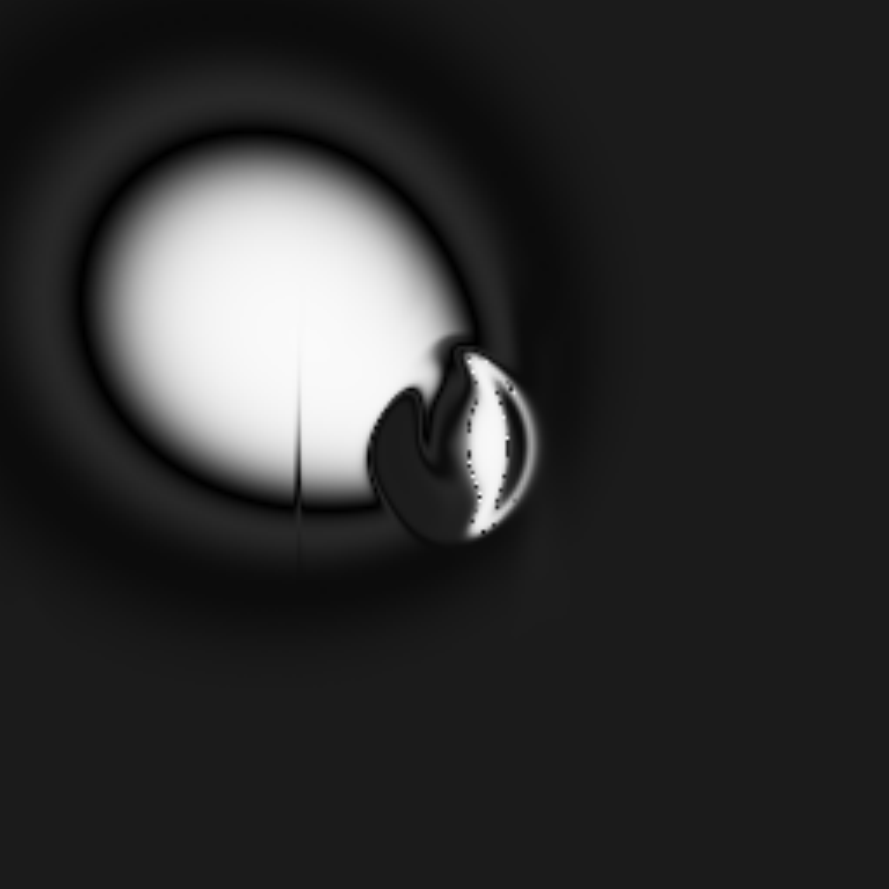}
\includegraphics[width=0.15\figwidth]{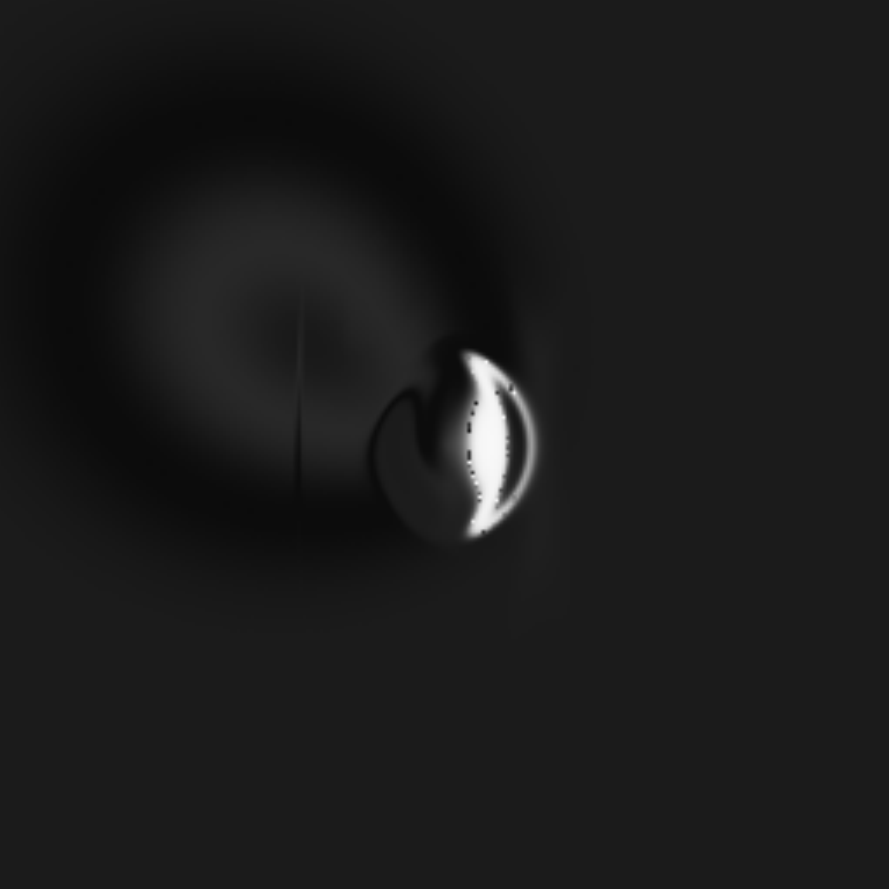}
\includegraphics[width=0.15\figwidth]{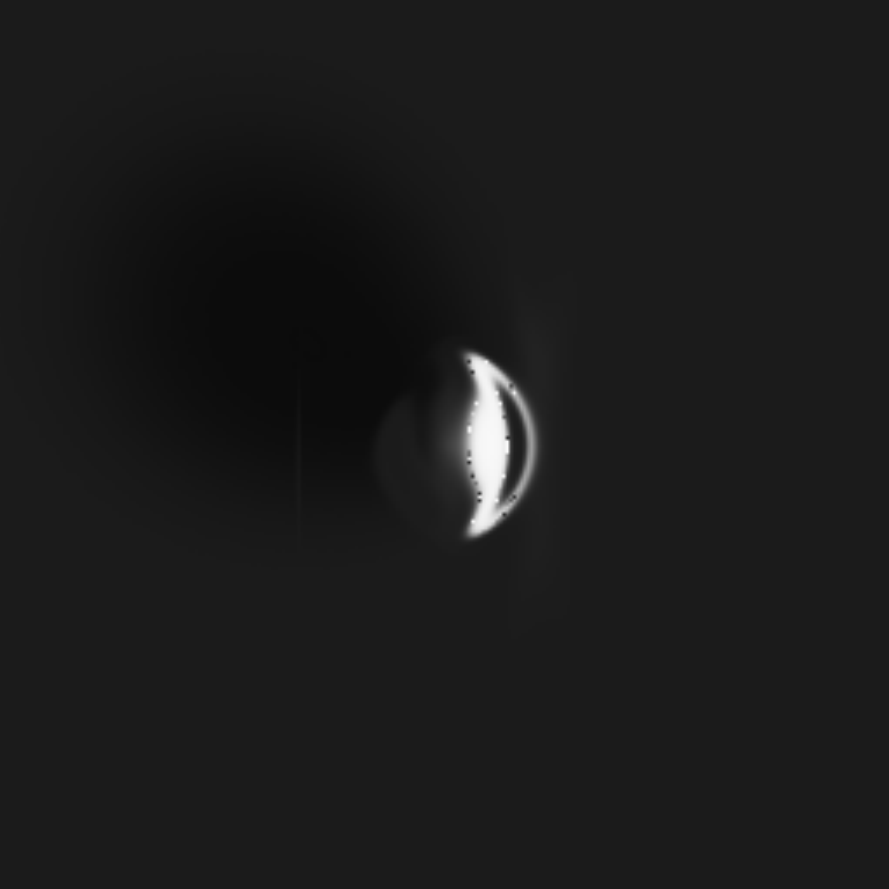}

\end{center}
\caption{\captionstart{}\textbf{The genome of the Spotlight Casting Shadow image, evolved on \picbreeder{},
 has canalized various dimensions of variation.} 
 The top panel shows the different components of the image. 
 Each row represents the effect of changing the values of a single gene (connection) from low to high
 in the underlying CPPN genome (Fig.~\ref{fig:networkTrans}, labels correspond to the link numbers).}
\label{fig:#1}
\end{figure}
}
\figlimelight{limelight}

From a visual inspection of the genome, color-coded to show the effect of each connection, 
it is apparent that the connections that control independent dimensions of variation are not randomly 
distributed throughout the genome (Fig.~\ref{fig:networkTrans}). 
Instead, the genome exhibits a modular and hierarchical organization whereby different clusters of 
connections enable the manipulation of different dimension of variation in the image,
and where ``higher level'' modules affect multiple aspects of the image while ``lower level'' modules
affect single aspects of the image.
The genome starts with two ``high level'' modules, one affecting the object and the shadow, 
and the other affecting the spotlight.
The ``Object And Shadow'' module feeds into two ``lower level'' modules 
that affect only the object or only the shadow of the object.
Lastly, all information is aggregated into a ``Global Lighting'' module,
which affects the brightness of the entire image without affecting the shapes within the image.

Curiously, the ``Object Only'' module also feeds into the ``Shadow Only'' module,
which raises the question why changes to the ``Object Only'' module do not affect the shadow.
This observation is important because, if changes to the ``Object Only'' module did affect the shadow,
there would be no ``Object Only'' module, and the image would have lost several independent dimensions of variation.
It turns out that the final image can be faithfully reconstructed by replacing the connections between 
the ``Object Only'' module and the ``Shadow Only'' module with 
connections from the bias input to the ``Shadow Only'' module,
thus demonstrating that these connections from the ``Object Only'' module to 
the ``Shadow Only" module only serve as a bias (SI Fig.~\ref{SI-fig:spotlight_bias}).
As such, the ``Object Only'' and ``Shadow Only'' modules are two low-level modules that 
are only related because they both receive their information from
the higher level ``Object And Shadow'' module.

This structural organization directly corresponds to
the different dimensions of variation that have been canalized;
the object and the shadow can be manipulated both together and separately 
because there exist hierarchically organized modules that process those aspects together and separately.
This also explains why there are no connections that affect only the object and the spotlight together (but not the shadow);
there is simply no location in the genome where only these two properties are processed together.
As such, the Spotlight Casting Shadow genome is a practical example 
of how structural organization can lead to canalization.
Witnessing this structural organization in the Object Casting Shadow and many other genomes
(SI Sec.~\ref{SI-sec:si_analyzed_genomes}), led us to hypothesize that, 
in general, some canalizations in CPPN genomes 
are facilitated by a modular and hierarchical structure. 
Because modularity and hierarchy can be quantitatively measured, 
we investigated whether these properties exist at elevated levels in Picbreeder genomes.

\newcommand{\fignetworktrans}[1]{%
\begin{figure}
\centering
\includegraphics[width=0.8\figwidth]{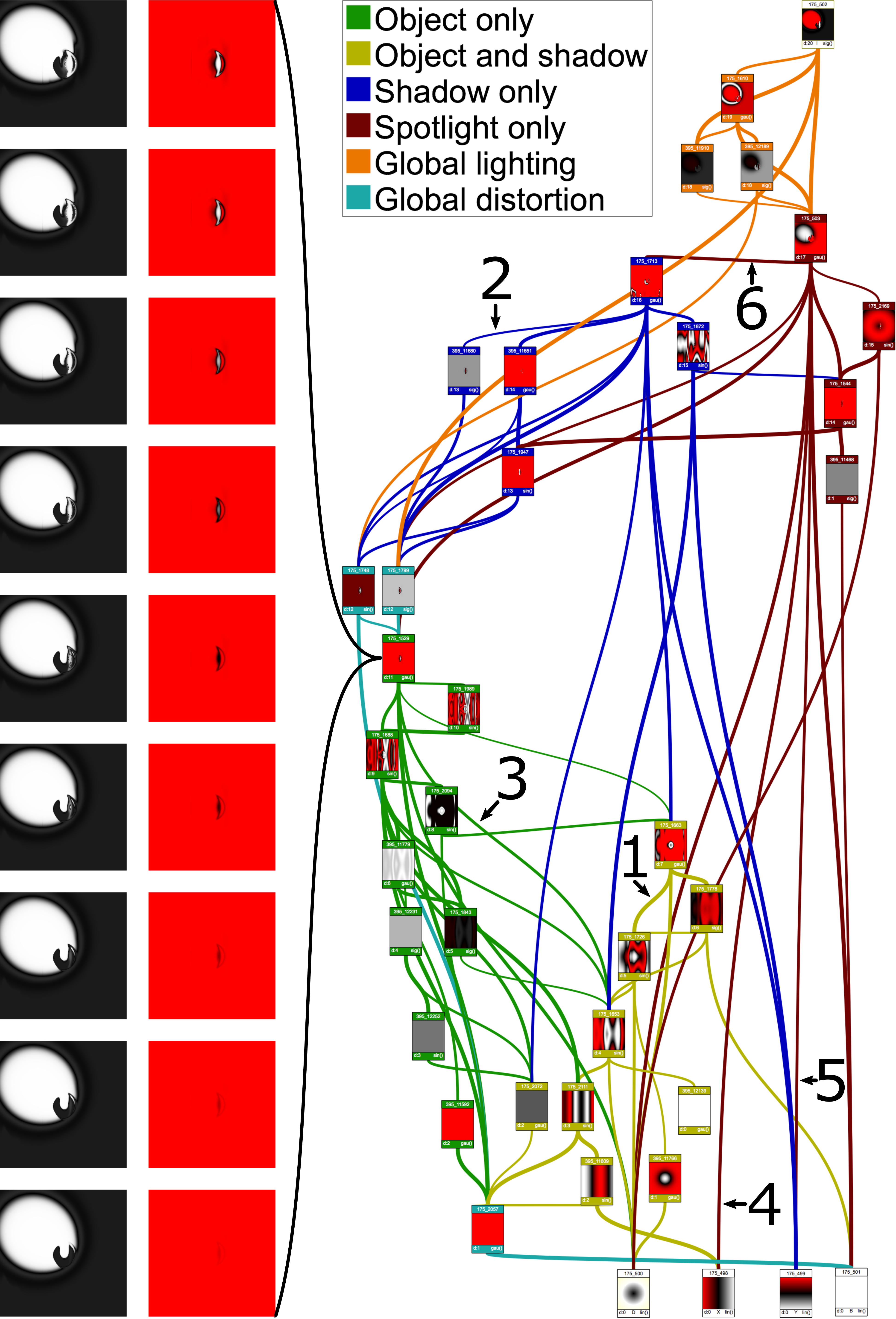}
\caption{\captionstart{}\textbf{The canalization of various dimensions of variation are expressed 
in the connections of the CPPN genome for the Spotlight Casting Shadow image.} 
\textbf{Right:} the CPPN genome with arrows pointing to links that 
individually change a canalized aspect of the image. 
The colors indicate which part of the network controls which aspect of the final image (see legend for details).
For visual clarity, connections that had no or little effect have been removed in this visualization 
(the full genome can be seen in SI Sec.~\ref{SI-sec:spotlightCastingShadow}). 
The numbers refer to link numbers from Fig.~\ref{fig:limelight}.
\textbf{Left:} the final and intermediate images when sweeping over link 3.
}
\label{fig:#1}
\end{figure}
}
\fignetworktrans{networkTrans}

We first tested the hypothesis that these genomes have evolved to have elevated levels of modularity. 
If \picbreeder{} images have indeed canalized dimensions of variation through modularity,
and if those canalizations provide an evolutionary advantage,
one would expect that, on average, \picbreeder{} genomes are more modular than randomly generated null models.
To test this, we approximated the maximal modularity-Q score for directed networks~\cite{Leicht2008} for each 
network in the \picbreeder{} database, which at the time we were given a copy of it contained 9585 genomes,
and we compared those values against the similarly approximated maximal modularity-Q score of random null models.
The modularity Q-score of a network, given a particular division of the network into modules, 
indicates the fraction of edges that lie within a module 
(as opposed to connecting two different modules), 
minus the expected value of that same fraction for a randomly connected network~\cite{Leicht2008}.
The network division that maximizes the modularity Q-score is known as an optimal split 
(approximated with an efficient, eigenvector-based method~\cite{Newman2006, Leicht2008}) 
and the corresponding Q-score of that optimal split is widely accepted  
as a measure of network modularity~\cite{Kashtan2005, 
clune2013originModularity, mengistu2016evolutionary, bullmore2012economy, meunier2010hierarchical, hagmann2008mapping}.

To generate fair null-models, two factors needed to be controlled for:
1) that the \picbreeder{} genomes were produced under a set of very specific constraints related to the NEAT algorithm
(e.g. a fixed number of inputs and outputs, no recurrent connections, no disconnected nodes),
and 2) that the networks in our data set are not all independent from each other.
To control for these factors, 10 null models were generated for each Picbreeder network,
where each null model was generated by iteratively applying NEAT ``add node'' and ``add connection'' mutations 
to the parent of the Picbreeder network until the null model
had the exact same number of nodes and connections as the \picbreeder{} network.
Here, the parent network refers to the most recent published ancestor of the Picbreeder network if it was branched,
or the minimal starting network if it was created from scratch.
Because \picbreeder{} does not feature any deletion mutations, 
this ensured that the null models underwent the same types of mutations as the actual network.
This way, the only explanation for a difference between a network and its null models lies with the selection
processes that happened between the parent and the child.
Finally, the average modularity Q-score of the null models was subtracted from the 
modularity Q-score of the real network to arrive at the residual modularity score provided throughout this paper.
Because we subtract null-model modularity-Q, 
a residual modularity greater than zero indicates that a network is more modular than expected by chance. 
We found that \picbreeder{} genomes are significantly more modular than the random null-models 
(median residual modularity: $0.0039 [0.0034, 0.0045]$ with $95\%$ bootstrapped confidence intervals, 
$p=0$ Wilcoxon signed rank test).

Another way to check whether modular genomes provide an evolutionary advantage in \picbreeder{}
is to see whether there exists a positive correlation between (residual) modularity and fitness,
where fitness is expressed as the number of times an image was branched and subsequently saved by a user. 
There exists a significant relationship between residual modularity and fitness 
(Pearson's correlation coefficient: $0.026$, $p=0.012$), 
suggesting that modularity does indeed have a positive effect on successful reproduction (Fig.~\ref{fig:branchingModularityCorrelation}a).

\newcommand{\figbranchingmodularity}[1]{%
\begin{figure}[tb!]
\centering
\labfigo{a}{0.48}{70}{{"Figures/BranchingFactorCorrelations/corrected_mod_vs_branching_factor"}.pdf}
\labfigo{b}{0.48}{70}{{"Figures/BranchingFactorCorrelations/corrected_backward_hierarchy_vs_branching_factor"}.pdf}
\caption{\captionstart{}\textbf{Genomic modularity and hierarchy have a positive correlation with fitness.} 
\textbf{(a)} There exists a positive and significant correlation} between residual modularity and fitness 
(Pearson's correlation coefficient: $0.026$, $p=0.012$). 
\textbf{(b)} Similarly, there exists a positive and significant correlation between residual hierarchy and fitness 
(Pearson's correlation coefficient: $0.037$, $p=0.00028$). 
For genomes with different residual modularity and residual hierarchy levels, 
the bars indicate average fitness and the whiskers indicate the $95\%$ bootstrapped confidence intervals 
of the average (obtained by resampling 5000 times). 
Colors represent the number of images represented by each bar. 
Note that the color map depicts a log-scale, 
meaning that the lightly colored bars represent orders of magnitude fewer data points than the darkly colored bars. 
The black line indicates the best linear fit of the underlying data.
\label{fig:#1}
\end{figure}
}
\figbranchingmodularity{branchingModularityCorrelation}

Ideally, we would test whether modularity correlates with canalization,
but we currently have no general way of quantifying canalization.
Instead, we examined whether the algorithmically detected modules 
correspond to our manually annotated decompositions of the genomes,
which would indicate that the algorithmically detected modules are indeed associated with a particular function.
While not as fine-grained, 
the automatically detected modules, in terms of connections, do correspond roughly with the 
manually labeled modules for the Spotlight Casting Shadow image (Fig.~\ref{fig:modulesWonton}).
However, this alignment is not as clear for other images, 
especially when the objects in the image are more overlapped (SI Sec.~\ref{SI-sec:si_analyzed_genomes}). Thus, 
while we have shown that modularity provides an evolutionary advantage,
it is still an open question to what degree modularity leads to canalization in \picbreeder{} images.

\newcommand{\figmoduleswonton}[1]{%
\begin{figure}
\labfigo{a}{0.45}{97}{{"Figures/CppnObjectCastingShadow/395_Spotlight_Casting_Shadow_Auto_Decomp_Main"}.pdf}
\hspace{1cm}
\labfigof{b}{0.45}{97}{-5}{{"Figures/CppnObjectCastingShadow/395_Spotlight_Casting_Shadow_Manual_Decomp_Main"}.pdf}
\caption{\captionstart{}\textbf{Canalization may occur through hierarchical and modular structures.} 
The modules found to produce the maximum modularity score are shown on the 
left (\textbf{a}) and those found after manual functional analysis are shown on the right (\textbf{b}). 
While only discovering two modules, 
the split found algorithmically to produce the maximum modularity value corresponds to the functional analysis of the network, 
roughly dividing the network in an ``Object'' module (left, blue) and a ``Spotlight And Shadow'' module (left, red).}
\label{fig:#1}
\end{figure}
}
\figmoduleswonton{modulesWonton}

Second, we quantitatively investigated the role of hierarchy within the subject genomes.
As with modularity, 
we first examined whether \picbreeder{} images are more hierarchical than randomly generated networks.
To do so, we quantified network hierarchy based on a metric described by Mones et al.\ (2012)~\cite{mones2012hierarchy}.
The idea of this metric is that, in hierarchical networks, 
a small number of nodes have a large influence while most nodes have little influence, 
whereas in a non-hierarchical network nodes tend to have more similar levels of influence. 
Thus, if a network is hierarchically organized, we expect a greater variance in node influence 
within the network than in a non-hierarchical network. 
Mones et al. (2012)~\cite{mones2012hierarchy} quantify influence in terms of Local Reaching Centrality (LRC), 
originally defined as a function of the number of reachable nodes and the weights along the paths to those nodes.
Because we are interested in CPPN structure regardless of weights,
we define LRC solely based on the number of reachable nodes,
as described in previous work~\cite{mengistu2016evolutionary}.
Given an LRC value for each node, the raw hierarchy can be calculated as the average of the 
normalized differences between each node and the maximum LRC in the network.
As in previous work that measured the hierarchy of feedforward networks with this metric, 
we reverse all edges before we apply the measure to avoid certain pathological results~\cite{mengistu2016evolutionary}.
In addition, to control for the effect of evolutionary constraints and interdependence between related networks, 
the average raw hierarchy of the ten null models described previously was subtracted from the hierarchy score of the original network to arrive at the residual hierarchy score reported throughout this paper.

On average, \picbreeder{} networks are significantly more hierarchical than the randomly generated null models 
(median residual hierarchy: $0.00018$ [$1.1 \cdot 10^{-16}$, $0.00052$] and $95\%$ bootstrapped confidence intervals, 
$p=8.0 \cdot 10^{-118}$ Wilcoxon signed rank test). 
In addition, there exists a significant and positive correlation between residual hierarchy and fitness 
(Pearson's correlation coefficient: $0.037$, $p=0.00028$), 
where networks with a higher residual hierarchy have an 
increased fitness (Fig.~\ref{fig:branchingModularityCorrelation}b).
Thus, it appears that in addition to modularity, hierarchy also has a positive effect on the reproductive success of images,
meaning that users unknowingly select for genomes that have these organizational properties.
Because the hierarchy measure does not provide a hierarchical decomposition,
we cannot check whether the measured hierarchy does indeed correspond with the manually observed hierarchy.
As such, testing to what degree hierarchy leads to canalization in \picbreeder{} images remains a topic for future work.

Thus far, we have shown images that have canalized various dimensions of variation, 
as well as an example of how, in some cases, structural organization can be the root of such canalization.
This leaves us with the question of how these canalizations behave over evolutionary time.
To answer this question, we examined the role of single connections in all
ancestors of an image titled ``Man Standing Silhouette,''
a descendant of the Spotlight Casting Shadow image. 
Because the NEAT algorithm in Picbreeder labels connections with historical markings~\cite{stanley2002evolving}, 
it is straightforward to track connections over generations.
We found that these canalizations and their genetic causes persist across evolutionary time, 
serving similar functional roles even in very different looking images (Fig.~\ref{fig:modulesOverTime}). 
The fact that these innovations are preserved over evolutionary time may explain why canalization appears to be so ubiquitous
on \picbreeder{}; even if the emergence of these innovations is rare, 
their persistence means that genomes can accumulate them over time.

\newcommand{\figmodulesovertime}[1]{%
\begin{figure}
\includegraphics[width=1.0\textwidth]{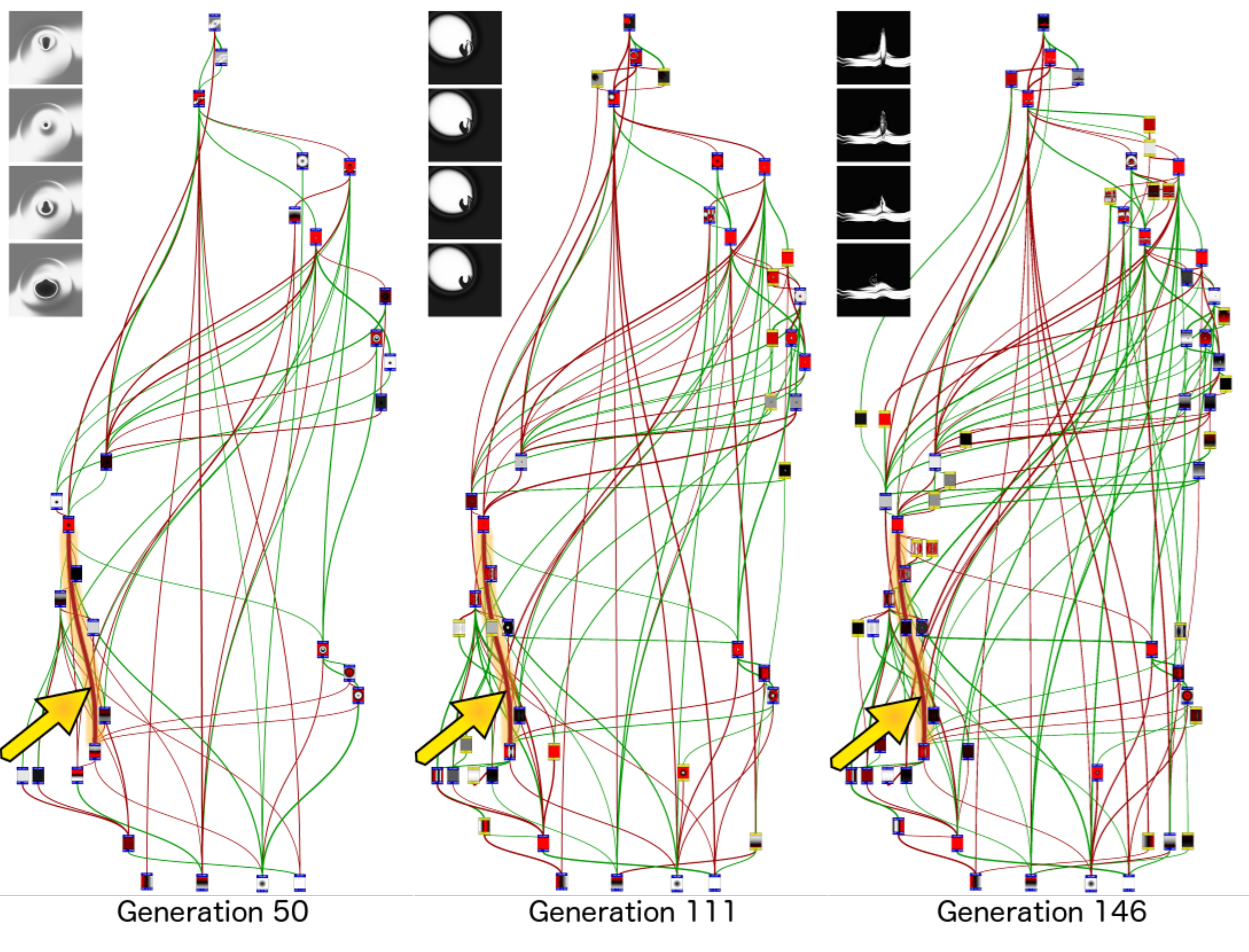}
\caption{\captionstart{}\textbf{Once discovered, canalizations are preserved.} 
The small images at the top of the figure display the variation when sweeping over different values 
for the connection indicated by the arrow. 
\textbf{Left:} A single link (indicated by the arrow) affects only the 
size and shape of the keyhole in an object a user titled ``Doorknob.'' 
\textbf{Middle:} The same link in a descendant 61 generations later only affects 
the size of the object in the Spotlight Casting Shadow image. 
\textbf{Right:} Another 35 generations later the same link only affects the size 
and shape of the man in an image titled ``Man Standing Silhouette.''}
\label{fig:#1}
\end{figure}
}
\figmodulesovertime{modulesOverTime}

\newcommand{\figlineage}[1]{%
\begin{figure}
\centering
\includegraphics[width=0.90\figwidth]{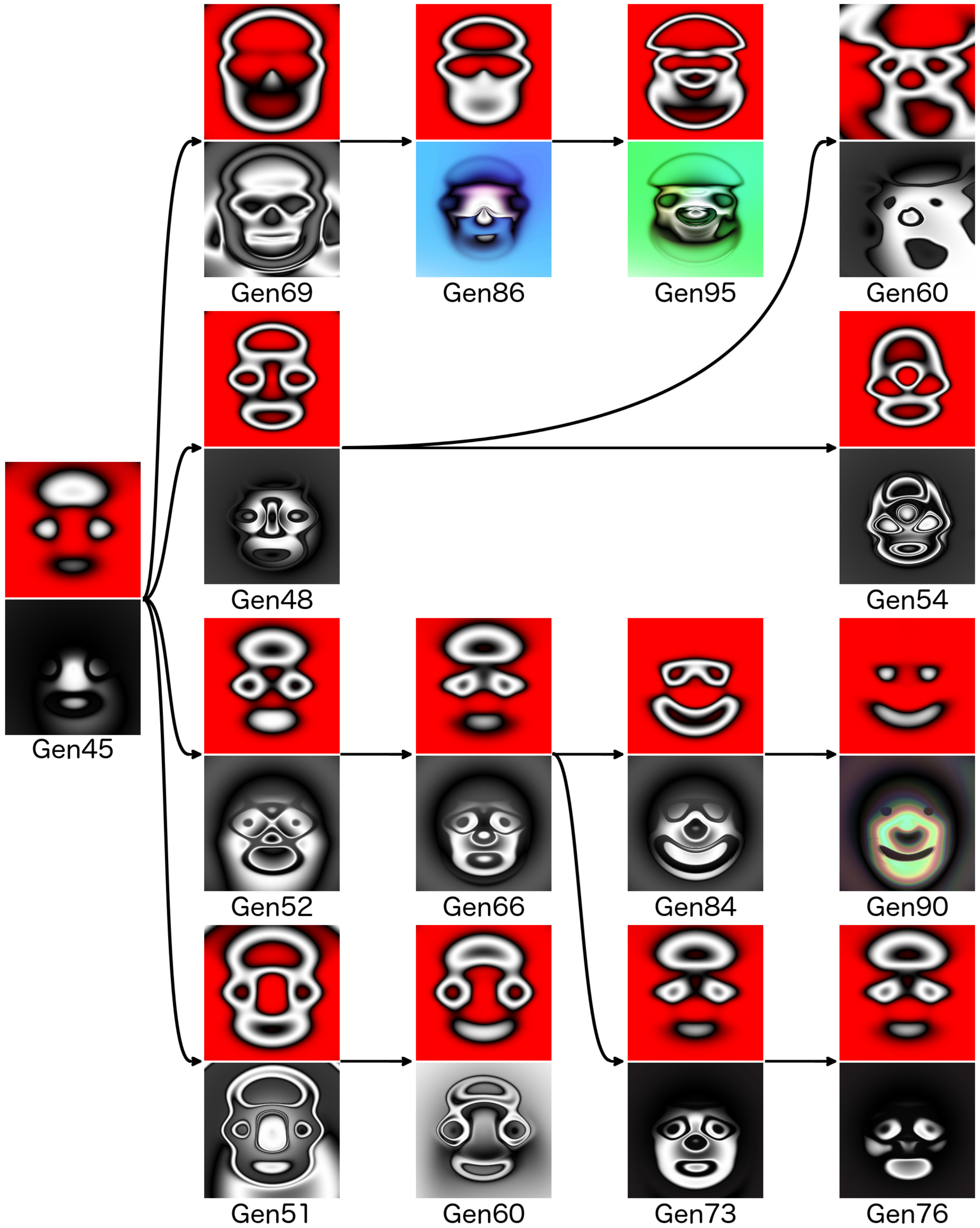}
\caption{\captionstart{}\textbf{Once discovered, canalizations can radiate.} 
Different lineages all containing the same protoface. 
For every pair of images, the top image shows the internal protoface 
(the same node is preserved and shown in all descendants), 
and the bottom image shows the final image 
(i.e.\ the output of the CPPN).  
The discovery of the protoface was a major innovation in \picbreeder{} 
(occurring first in the leftmost image, named ``Ghost Face Spooky''). 
This protoface enabled the evolution of a wide variety of different faces (all the other images shown).
In some lineages the protoface changed dramatically, 
while in other lineages the protoface remained virtually the same, 
and most variation there is due to changes outside of the protoface elsewhere in the genome.
The full CPPN genome with the annotated location of the protoface is available for four of these images in the SI 
(SI Sec.~\ref{SI-sec:sadface},~\ref{SI-sec:letterg},~\ref{SI-sec:face}, and~\ref{SI-sec:skull}). 
}
\label{fig:#1}
\end{figure}
}
\figlineage{lineage}

Similarly, once an interesting image structure is discovered, 
the genetic structures that encode it can be preserved throughout the evolutionary process.
One especially striking example of this process comes from an image named ``Ghost Face Spooky,''
which contains a genetic structure that gives rise to a protoface
(Fig.~\ref{fig:lineage} left and SI Fig.~\ref{SI-fig:minimal_face_mask}).
In its descendants, the underlying image concept (e.g.\ a face) is still present, 
but it can be altered in a variety of ways to result in very different face images (Fig.~\ref{fig:lineage}).
In other words, the nodes and connections that produce the face concept are roughly preserved, 
but the exact weights and connectivity of those structures, 
and thus the pixel-by-pixel image pattern that results from it, can change dramatically.
As such, the genetic structures that give rise to this face concept
functionally act as a ``face'' module that is preserved yet modified throughout generations.
This phenomenon is somewhat reminiscent of ``adaptive radiations''~\cite{gould1977ontogeny}, 
where once an evolutionary innovation is discovered (e.g. the four-legged body plan) 
there is a cascade of new evolutionary species that take advantage of the new innovation, 
but apply it in very different ways (e.g.\ elephants, dolphins, crocodiles, kangaroos, different apes, etc.).

\section{Discussion}

While we have shown examples of canalization in \picbreeder{} images, 
it is still unclear why canalizations have evolved in this system, but not in others.
Clearly, the user somehow selects for genomes that have canalized various dimensions of variation. 
However, the genome is not visible to the user and canalization, 
being a property solely describing how an image might change, 
is not directly discernible from the image itself.
In fact, one might have expected the dolphin (Fig.~\ref{fig:concept} right) to be better canalized, 
but it is not, potentially explaining why its descendants are generally of poor quality 
and have a low fitness (in terms of the number of times branched). 
Thus, to answer why canalizations have evolved on Picbreeder, 
we will have to examine how a collection of independent users was able to consistently select for canalized genomes.

One concept that has often been presented as a driver for the emergence of structural organization and evolvability, 
if not canalization directly, is the idea of a changing, rather than a static, 
environment~\cite{kashtan2007varying, draghi2008evolution}.
While previous research generally alternated between a few fixed 
environments~\cite{kashtan2007varying, draghi2008evolution},
the Picbreeder system takes such selection to extremes, 
because the multitude of different users, 
and how those users may change their objectives,
results in a highly dynamic environment where there exists selection 
in many different directions that continually change over evolutionary time.
Such a selection regime, where there exists selection in many different directions,
has also been referred to as \emph{divergent search}~\cite{pugh2016quality}.
The divergent nature of Picbreeder has been posited as an essential property for its success,
as the resulting images often do not resemble the intermediate stepping 
stones~\cite{woolley2011deleterious} (see also Fig.~\ref{fig:lineage}).
Consider the case of evolving the Spotlight Casting Shadow image (Fig.~\ref{fig:modulesOverTime} middle).
Previous research has shown that direct selection for any particular target image, 
such as by taking the difference in pixel intensities as a fitness measure,
only works for very simple shapes, such as the circle of the spotlight~\cite{woolley2011deleterious}.
More complex shapes, like the object with its shadow, are unlikely to ever be discovered this way.
However, from its evolutionary history we know that
to discover the Object Casting Shadow image, we may first have to
select for something that resembles a door-knob with a keyhole (Fig.~\ref{fig:modulesOverTime} left).
While we know the evolutionary history for this particular image, 
there is no way of knowing the intermediate stepping stones for any yet to be evolved image.
Evolving the image of a house might require selecting images that resemble a fire hose, teakettle, and school bus first.
The many different selection pressures present in a system like 
Picbreeder circumvent the issue of unknown stepping stones by providing 
evolutionary advantages for anything that looks interestingly different,
thus preserving all potential stepping stones.

To understand how divergent search can increase evolvability,
it is helpful to examine what happens with the genome under different selection regimes.
When a genome is subject to selection towards a particular goal,
the genome tends to expand in size as it collects and preserves small beneficial 
mutations~\cite{woolley2011deleterious}.
For example, in the domain of images, 
such mutations may cause a small number of pixels to get closer to their desired intensities.
This way, the genome incrementally acquires a structure that allows for small, local changes.
The downside of such genomic growth is that, in the extreme, 
every aspect of the phenotype can be changed independently,
making coordinated changes much less likely.
In the Spotlight Casting Shadow image, changing the size of the spotlight 
would require hundreds of coordinated mutations if every pixel had to be adjusted independently.
Such fine-grained genetic representations are far less likely to evolve in a system 
with a divergent selection regime, such as Picbreeder.
When there is selection for interesting change,
rather than selection towards a particular goal,
mutations that are most likely to be preserved are mutations that have large,
yet coordinated effects (large uncoordinated effects, 
such as flipping every pixel in the image randomly, are generally not considered interesting).
For example, in the Doorknob image (Fig.~\ref{fig:modulesOverTime} left), 
a user is much more likely to select an image where the size of the 
keyhole is changed as a whole than an image where some individual pixels of the keyhole change color,
or where every pixel in the image changes color.
This way, selection for interesting change is likely to result in genetic structures that favor 
coordinated changes over small incremental changes, 
or large uncoordinated changes.

While single-objective algorithms~\cite{de2006evolutionary, eiben2003introduction}
(or algorithms with a few fixed objectives~\cite{Deb2002})
are still the norm within the field of evolutionary 
computation~\cite{de2006evolutionary, eiben2003introduction, Deb2002},
a new family of evolutionary search algorithms explicitly focuses on divergent 
search~\cite{lehman2008exploiting, mouret2015illuminating, nguyen2015innovation, lehman2011abandoning, pugh2016quality}.
These algorithms, which are often referred to as \emph{illumination} algorithms or \emph{quality diversity} 
algorithms~\cite{mouret2015illuminating, pugh2016quality}, 
attempt to find the unknown stepping stones towards solutions by selecting 
for individuals that are interestingly different from anything found before.
The main challenge for these algorithms is to quantify ``interestingness,'' 
because most problem domains allow individuals to be different in ways that are 
unlikely to result in stepping stones towards anything (e.g. white noise in image space).
The algorithm known as Novelty Search selects individuals with the help of a distance function, 
where individuals that are far away from previously discovered individuals have an evolutionary 
advantage~\cite{lehman2008exploiting, lehman2011abandoning}.
Essential for the success of Novelty Search is the choice of distance function,
because this function determines whether any particular difference is 
interesting~\cite{lehman2008exploiting, lehman2011abandoning, meyerson2016learning}.
Another algorithm, known as MAP-Elites, offers a large number of different niches
reserved for individuals with particular phenotypic characteristics~\cite{mouret2015illuminating}.
Here the success of the algorithm depends on the choice of characteristics,
which need to be descriptive enough to preserve potential stepping stones
without making the number of niches intractably large~\cite{mouret2015illuminating, tarapore2016different, cully2015robots}.
Directly based on MAP-Elites is an algorithm known as the Innovation Engine,
where the different niches are not merely reserved for individuals with different phenotypic characteristics,
but where each niche may have a completely different fitness function~\cite{nguyen2015innovation}.
For example, when evolving images, one niche may favor individuals resembling cars, 
while another niche may favor individuals resembling dolphins.
In its basic implementation the niches are determined in advance,
and thus the choice of niches will determine the success of the algorithm,
but in future implementations, niches may be determined dynamically~\cite{nguyen2015innovation}.
While all of these algorithms incorporate the divergent, goalless property of \picbreeder{} to a greater or lesser extent,
none of them have thus far explicitly reported forms canalization.
Examining the genomes produced by these algorithms,
and looking into the differences between them and \picbreeder{},
may shed more light on the origins of canalization and evolvability.

It is debatable to what extent the ever-shifting \picbreeder{} environment resembles natural evolution.
On the one hand, natural evolution involves the presence of many different niches and environments,
which may appear, disappear and change over 
time~\cite{potts1998variability, demenocal2004african, erwin2008macroevolution, odling2003niche},
thus resulting in divergent selection.
On the other hand, natural evolution also includes long periods in which environments remain stable,
emphasizing the effects of directional and stabilizing selection 
towards exploiting established niches~\cite{endler1986natural, lemos2005rates},
selective forces which are mostly absent in the Picbreeder system 
(there are few evolutionary advantages for an image to remain identical,
because users are unlikely to publish an image if they failed to produce at least some kind of change).
It is probably a combination of both forces that define natural evolution, 
which might explain why natural genomes contain both potential for variation,
but also feature many innovations whose main purpose is to reduce arbitrary variation and mutations in descendants.

Our understanding of the relationship between canalization and genomic structural organization remains incomplete.
It is clear that, in some cases, structural organization can directly lead to canalization.
However, we have also observed cases where networks with low modularity and hierarchy 
scores have canalized various dimensions of variation, 
and it is not hard to imagine a network that features structural organization, but not canalization.
As such, we expect these properties to be correlated, but not always causally related.
Given a good quantification of canalization it would be straightforward to test this hypothesis,
but for now that remains an open challenge for future work.

\section{Conclusion}

While ubiquitous in nature, canalizations 
--- the propensity of genomic structures to be mutationally robust against changes in some dimensions of variation
(ways in which individual or combinations of phenotypic traits can change), 
such that other, possibly more adaptive dimensions of variation, are free to vary ---
rarely emerge in computational simulation.
We have shown the emergence of canalization in the goalless and open-ended system of \picbreeder{},
a website for the interactive evolution of images.
An example was investigated where these canalizations are the result of structural organization in the genotype,
in the form of modularity and hierarchy,
and such genomic structural organization increases the reproductive success of individuals.
Lastly, we argued that the divergent, goal-free nature of \picbreeder{} may be an important driver 
for the spontaneous evolution of canalization.

\section{Acknowledgments}

Support came from the Santa Fe Institute to KS and JC, a National Science Foundation CAREER award (CAREER: 1453549) to JC, and a National Science Foundation Robust Intelligence grant no. IIS-1421925 to KS.  Any opinions, findings, and conclusions or recommendations expressed in this material are those of the authors and do not necessarily reflect the views of the National Science Foundation. 

\bibliographystyle{apa-good}

\bibliography{references_abbreviations_full,references}{}

\documentfooter

\end{document}